\documentclass{article}

\PassOptionsToPackage{numbers,compress}{natbib}

\usepackage[main,final]{neurips_2026}

\usepackage[utf8]{inputenc}
\usepackage[T1]{fontenc}
\usepackage{hyperref}
\usepackage{url}
\usepackage{booktabs}
\usepackage{amsfonts}
\usepackage{nicefrac}
\usepackage{microtype}
\usepackage{xcolor}

\usepackage{amsmath,amssymb,amsthm,mathtools,bm}
\usepackage{multirow,array,tabularx}
\usepackage{enumitem}
\usepackage{graphicx}
\usepackage{caption}
\usepackage{subcaption}
\usepackage[nameinlink,noabbrev]{cleveref}
\usepackage{algorithm}
\usepackage{algpseudocode}
\usepackage{setspace}
\usepackage{placeins}
\crefname{table}{Table}{Tables}
\Crefname{table}{Table}{Tables}
\crefname{figure}{Fig.}{Figs.}
\Crefname{figure}{Fig.}{Figs.}

\newtheorem{theorem}{Theorem}
\newtheorem{proposition}{Proposition}

\newtheorem{definition}{Definition}

\newcommand{\R}{\mathbb{R}}
\newcommand{\E}{\mathbb{E}}
\newcommand{\Cov}{\mathrm{Cov}}
\newcommand{\diag}{\mathrm{diag}}
\newcommand{\softmax}{\mathrm{softmax}}

\newif\ifcameraready\camerareadyfalse
\newcommand{\cameraready}[1]{\ifcameraready\textcolor{blue}{#1}\else#1\fi}
\newcommand{\citechk}[1]{\ifcameraready\colorbox{yellow}{\cite{#1}}\else\cite{#1}\fi}

\title{\textsc{Elpac}: Endpoint-Anchored Latent Progression with Stage-Varying Multimodal Coordination}

\author{%
  Shengxian Ding \\
  \texttt{naomi.ding@yale.edu} \\
  \And
  Yifei Zhang \\
  \texttt{yifei.zhang.yz2489@yale.edu} \\
  \And
  Xinyuan Tian \\
  \texttt{x.tian@yale.edu} \\
  \AND
  Yuanshuang Guo \\
  \texttt{yuanshuang.guo@yale.edu} \\
  \And
  Yize Zhao \\
  \texttt{yize.zhao@yale.edu} \\
  \AND
  \mdseries Department of Biostatistics, Yale School of Public Health, Yale University \\
  \mdseries New Haven, CT 06510
}

\begin{document}
\maketitle

\begin{abstract}
Aging cohorts increasingly combine imaging, plasma, and cognitive measurements, but most subjects provide only a partially observed cross-sectional snapshot, leaving latent stage, progression subtype, and stage-varying multimodal coordination to be inferred jointly. Existing methods estimate the cohort-level aging process, model covariance under observed covariates, or fuse modalities for prediction. However, they rarely target the joint problem of endpoint-anchored staging and progression-resolved residual coordination across modalities, even though subject staging, subtype assignment, and multimodal interplay are mutually dependent. We introduce \textsc{Elpac}, a Bayesian framework with two interlocking constructions. The first integrates subtype trajectories backward from clinically reliable endpoint groups along biologically structured velocity components encoding modality-specific monotone direction and optional sequential ordering priors. The second factorizes residual multimodal variation around the inferred stage axis into sparse coordination axes whose loadings and activations expose which multimodal patterns are operative in each subtype and when they emerge, with stage-resolved variation supplying the auxiliary structure for conditional identifiability. We use a scalable amortized variational inference scheme that enables joint posterior inference at cohort scale with single-pass deployment on new subjects. On synthetic data and a harmonized ADNI+NACC/SCAN Alzheimer's cohort, \textsc{Elpac} demonstrates accurate latent-structure recovery, well-calibrated prediction intervals, and biologically interpretable multimodal coordination programs that differ by subtype.
\end{abstract}

\section{Introduction}
Aging-related neurodegeneration is observed indirectly: pathology accumulates for years before symptoms emerge, and two patients indistinguishable on a cognitive test may differ markedly in pathology, structural change, and future risk \cite{Young2024,Jack2024}. For each subject, three latent quantities are of central scientific interest: the progression subtype that best describes them, including preserved aging; their continuous stage along that subtype trajectory; and the stage-specific structure of how measurements across modalities covary during progression. \cameraready{For example, consider two subjects with the same Mini-Mental State Examination (MMSE) score of 26, one 30\% along the early-onset AD (EOAD) amyloid+tau trajectory (stage $S_i{\approx}0.3$), the other far along healthy aging (stage $S_i{\approx}0.9$), which is advanced healthy aging rather than advanced disease.}

None of these quantities is directly observed. Aging studies now routinely collect brain imaging, plasma assays, and cognitive measurements \cite{Salvado2024,Palmqvist2025}, yet most participants contribute only a single, partially observed cross-sectional snapshot, without repeated measurements or follow-up diagnoses that would reveal where they lie along progression. This creates two coupled inference problems. First, supervision is asymmetric: endpoint groups are well characterized, while early and intermediate subjects, the part of the cohort whose stage and risk are most uncertain, carry the noisiest labels. A model that grows trajectories forward from this uncertain origin therefore rests on the cohort's weakest signal, whereas one anchored at the endpoints rests on its strongest. Second, dependence is non-stationary: the coupling between measurements changes along progression, so marginal correlations mix subjects from different latent stages and subtypes and can wash out biologically meaningful coordination. The target is therefore not subtype, stage, or covariance separately, but a joint posterior over subtype assignment, continuous stage, and stage-varying multimodal dependence, with uncertainty propagated across all of them.

Prior work has advanced each ingredient of this problem, but mostly in isolation. Disease progression models \cite{Fonteijn2012,Young2018,Venkatraghavan2019,Wijeratne2024,Koval2021,Schiratti2017,Lespinasse2023,Viani2025,Zhang2024TFM,Liu2025DiPro} recover event orderings or subtype-specific mean trajectories, targeting the expected trajectory, not the multivariate dependence around it. Condition-varying covariance and dynamic factor models \cite{Wilson2011,HoffNiu2012,Nejatbakhsh2023,Tu2024} learn covariance as a function of an \emph{observed} condition; in our setting that condition is latent. Multimodal biomedical architectures \cite{Chen2021MCAT,Chen2022Pathomic,WuGoodman2018PoE,Sutter2021MoPoE,Hemker2024} fuse heterogeneous data and absorb missingness, but read it through a predictive lens rather than a progression-resolved one. Identifiable latent variable learning \cite{HyvarinenMorioka2016,ZhengZhang2023,vonKugelgen2023,AndersonRubin1956,BhattacharyaDunson2011,Pati2014} shows that auxiliary structure resolves rotational and permutation ambiguities in factor models, but typically requires the auxiliary variable to be observed, not inferred. None learns the latent disease axis and the dependence around it as mutually identifying objects.

\begin{figure}[t]
    \centering
    \includegraphics[width=\linewidth]{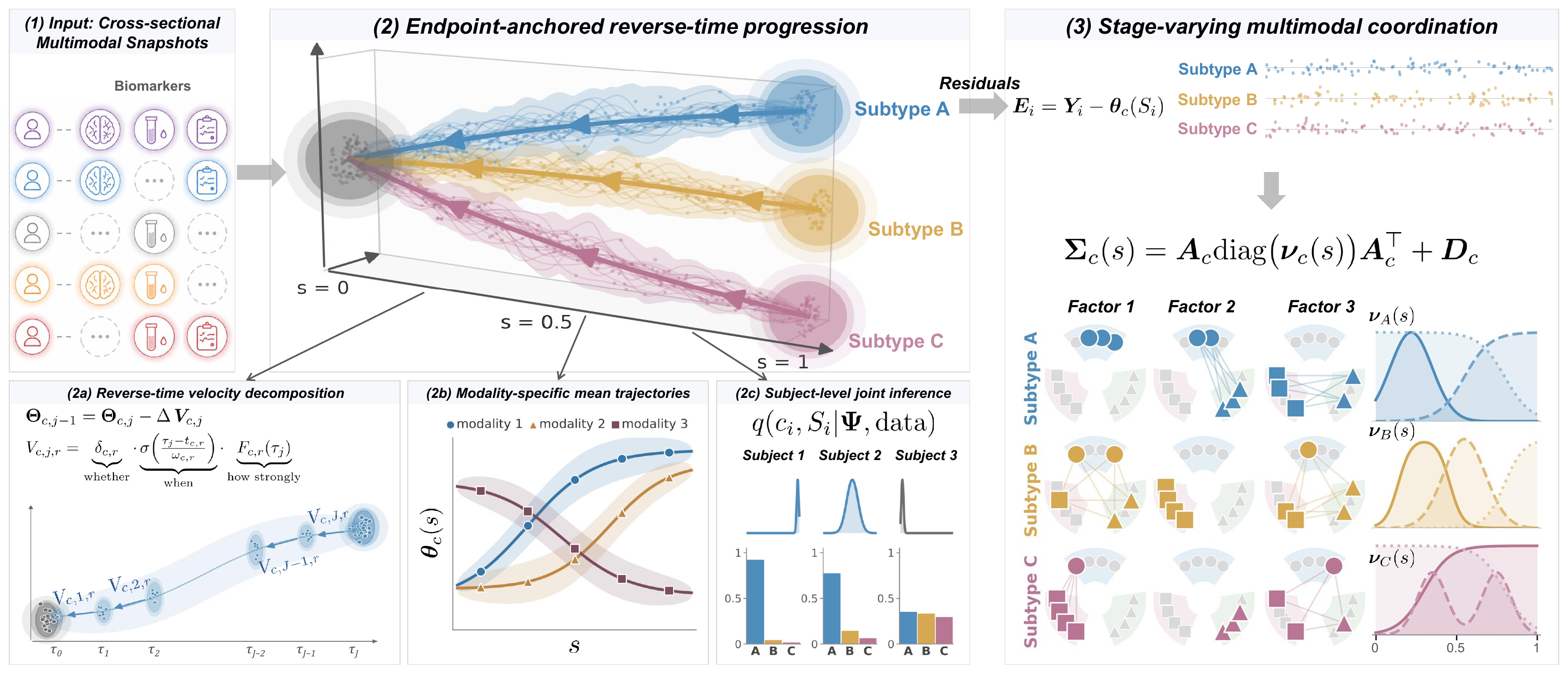}
    \caption{\textbf{\textsc{Elpac} workflow.} 
    Anchor-supervised multimodal cross-sectional data feed two interlocking constructions: backward integration from endpoints toward a shared healthy origin along biologically structured velocity yields subtype mean trajectories and the joint posterior over subtype and stage; residuals around the recovered stage axis factorize into sparse coordination axes whose loadings and activations expose \emph{which} multimodal programs operate per subtype and \emph{when} they emerge.}
    \label{fig:workflow}
\end{figure}

To address the gap, we introduce \textbf{\textsc{Elpac} (Endpoint-anchored Latent Progression And Coordination)}, a Bayesian framework with two interlocking constructions that recover the latent progression and its multimodal coordination from cross-sectional snapshots (\cref{fig:workflow}). 
We integrate subtype trajectories backward from clinically reliable endpoint groups towards a shared healthy reference along per-modality velocity components that encode monotone progression, modality-specific direction, and optional sequential ordering priors. These biological constraints supply the inductive bias absent from cross-sectional snapshots, while endpoint anchoring calibrates the stage axis from the cohort's best supervised subjects. 
Around that axis, residual multimodal variation decomposes into coordination axes, each a sparse coalition of regions and modalities whose covariation strengthens or attenuates with stage; the loadings expose \emph{which} programs are operative in each subtype, the activations expose \emph{when} they emerge. 
The two constructions are fitted jointly: stage-resolved variation supplies the auxiliary structure for the coordination decomposition, and the resolved coordination feeds back through the multivariate likelihood to sharpen the posterior over subtype and stage. 
We use a scalable amortized variational scheme paired with structured low-rank updates over the global parameters, enabling joint posterior inference at cohort scale with single-pass deployment on new subjects.

These design choices yield three contributions:
\begin{itemize}[leftmargin=*,itemsep=1pt,topsep=2pt]
  \item We incorporate asymmetric endpoint supervision as an inductive bias for subject-level continuous stage inference, with subtype trajectories anchored at clinically reliable groups and integrated backward through biologically structured velocity components.
  \item We establish conditional identifiability of the stage-varying multimodal coordination via the inferred stage axis as auxiliary variation, yielding interpretable sparse coordination axes across and within multimodal measurements.
  \item We validate \textsc{Elpac} on synthetic data and on a harmonized ADNI~+~NACC/SCAN cohort ($N=7{,}501$; \cameraready{NACC} March~2026 data freeze), demonstrating accurate latent-structure recovery, well-calibrated prediction intervals, and biologically interpretable coordination programs that differ by subtype.
\end{itemize}

\section{Method}\label{sec:method}
\begin{table}[t]
{\ifcameraready\color{blue}\fi
\centering\footnotesize\setlength{\tabcolsep}{2pt}
\caption{\cameraready{\textbf{Notation used in \cref{sec:method}.}}}
\label{tab:notation}
\begin{tabularx}{\linewidth}{@{}l X l X@{}}
\toprule
Symbol & Meaning & Symbol & Meaning\\
\midrule
$c_i$; $S_i\in(0,1)$ & subtype; progression stage & $C$; $R$; $m(r)\in\mathcal{M}$ & no.\ subtypes, biomarkers; modality of $r$\\
$\bm{Y}_i$; $\mathcal{O}_i$ & biomarker vector; observed entries & $\bm{X}_i$; $Y_i^{\mathrm{cog}}$; $\chi_c(s)$ & covariates; cognition; its stage trend\\
$\mathcal{A}_c$; $\mathcal{H}$; $\mathcal{U}$ & anchors; healthy reference; unlabeled & $s_A$; $s_H$ & anchor; healthy-reference stage location\\
$\bm{\theta}_c(s)$; $\bm{\Sigma}_c(s)$ & mean trajectory; within-stage covariance & $\bm{\mu}_c$; $\bm{\Theta}_0$ & subtype endpoint; shared healthy origin\\
$\bm{\Theta}_{c,j}$; $\bm{V}_{c,j}$; $J$ & shell state; shell velocity; no.\ shells & $\mathrm{sgn}_{m}$; $\tilde t_{c,m}$ & modality direction; modality onset center\\
$\delta_{c,r}$; $t_{c,r}$; $\omega_{c,r}$ & inclusion; onset timing; onset sharpness & $F_{c,r}$; $B_q$, $Q$ & signed velocity magnitude; spline basis, size\\
$\bm{E}_i$; $\bm{z}_i$; $\bm{\eta}_i$ & residual; factor scores; idiosyncratic noise & $\bm{A}_c$; $\bm{\nu}_c(s)$; $\bm{D}_c$ & loadings; activations; noise covariance\\
$K$; $\psi_p$, $P$ & no.\ factors; activation B-spline basis, size & $\bm{\Psi}$; $q_{\phi,\psi}$ & global parameters; variational family\\
\bottomrule
\end{tabularx}}
\end{table}

\subsection{Problem formulation}\label{sec:problem}
We observe a cross-sectional cohort $\mathcal{D}=\{(\bm{Y}_i,\bm{X}_i,Y_i^{\mathrm{cog}})\}_{i=1}^n$, where $\bm{Y}_i\in\R^R$ is a partially observed multimodal biomarker vector with each coordinate $r\in\{1,\ldots,R\}$ assigned to a modality $m(r)\in\mathcal{M}$, $\bm{X}_i\in\R^{p}$ collects demographic and genetic covariates, and $Y_i^{\mathrm{cog}}\in\R$ is an auxiliary cognitive outcome that may be missing. Each subject carries a latent subtype $c_i\in\{1,\ldots,C\}$, with subtypes defined by clinically distinct endpoint phenotypes, and a latent progression stage $S_i\in(0,1)$. The cohort comprises three strata: endpoint anchors $\mathcal{A}_c$ with clinically confirmed subtype $c$ (near $s\!\to\!1$); a healthy reference set $\mathcal{H}$ near $s\!\approx\!0$; and a general unsupervised population $\mathcal{U}$ with unknown $(c_i,S_i)$. The objects of inference are the subject-level latents $(c_i,S_i)$, the subtype-specific mean trajectories $\bm{\theta}_c(s)\in\R^R$, and the stage-varying within-stage covariance $\bm{\Sigma}_c(s)\in\R^{R\times R}$. Boundary supervision from $\mathcal{A}_c$ and $\mathcal{H}$ grounds the latent stage, and the resulting stage axis serves as the auxiliary variation under which $\bm{\Sigma}_c(s)$ is identifiable. \cameraready{\Cref{tab:notation} summarizes the notation.}

\subsection{Observation model}\label{sec:observation_model}
Let $M_{ir}\in\{0,1\}$ indicate whether biomarker $r$ is observed for subject $i$, with $\mathcal{O}_i=\{r:M_{ir}=1\}$, and let $O_i^{\mathrm{cog}}\in\{0,1\}$ indicate whether the auxiliary cognitive outcome $Y_i^{\mathrm{cog}}$ is observed. The observed biomarker likelihood is
\begin{equation}
\bm{Y}_{i,\mathcal{O}_i}\mid c_i,S_i \sim \mathcal{N}\!\left( \bm{\theta}_{c_i,\mathcal{O}_i}(S_i),\, \bm{\Sigma}_{c_i,\mathcal{O}_i}(S_i) \right),
\label{eq:observation}
\end{equation}
where $\bm{\theta}_c(s)\in\R^R$ and $\bm{\Sigma}_c(s)\in\R^{R\times R}$ denote the subtype-$c$ mean trajectory and within-stage covariance. The subscript $\mathcal{O}_i$ selects the observed entries and corresponding submatrix, so unobserved biomarker coordinates are integrated out under the multivariate Gaussian model. 
When $O_i^{\mathrm{cog}}{=}1$, the cognitive outcome enters as
\begin{equation}
Y_i^{\mathrm{cog}}\mid c_i,S_i,\bm{X}_i \sim \mathcal{N}\!\left(\gamma_0+\bm{\gamma}_X^\top \bm{X}_i+\chi_{c_i}(S_i),\,\sigma_{\mathrm{cog}}^2\right),
\label{eq:cog}
\end{equation}
where $\gamma_0\!\sim\!\mathcal{N}(0,1)$, $\bm{\gamma}_X\!\sim\!\mathcal{N}(\bm{0},\bm{I}_p)$, $\log\sigma_{\mathrm{cog}}\!\sim\!\mathcal{N}(0,1)$, and $\chi_c(s)=\bm{b}^{\mathrm{I}}(s)^\top\bm{\xi}_c$ is a monotone I-spline with non-negative coefficients $\xi_{c,j}\!\sim\!\mathrm{Half\text{-}Normal}(1)$ that encodes the monotone progression-conditioned trend of $Y_i^{\mathrm{cog}}$. The corresponding factor is omitted from the joint likelihood when $O_i^{\mathrm{cog}}{=}0$.

\subsection{Construction I: endpoint-anchored progression module}\label{sec:construction1}
The subtype mean trajectory $\bm{\theta}_c(s)$ is defined as a reverse-time integral on $[0,1]$ between two latent boundary states. Partition the interval by shells $0{=}\tau_0{<}\cdots{<}\tau_J{=}1$ with width $\Delta{=}1/J$, and let $\bm{\theta}_c(s)=\mathcal{I}\bigl(\{\bm{\Theta}_{c,j},\bm{V}_{c,j}\}_{j=0}^J;s\bigr)$ denote the cubic Hermite interpolant of shell states $\bm{\Theta}_{c,j}\!\in\!\R^R$ and shell-endpoint velocities $\bm{V}_{c,j}\!\in\!\R^R$ (explicit form in \cref{eq:hermite}) satisfying
\begin{equation}
\bm{\Theta}_{c,j-1}=\bm{\Theta}_{c,j}-\Delta\,\bm{V}_{c,j},\qquad j=J,J-1,\ldots,1.
\label{eq:backward}
\end{equation}
This right-endpoint update interprets $\bm{V}_{c,j}$ as the arrival velocity at shell boundary $\tau_j$, yielding a triangular reverse-time construction in which $\bm{\Theta}_{c,j}$ depends on $(\bm{\Theta}_{c,J},\bm{V}_{c,j+1:J})$ alone. We assign weakly informative Gaussian hyperpriors to the subtype endpoint $\bm{\mu}_c$ and the shared healthy-reference origin $\bm{\Theta}_0$,
\begin{equation}
\bm{\mu}_c\sim\mathcal{N}(\bm{0},\tau_\mu^2 \bm{I}_R),\quad \bm{\Theta}_{c,J}\mid\bm{\mu}_c\sim\mathcal{N}(\bm{\mu}_c,\sigma_T^2 \bm{I}_R),\quad \bm{\Theta}_0\sim\mathcal{N}(\bm{0},\tau_0^2 \bm{I}_R).
\label{eq:boundary}
\end{equation}
The recursion \cref{eq:backward} fixes $\bm{\Theta}_{c,0}$ deterministically as the integrated origin $\bm{\Theta}_{c,0}=\bm{\Theta}_{c,J}-\Delta\sum_{j=1}^J \bm{V}_{c,j}$, and we attach a Gaussian coupling potential that draws each subtype's integrated origin toward the shared healthy-reference origin,
\begin{equation}
\log\psi_{\mathrm{soft}}\bigl(\bm{\Theta}_0,\{\bm{\Theta}_{c,J},\bm{V}_{c,1:J}\}_{c=1}^{C}\bigr) \;=\; -\tfrac{1}{2\sigma_0^2}\sum_{c=1}^{C}\bigl\|\bm{\Theta}_{c,0}(\bm{\Psi})-\bm{\Theta}_0\bigr\|^2 + \mathrm{const},
\label{eq:origin-soft}
\end{equation}
where $\sigma_0$ controls the strength of the coupling\cameraready{: all subtypes depart from a common healthy state, while each subtype's integrated origin may deviate from it}. \cref{eq:origin-soft} is a coupling factor on the joint over $(\bm{\Theta}_0,\{\bm{\Theta}_{c,J},\bm{V}_{c,1:J}\})$, not a conditional density on the deterministic $\bm{\Theta}_{c,0}(\bm{\Psi})$; combined with the Gaussian boundary priors of \cref{eq:boundary} it yields a normalized joint prior since $\psi_{\mathrm{soft}}\!\le\!1$ pointwise. The posteriors of $\bm{\mu}_c$ and $\bm{\Theta}_0$ are informed by the anchor and healthy-reference cohorts $\mathcal{A}_c$ and $\mathcal{H}$ through the observation likelihood \cref{eq:observation} evaluated at the boundary stages introduced below. The recursion \cref{eq:backward} together with \cref{eq:origin-soft} then determines $\bm{\theta}_c(s)$ from the velocity field $\bm{V}_{c,1:J}$.

The velocity admits a biologically structured decomposition,
\begin{equation}
V_{c,j,r}=\underbrace{\delta_{c,r}}_{\text{whether}}\cdot\underbrace{\sigma\!\left(\tfrac{\tau_j-t_{c,r}}{\omega_{c,r}}\right)}_{\text{when}}\cdot\underbrace{F_{c,r}(\tau_j)}_{\text{how strongly}},\quad F_{c,r}(\tau_j)=\mathrm{sgn}_{m(r)}\cdot\mathrm{softplus}\!\left(\sum_{q=1}^Q B_q(\tau_j)\,\beta_{c,r,q}\right),
\label{eq:velocity}
\end{equation}
in which $\delta_{c,r}\!\in\!\{0,1\}$ encodes biomarker inclusion in subtype $c$, $(t_{c,r},\omega_{c,r})$ govern onset timing and sharpness, $\{B_q\}$ is a $Q$-dimensional B-spline basis with coefficients $\{\beta_{c,r,q}\}$, and the modality sign $\mathrm{sgn}_{m(r)}\!\in\!\{+1,-1\}$ is a fixed modality-level constant encoding the direction of evolution (e.g.,\ monotone accumulation versus monotone atrophy). The associated priors are
\begin{equation}
\delta_{c,r}\sim\mathrm{Bern}(\pi_{c,m(r)}),\quad \pi_{c,m}\sim\mathrm{Beta}(2,1),\quad t_{c,r}{=}\sigma(\eta_{c,r}),\;\eta_{c,r}\sim\mathcal{N}(\tilde t_{c,m(r)},\sigma_\eta^2),
\label{eq:vel-prior-1}
\end{equation}
\begin{equation}
\tilde t_{c,m}\sim\mathcal{N}(0.5,1),\quad \omega_{c,r}{=}e^{\varrho_{c,r}},\;\varrho_{c,r}\sim\mathcal{N}(m_\omega,\sigma_\varrho^2),\quad \beta_{c,r,q}\sim\mathcal{N}(0,\sigma_\beta^2).
\label{eq:vel-prior-2}
\end{equation}
The modality-monotone direction encoded by $\mathrm{sgn}_{m(r)}$ is an inductive bias common to event-based progression models~\cite{Fonteijn2012,Young2018}. Domain knowledge enters through the modality-level onset centers $\tilde t_{c,m}$: when biomarker ordering or late-onset behavior is supported by prior evidence, the diffuse $\mathcal{N}(0.5,1)$ on $\tilde t_{c,m}$ is replaced by an informative location or ordering prior.

Subtype assignment is modeled as a covariate-modulated categorical, and the subtype-conditional stage as logit-normal. For unsupervised subjects $i\!\in\!\mathcal{U}$,
\begin{equation}
c_i\mid \bm{X}_i\sim\mathrm{Cat}\!\bigl(\softmax(\bm{\alpha} + \bm{B} \bm{X}_i)\bigr),\qquad \mathrm{logit}(S_i)\mid i\in\mathcal{U},\,c_i{=}c,\,\bm{X}_i\sim\mathcal{N}\!\bigl(\bm{b}_c^\top \bm{X}_i,\,\sigma_c^2\bigr),
\label{eq:subtypestageprior}
\end{equation}
with $\bm{\alpha}\!\sim\!\mathcal{N}(\bm{0},\bm{I}_C)$, $\bm{B}\!\sim\!\mathcal{N}(\bm{0},\bm{I}_{C\times p})$, $\bm{b}_c\!\sim\!\mathcal{N}(\bm{0},\bm{I}_p)$, and $\sigma_c\!\sim\!\mathrm{Half\text{-}Cauchy}(0,1)$. The diffuse prior on $\bm{b}_c$ is replaced by an informative location prior on the relevant coordinate when subtype-specific covariate effects (e.g., age-of-onset) are supported by external evidence. For anchored subjects $i\!\in\!\mathcal{A}_c$, the subtype is fixed at $c$ and the stage is concentrated near the endpoint,
\begin{equation}
\mathrm{logit}(S_i)\mid i\in\mathcal{A}_c\sim\mathcal{N}\!\bigl(\mathrm{logit}(s_A),\,\epsilon_S^2\bigr),
\label{eq:anchor}
\end{equation}
and for healthy-reference subjects $i\!\in\!\mathcal{H}$, the subtype is unconstrained and the stage is concentrated near the origin,
\begin{equation}
\mathrm{logit}(S_i)\mid i\in\mathcal{H},\,c_i{=}c\sim\mathcal{N}\!\bigl(\mathrm{logit}(s_H),\,\epsilon_S^2\bigr)\quad\forall c.
\label{eq:control}
\end{equation}
Here $s_A, ~ s_H \in (0,1)$ are hyperparameters localizing anchors near the endpoint and healthy subjects near the origin. 

\subsection{Construction II: stage-varying coordination module}\label{sec:construction2}
With $(c,s)$ latent, the iterated law of total covariance separates raw pairwise correlation into a within-stage coordination term and a between-$(c,s)$ mean effect,
\begin{equation}
\Cov(Y_r, Y_{r'}) = \underbrace{\E_{c,s}\!\left[\Cov(Y_r, Y_{r'}\mid c, s)\right]}_{\text{within-stage coordination}} + \underbrace{\Cov_{c,s}\!\left(\E[Y_r\mid c, s],\,\E[Y_{r'}\mid c, s]\right)}_{\text{between-}(c,s)\text{ mean effect}}.
\label{eq:ltc}
\end{equation}
The second term is captured by the mean trajectory $\bm{\theta}_c(s)$ of Construction~I. We model the first by endowing the residual $\bm{E}_i\equiv \bm{Y}_i-\bm{\theta}_{c_i}(S_i)$ with a stage-varying latent factor representation,
\begin{equation}
\bm{E}_i=\bm{A}_{c_i}\diag\!\bigl(\bm{\nu}_{c_i}(S_i)\bigr)^{1/2}\bm{z}_i+\bm{\eta}_i,\qquad \bm{z}_i\sim\mathcal{N}(\bm{0},\bm{I}_K),\quad \bm{\eta}_i\sim\mathcal{N}(\bm{0},\bm{D}_{c_i}),
\label{eq:latentfactor}
\end{equation}
where $\bm{z}_i\!\in\!\R^K$ are latent factor scores and $\bm{\eta}_i$ is idiosyncratic noise. Each column $\bm{A}_{c,\cdot,k}$ defines a coordination axis, encoding \emph{which} biomarkers and modalities co-vary, and the activation $\nu_{c,k}(s)$ governs \emph{when} along progression that axis strengthens or vanishes; a column may concentrate within a single modality or span multiple modalities, capturing within-modality and across-modality coordination respectively. The induced within-stage covariance has the rank-$K$-plus-diagonal form
\begin{equation}
\bm{\Sigma}_c(s)=\bm{A}_c\diag\!\bigl(\bm{\nu}_c(s)\bigr)\bm{A}_c^\top + \bm{D}_c.
\label{eq:sigma_factor}
\end{equation}

The loadings matrix $\bm{A}_c\!\in\!\R^{R\times K}$ is constrained lower triangular with positive diagonal, fixing the rotational ambiguity inherent to factor models \cite{AndersonRubin1956,Geweke1996}. A multiplicative-gamma cascade \cite{BhattacharyaDunson2011,Pati2014} adaptively controls the effective number of factors:
$\kappa_{c,1}\!\sim\!\mathrm{Gamma}(2,1)$, $\kappa_{c,h\ge 2}\!\sim\!\mathrm{Gamma}(a_2,1)$ with $a_2{>}1$, $\Lambda_{c,k}{=}\prod_{h\le k}\kappa_{c,h}$, and, conditional on $\Lambda_{c,k}$,
$A_{c,r,k}\!\sim\!\mathcal{N}(0,\Lambda_{c,k}^{-1})$ for $r{>}k$, $A_{c,k,k}\!\sim\!\mathrm{Half\text{-}Normal}(\Lambda_{c,k}^{-1/2})$. Since $\E[\Lambda_{c,k}]$ grows with $k$, later columns are increasingly shrunk toward zero. The activation $\nu_{c,k}(s){=}\exp\bigl(\sum_p w_{c,k,p}\psi_p(s)\bigr)$ is expanded on a $P$-dimensional B-spline basis $\{\psi_p\}$. A constant shift in $\log\nu_{c,k}$ may be absorbed into $\bm{A}_{c,\cdot,k}$, so we resolve the resulting scale ambiguity by supporting $\bm{w}_{c,k}$ on the centered subspace $\mathcal{P}_0{=}\{\bm{w}\!\in\!\R^P:\mathbf{1}^\top \bm{w}{=}0\}$, via the projection $\bm{\Pi}_0{=}\bm{I}_P{-}\tfrac{1}{P}\mathbf{1}\mathbf{1}^\top$, with the singular Gaussian prior
\begin{equation}
\bm{w}_{c,k}\sim\mathcal{N}\!\bigl(\bm{0},\,\tau_k^2\,\bm{\Pi}_0(\bm{K}_B+\epsilon_B \bm{I}_P)^{-1}\bm{\Pi}_0\bigr),\qquad \tau_k\sim\mathrm{Half\text{-}Normal}(\sigma_\tau),
\label{eq:nu}
\end{equation}
where $\bm{K}_B$ denotes the second-difference smoothness penalty matrix\cameraready{, so coordination strengthens or fades gradually along progression rather than jumping between adjacent stages,} and $\epsilon_B$ is a small ridge\cameraready{; $\tau_k$ scales how strongly factor $k$'s activation varies with stage}. The prior covariance has rank $P{-}1$ and is supported on $\mathcal{P}_0$, so the centering identity $\mathbf{1}^\top \bm{w}_{c,k}{=}0$ holds almost surely. \cameraready{$\bm{D}_c$ is diagonal, so all within-stage cross-biomarker dependence passes through the coordination axes; the} idiosyncratic noise variances follow a log-Gaussian prior, $\log d_{c,r}\!\sim\!\mathcal{N}(m_D,\sigma_D^2)$, with $\bm{D}_c{=}\diag(d_{c,r}+d_{\min})$ and a small positive floor $d_{\min}$ \cameraready{that keeps every noise variance bounded away from zero}. Downstream summaries, including stage-dependent pairwise correlations $\rho_c(r,r';s)$ and Fisher modality informativeness $w^{\mathrm{FI}}_{c,m}(s)$, are deterministic functionals of $(\bm{A}_c,\bm{\nu}_c(\cdot),\bm{D}_c)$.

\subsection{Conditional identifiability}\label{sec:identifiability}
The latent stage recovered through Construction~I supplies the auxiliary variation under which the Construction~II decomposition $(\bm{A}_c,\bm{\nu}_c(\cdot),\bm{D}_c)$ is identifiable, and the resolved $\bm{\Sigma}_c(s)$ feeds back through the likelihood \cref{eq:observation} to sharpen the $(c_i,S_i)$ posterior; the two blocks are fitted jointly.
\begin{theorem}[Conditional identifiability of the dynamic factor block]\label{thm:identifiability}
Fix a subtype and suppress its index. Within the structural model of \cref{eq:sigma_factor,eq:nu} ($\bm{A}\in\R^{R\times K}$ lower triangular with positive diagonal; activation spline coefficients $\bm{w}_k$ satisfying $\mathbf{1}^\top \bm{w}_k=0$; $\bm{D}$ diagonal with strictly positive entries; data dimension $R\ge 2K+1$; activations $\nu_k(\cdot)$ non-constant; plus the generic regularity conditions of \cref{app:thm_proof}), suppose the within-stage covariance is identified at two stages $s_1\ne s_2$ whose activation ratios are pairwise distinct,
\begin{equation}
\frac{\nu_j(s_2)}{\nu_j(s_1)}\;\neq\;\frac{\nu_k(s_2)}{\nu_k(s_1)}\qquad\text{for all }j\neq k.
\label{eq:ratio}
\end{equation}
Then the loadings $\bm{A}$, the activation function $\bm{\nu}(\cdot)$, and the idiosyncratic noise $\bm{D}$ are uniquely determined.
\end{theorem}

\emph{Proof sketch.} \textbf{(i)} At each fixed $s$, the Anderson--Rubin row-deletion condition with lower-triangular $\bm{A}$ and $R\ge 2K{+}1$ identifies $\bm{D}$ and $\bm{A}\diag(\bm{\nu}(s))\bm{A}^\top$ \cite{AndersonRubin1956,Geweke1996}. \textbf{(ii)} Setting $\bm{\Pi}_\bullet{:=}\bm{\Sigma}(s_\bullet)-\bm{D}$, the generalized eigenproblem $\bm{\Pi}_2\bm{v}{=}\lambda\,\bm{\Pi}_1\bm{v}$ has eigenvalues $\nu_k(s_2)/\nu_k(s_1)$, distinct by \cref{eq:ratio}; primal columns recover as $\tilde{\bm{a}}_k{=}\bm{\Pi}_1\bm{v}_k/\|\bm{\Pi}_1\bm{v}_k\|_2$, and the lower-triangular positive-diagonal normalization pins scale and sign. \textbf{(iii)} The centering identity $\mathbf{1}^\top \bm{w}_k{=}0$ (with partition-of-unity $\{\psi_p\}$) pins $\bm{\nu}(s_1),\bm{\nu}(s_2)$ pointwise. \textbf{(iv)} Spline-design coverage extends $\bm{\nu}(\cdot)$ to all of $[0,1]$; boundary anchoring \cref{eq:anchor,eq:control} together with the trajectory-feature regularity rules out non-identity boundary-fixing warps. Full proof, warp invariance, and full-model identification in \cref{app:ident}.

\subsection{Inference}\label{sec:inference}
Let $\bm{\Psi}$ denote the global parameters of \crefrange{eq:observation}{eq:control} and $\mathcal{Z}{=}\{(c_i,S_i)\}_{i=1}^n$ the subject-level latents. We approximate $p(\bm{\Psi},\mathcal{Z}\mid\mathcal{D})$ by a structured variational family that pairs a low-rank multivariate Gaussian on $\bm{\Psi}$ with $C$ subtype-conditional Gaussians on $\mathrm{logit}(S_i)$ per subject,
\begin{equation}
\begin{aligned}
q_{\phi,\psi}\bigl(\bm{\Psi},\{S_i^{(c)}\}\bigr) \;=\;& \mathcal{N}\!\bigl(\bm{\Psi};\,\bm{\mu}_\phi,\,\bm{W}_\phi \bm{W}_\phi^\top + \diag(\bm{d}_\phi^2)\bigr) \\
&\times \prod_{i=1}^n\prod_{c=1}^{C} \mathcal{N}\!\bigl(\mathrm{logit}(S_i^{(c)});\,\mu_\psi^{(c)}(\bm{Y}_{i,\mathcal{O}_i},\bm{X}_i),\,\tau_\psi^{(c)}(\bm{Y}_{i,\mathcal{O}_i},\bm{X}_i)^2\bigr),
\end{aligned}
\label{eq:guide}
\end{equation}
where $\bm{W}_\phi\!\in\!\R^{|\bm{\Psi}|\times r_g}$ is the low-rank factor of the global Gaussian (rank $r_g$), and $(\mu_\psi^{(c)},\tau_\psi^{(c)})$ are produced by an amortized recognition network with weights $\psi$ shared across subjects \cite{KingmaWelling2014VAE,RezendeMohamed2014}.
Marginalizing $c_i$ exactly inside the evidence lower bound gives the collapsed bound
\begin{equation}
\begin{aligned}
\mathcal{L}(\phi,\psi) \;=\;& \sum_{i=1}^n \E_{q_{\phi,\psi}}\!\Bigl[\log\sum_{c=1}^{C} p(c\mid \bm{X}_i,\bm{\Psi})\,\ell_i^{(c)}\bigl(\bm{\Psi},S_i^{(c)}\bigr)\Bigr] \\
&- \mathrm{KL}\bigl(q_\phi^{\mathrm{glob}}(\bm{\Psi})\,\|\,p(\bm{\Psi})\bigr) - \sum_{i,c}\E_{q_\phi^{\mathrm{glob}}}\!\Bigl[\mathrm{KL}\bigl(q_\psi(S_i^{(c)}\mid \bm{Y}_i,\bm{X}_i)\,\|\,p(S_i\mid c_i{=}c,\bm{X}_i,\bm{\Psi})\bigr)\Bigr],
\end{aligned}
\label{eq:elbo}
\end{equation}
with per-candidate-subtype data factor $\ell_i^{(c)}{=}p(\bm{Y}_{i,\mathcal{O}_i}\mid c,S_i^{(c)},\bm{\Psi})\,p(Y_i^{\mathrm{cog}}\mid c,S_i^{(c)},\bm{X}_i,\bm{\Psi})^{O_i^{\mathrm{cog}}}$. The expectations in \cref{eq:elbo} are estimated by single-sample Monte Carlo with reparameterized noise,
\begin{equation}
\bm{\Psi} = \bm{\mu}_\phi + \bm{W}_\phi\bm{\epsilon}^{(W)} + \bm{d}_\phi\odot\bm{\epsilon}^{(d)},\qquad \mathrm{logit}(S_i^{(c)}) = \mu_\psi^{(c)} + \tau_\psi^{(c)}\,\epsilon_i^{(c)},
\label{eq:reparam}
\end{equation}
with $\bm{\epsilon}^{(W)}\!\sim\!\mathcal{N}(\bm{0},\bm{I}_{r_g})$, $\bm{\epsilon}^{(d)}\!\sim\!\mathcal{N}(\bm{0},\bm{I}_{|\bm{\Psi}|})$, and $\epsilon_i^{(c)}\!\sim\!\mathcal{N}(0,1)$, and $(\phi,\psi)$ are optimized with Adam \cite{KingmaBa2015Adam}. Each $\ell_i^{(c)}$ is evaluated under the rank-$K$-plus-diagonal covariance $\bm{\Sigma}_{c,\mathcal{O}_i}(S_i^{(c)})$ via the Sylvester--Woodbury identity \cite{HendersonSearle1981} at cost $O(RK^2{+}K^3)$. \Cref{alg:svi} provides the pseudocode.

The subtype posterior is
\begin{equation}
\hat p(c_i{=}c\mid \bm{Y}_i,\bm{X}_i,\mathcal{D}) = \E_{q_\phi^{\mathrm{glob}}(\bm{\Psi})}\!\left[\frac{p(c\mid \bm{X}_i,\bm{\Psi})\,\cameraready{\int_0^1 p(s\mid c,\bm{X}_i,\bm{\Psi})\,\ell_i^{(c)}(\bm{\Psi},s)\,\mathrm{d}s}}{\sum_{c'}p(c'\mid \bm{X}_i,\bm{\Psi})\,\cameraready{\int_0^1 p(s\mid c',\bm{X}_i,\bm{\Psi})\,\ell_i^{(c')}(\bm{\Psi},s)\,\mathrm{d}s}}\right],
\label{eq:posterior_c}
\end{equation}
and the subtype-conditional stage posterior is $\hat p(S_i\mid c_i{=}c, \bm{Y}_i, \bm{X}_i, \mathcal{D}){=}q_\psi(S_i^{(c)}\mid \bm{Y}_i, \bm{X}_i)$. Marginalizing the subtype yields a $C$-component Gaussian mixture on $\mathrm{logit}(S_i)$,
\begin{equation}
\hat p(S_i\mid \bm{Y}_i,\bm{X}_i,\mathcal{D}) = \sum_{c=1}^{C} \hat p(c_i{=}c\mid \bm{Y}_i,\bm{X}_i,\mathcal{D})\,q_\psi(S_i^{(c)}\mid \bm{Y}_i,\bm{X}_i).
\label{eq:posterior_joint}
\end{equation}
Posterior expectations and credible intervals are obtained from \cref{eq:posterior_c,eq:posterior_joint} by Monte Carlo over $\bm{\Psi}\!\sim\!q_\phi^{\mathrm{glob}}$, with the inner \cameraready{stage integral of $\ell_i^{(c)}$ against the stage prior $p(s\mid c,\bm{X}_i,\bm{\Psi})$} computed by quadrature on a logit-stage grid. For a held-out subject the encoder is evaluated in a single forward pass.

\section{Experiments}\label{sec:experiments}
We evaluate \textsc{Elpac} along two axes that test the two constructions of \cref{sec:method}. Synthetic experiments\footnote{Code and synthetic-experiment reproduction: \cameraready{\url{https://github.com/Naomi-Ding/ELPAC}}}
measure recoverability of the anchored stage and the stage-varying factor decomposition against ground truth. The harmonized ADNI+NACC/SCAN evaluation measures held-out generative fit, subtype and stage recovery, predictive-interval (PI) coverage of biomarker observations, and the biological structure of the recovered coordination programs in a regime where most subjects are partially observed and unlabeled. 
We compare \textsc{Elpac} against representative baselines per relevant family: SuStaIn~\cite{Young2018} and DEBM~\cite{Venkatraghavan2019} represent subtype-and-stage and event-based progression models, Static-FA~\cite{AndersonRubin1956} separates factor analysis from downstream subtype or stage prediction, MoPoE-VAE~\cite{Sutter2021MoPoE} and HEALNet~\cite{Hemker2024} represent multimodal neural fusion, while XGBoost~\cite{Chen2016XGBoost} and label propagation~\cite{Zhu2002LabelProp} directly use the available endpoint labels and serve as supervised reference. 

\textbf{Metrics.}
We report held-out NLL/cell for generative fit; reconstruction RMSE and masked NLL (one modality held out) for imputation; BAcc and Macro-F1 for subtype recovery; Spearman $\rho$ of recovered stage against ground-truth latent (synthetic) or MMSE (real); AUC against case--control labels for diagnostic discrimination; $|\mathrm{Cog}\,r|$ for cognition alignment; and PI coverage at the $90\%$ nominal level for calibration. Metrics scale by $100$; \textbf{best} bolded, \underline{second} underlined.

\begin{table}
\centering
\scriptsize
\setlength{\tabcolsep}{3pt}
\caption{\textbf{Synthetic benchmark: \textsc{Elpac} recovers subtype, stage, diagnostic, and cognition on synthetic cohorts.} $N{=}1{,}000$ over $100$ replicate cohorts. }
\label{tab:synth}
\begin{tabularx}{\linewidth}{X c c c c c c c}
\toprule
Method & NLL/cell\,$\downarrow$ & BAcc\,$\uparrow$ & Macro-F1\,$\uparrow$ & Stage-score $\rho$\,$\uparrow$ & AUC$_{\{1,2\}/3}$\,$\uparrow$ & Masked NLL\,$\downarrow$ & $|$Cog $r|$\,$\uparrow$\\
\midrule
SuStaIn & -- & $44.0{\pm}0.8$ & $43.3{\pm}0.8$ & $45.4{\pm}0.6$ & $54.9{\pm}1.3$ & -- & $47.7{\pm}0.6$\\
DEBM & -- & -- & -- & $38.4{\pm}0.8$ & $71.5{\pm}0.6$ & -- & $13.4{\pm}0.6$\\
Static-FA & -- & $60.4{\pm}0.8$ & $59.2{\pm}0.9$ & $27.5{\pm}1.2$ & $52.1{\pm}1.2$ & -- & $42.1{\pm}1.3$\\
MoPoE-VAE & $\underline{1.47{\pm}0.01}$ & $47.8{\pm}0.8$ & $43.8{\pm}0.9$ & $39.7{\pm}0.9$ & $38.1{\pm}0.7$ & $\underline{1.46{\pm}0.01}$ & $38.5{\pm}0.9$\\
XGBoost & -- & $87.5{\pm}0.5$ & $86.4{\pm}0.5$ & $\underline{95.6{\pm}0.0}$ & $75.9{\pm}2.3$ & -- & $68.2{\pm}0.8$\\
Label-prop & -- & $87.6{\pm}0.5$ & $86.7{\pm}0.5$ & $43.9{\pm}0.7$ & $\underline{77.3{\pm}0.3}$ & -- & $42.3{\pm}0.9$\\
HEALNet & -- & $\underline{89.8{\pm}0.5}$ & $\underline{89.9{\pm}0.5}$ & $64.9{\pm}0.5$ & $48.1{\pm}0.6$ & -- & $\underline{87.8{\pm}0.3}$\\
\midrule
\textbf{\textsc{Elpac} (ours)} & $\mathbf{1.07{\pm}0.00}$ & $\mathbf{95.4{\pm}0.2}$ & $\mathbf{96.0{\pm}0.1}$ & $\mathbf{96.2{\pm}0.0}$ & $\mathbf{81.0{\pm}2.3}$ & $\mathbf{0.91{\pm}0.00}$ & $\mathbf{97.9{\pm}0.1}$\\
\bottomrule
\end{tabularx}
\end{table}

\begin{figure}
\centering
\includegraphics[width=\linewidth]{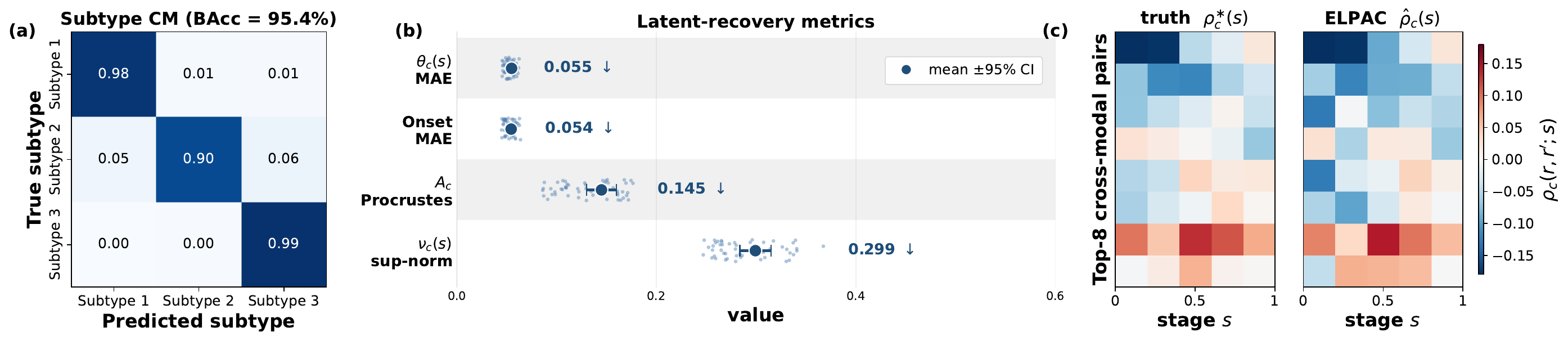}
\caption{\textbf{Synthetic latent recovery: the inferred posterior recovers subtype assignment, latent structures, and stage-varying cross-modal correlations.} \textbf{(a)} Subtype confusion matrix. \textbf{(b)} Latent-recovery metrics over 100 replicates. \textbf{(c)} Representative cross-modal correlations.}
\label{fig:synth}
\end{figure}

\begin{table}[tp]
\centering
\scriptsize
\caption{\textbf{Synthetic ablation: both constructions are necessary, removing either degrades a different recovery axis.} $N=1,000$ over 100 replicate cohorts.}
\label{tab:abl_synth}
\setlength{\tabcolsep}{4pt}
\begin{tabularx}{\linewidth}{X c c c c c c}
\toprule
Model & Held-out NLL\,$\downarrow$ & BAcc\,$\uparrow$ & Stage-score $\rho$\,$\uparrow$ & Masked NLL\,$\downarrow$ & Stage-binned PPC\,$\uparrow$ & $|$Cog $r|$\,$\uparrow$\\
\midrule
\textsc{Elpac}, no-anchor & $22.07{\pm}0.05$ & $48.9{\pm}0.6$ & $92.4{\pm}0.1$ & $1.109{\pm}0.003$ & $\underline{81.4{\pm}0.5}$ & $76.1{\pm}1.1$\\
Diagonal covariance & $23.97{\pm}0.05$ & $68.7{\pm}0.5$ & $95.7{\pm}0.0$ & $1.289{\pm}0.002$ & -- & $96.1{\pm}0.1$\\
Static-factor covariance & $\underline{19.38{\pm}0.05}$ & $\mathbf{95.5{\pm}0.2}$ & $\mathbf{97.4{\pm}0.0}$ & $\underline{0.912{\pm}0.002}$ & $78.7{\pm}0.5$ & $\mathbf{98.7{\pm}0.1}$\\
\midrule
\textbf{\textsc{Elpac} (full)} & $\mathbf{19.27{\pm}0.05}$ & $\underline{95.4{\pm}0.2}$ & $\underline{96.2{\pm}0.0}$ & $\mathbf{0.910{\pm}0.002}$ & $\mathbf{82.8{\pm}0.4}$ & $\underline{97.9{\pm}0.1}$\\
\bottomrule
\end{tabularx}
\end{table}

\subsection{Synthetic cohorts: joint latent-structure recovery and construction ablation}\label{sec:synth}
\textbf{Setup.} We generate synthetic data from the model in \cref{sec:method} using $C{=}3$ subtypes, $K{=}3$ latent coordination factors, $J{=}8$ trajectory shells, and $R{=}20$ biomarkers split across four modalities. Each replicate has $10\%$ endpoint anchors per subtype, $10\%$ healthy-reference controls, and $60\%$ unlabeled subjects. Observation masks are drawn independently per modality to match the sparse multimodal observation pattern in the real cohort. For each replicate, endpoint anchors are sampled with $\mathrm{logit}(S_i)\sim\mathcal{N}(\mathrm{logit}(s_A),\epsilon_S^2)$ for each subtype, healthy-reference controls with $\mathrm{logit}(S_i)\sim\mathcal{N}(\mathrm{logit}(s_H),\epsilon_S^2)$, and the remaining unlabeled subjects with $(c_i,S_i)$ drawn from \cref{eq:subtypestageprior}. We generate $100$ independent replicates at $N=1{,}000$, $2{,}000$, and $5{,}000$ subjects, partition each into a stratified $80/20$ train--test split, and evaluate predictive performance on held-out test subjects.

\textbf{Latent-structure recovery.}
\textsc{Elpac} jointly recovers subtype assignment, latent stage, mean trajectories, and stage-varying coordination, leading on every metric tied to the latent stage axis (\cref{tab:synth}). \cref{fig:synth} provides a closer look: the subtype posterior cleanly separates the unlabeled population, the recovered mean trajectories and onset locations are close to the truth, and the stage-varying cross-modal correlations track the truth pattern across stages.

\textbf{Ablation and sensitivity.} 
We evaluate the roles of endpoint anchoring and stage-varying coordination (\cref{tab:abl_synth}). Removing endpoint anchors damages subtype recovery (BAcc $-46.5$ pp), cognition alignment ($-21.8$ pp), and held-out likelihood ($+2.8$ nats), while leaving stage Spearman $\rho$ within $4$ pp of the full model: boundary supervision carries the subtype signal and orients the latent stage axis.
Replacing the factor covariance by a diagonal covariance damages held-out likelihood ($+4.7$ nats), masked reconstruction ($+0.38$ nats), and subtype recovery (BAcc $-26.7$ pp): residual covariance carries both density-modeling and subtype-discriminative information. The full model also leads stage-binned posterior predictive correlation (PPC) over the static-factor variant, evidence that the stage-varying activations carry information beyond a single stationary covariance.
Sensitivity analyses in \cref{app:sensitivity} show the robustness across the factor-module hyperparameters and sample-sizes. Sweeping each of the hyperparameters one at a time leaves held-out NLL within the $\pm 1.96\,\mathrm{SE}$ band (\cref{fig:appH_sensitivity}, $50$ paired cohorts/cell); scaling $N$ from $1{,}000$ to $5{,}000$ leaves all four recovery metrics within replicate-cohort SE of the $N{=}1{,}000$ values in \cref{tab:synth} (\cref{fig:appH_sample_size_100}).

\subsection{Harmonized AD cohort: stage-resolved coordination and clinical alignment}\label{sec:real_results}
\textbf{Cohort and protocol.} 
We evaluate \textsc{Elpac} on a harmonized cross-sectional Alzheimer's disease (AD) cohort of $N{=}7{,}501$ subjects pooled from ADNI~\cite{ADNI} and NACC~\cite{NACC} with SCAN-standardized imaging~\cite{SCAN} (\cameraready{NACC} March~2026 data freeze). Subjects are retained whenever any of the four modalities (amyloid PET, tau PET, structural MRI, plasma biomarkers) is observed, with the per-modality availability ranging from $13.3\%$ to $73.0\%$. The cohort comprises $641$ endpoint anchors, $219$ healthy-reference startpoints, and $6{,}641$ unlabeled subjects. The endpoint anchors split into early-onset AD (EOAD, $n{=}68$), late-onset AD (LOAD, $n{=}377$), and healthy aging (HA, $n{=}196$). Details of dataset characteristics are summarized in \cref{tab:cohort}. Performance is evaluated over 20 random $80/20$ train--test splits, each stratified by anchor stratum (per-subtype anchors, healthy reference, unlabeled). 

\begin{figure}
\centering
\includegraphics[width=\linewidth]{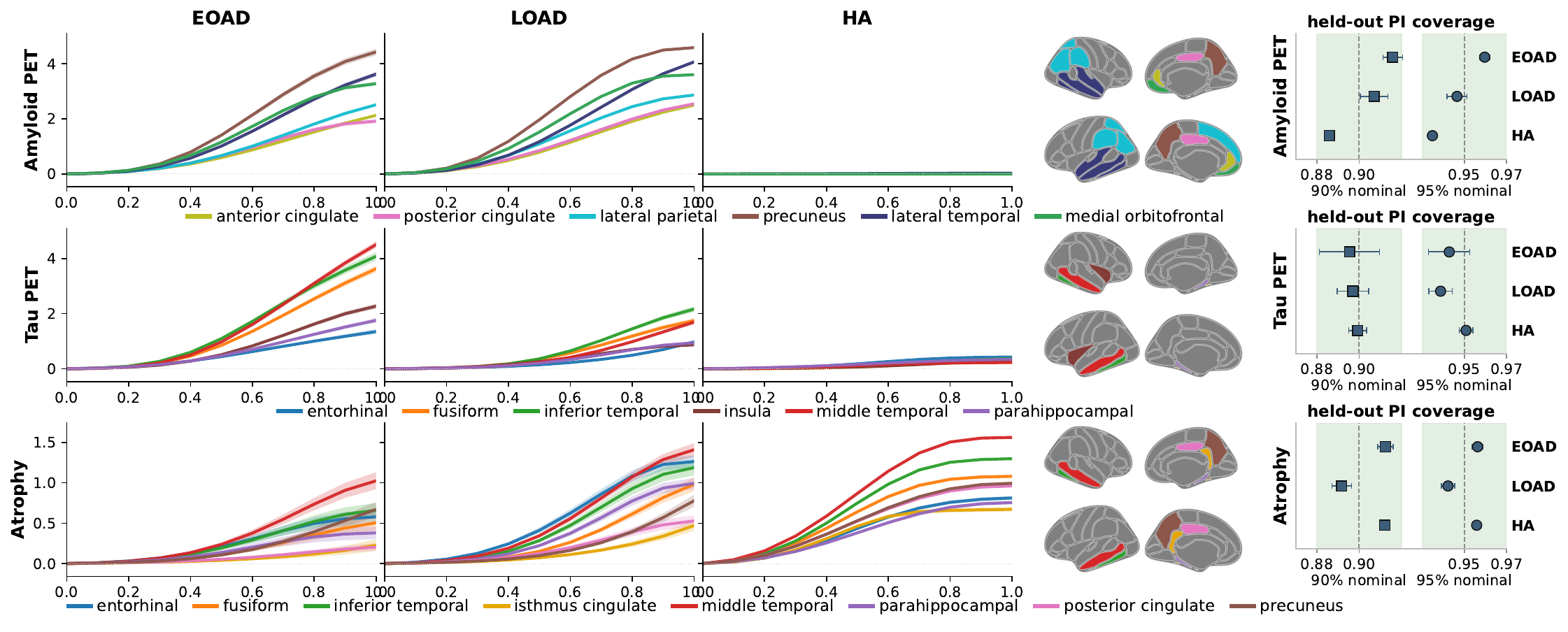}
\caption{\textbf{Recovered subtype trajectories $\bm{\theta}_c(s)$ and held-out predictive-interval (PI) coverage of biomarker observations.} ROIs map to Desikan--Killiany (DK) atlas.}
\label{fig:stage}
\end{figure}

\textbf{Generative fit, subtype and stage recovery, and calibrated prediction intervals.} 
\textsc{Elpac} leads generative fit, reconstruction, stage correlation with MMSE, and Macro-F1 in \cref{tab:stage}, validating the subtype and stage-axis recovery. 
HEALNet, paired with a supervised classification head trained on the endpoint anchors, leads diagnostic AUCs by ${\le}2.3$ pp over \textsc{Elpac}; \textsc{Elpac} additionally returns a coherent posterior over subtype, stage, and trajectories, with per-subtype, per-modality PI coverage tracking the $90\%$ and $95\%$ nominal targets (\cref{fig:stage}, column 5). 

\textbf{Ablation.} 
The AD ablation reproduces the synthetic per-axis attribution at smaller deltas (\cref{tab:abl_real}). Endpoint anchoring lifts subtype recovery (BAcc $+4.3$ pp) and both diagnostic AUCs (${+}1.7$ and $+1.9$ pp). The diagonal-covariance variant widens the held-out and masked NLL gaps by $+1.8$ nats each relative to the full model, and the static-factor variant drops BAcc by $-9.8$ pp ($58.7$ vs.\ $68.5$ full), supporting stage-varying activations as the source of subtype-discriminative residual structure. The full model leads held-out NLL, subtype recovery, both diagnostic AUCs, and masked reconstruction. 

\begin{table}
\centering
\scriptsize
\caption{\textbf{AD cohort benchmark: \textsc{Elpac} leads generative fit, biomarker reconstruction, MMSE-aligned staging, and subtype Macro-F1, while label-consuming baselines lead diagnostic classification.} Mean $\pm$ SE across $20$ held-out splits.}
\label{tab:stage}
\setlength{\tabcolsep}{3pt}
\begin{tabularx}{\linewidth}{X c c c c c c c}
\toprule
Method & NLL/cell\,$\downarrow$ & RMSE\,$\downarrow$ & $|\rho|$ MMSE\,$\uparrow$ & Macro-F1\,$\uparrow$ & BAcc\,$\uparrow$ & AUC AD/(CN{+}MCI)\,$\uparrow$ & AUC AD/MCI\,$\uparrow$\\
\midrule
SuStaIn & -- & -- & $35.6{\pm}1.0$ & $26.9{\pm}1.0$ & $36.6{\pm}0.5$ & $79.4{\pm}0.5$ & $73.0{\pm}0.5$\\
DEBM & -- & -- & $40.3{\pm}1.1$ & -- & -- & $75.2{\pm}0.7$ & $70.4{\pm}0.6$\\
Static-FA & -- & -- & $40.3{\pm}1.1$ & $28.7{\pm}1.4$ & $41.7{\pm}0.4$ & $79.7{\pm}0.5$ & $73.3{\pm}0.5$\\
MoPoE-VAE & $\underline{1.96{\pm}0.02}$ & $\underline{3.12{\pm}0.10}$ & $43.7{\pm}0.2$ & $38.7{\pm}0.2$ & $47.8{\pm}0.2$ & $82.5{\pm}0.3$ & $75.0{\pm}0.2$\\
XGBoost & -- & -- & $36.4{\pm}0.9$ & $63.4{\pm}0.3$ & $\underline{70.9{\pm}0.5}$ & $77.4{\pm}0.5$ & $71.5{\pm}0.4$\\
Label-prop & -- & -- & $42.0{\pm}0.8$ & $\underline{65.4{\pm}0.1}$ & $\mathbf{74.6{\pm}0.1}$ & $75.9{\pm}0.5$ & $70.3{\pm}0.4$\\
HEALNet & -- & -- & $\underline{49.2{\pm}0.4}$ & $50.4{\pm}0.4$ & $62.1{\pm}0.4$ & $\mathbf{87.8{\pm}0.1}$ & $\mathbf{81.2{\pm}0.2}$\\
\midrule
\textbf{\textsc{Elpac} (ours)} & $\mathbf{1.35{\pm}0.00}$ & $\mathbf{2.09{\pm}0.01}$ & $\mathbf{50.1{\pm}0.4}$ & $\mathbf{66.9{\pm}1.3}$ & $68.5{\pm}1.4$ & $\underline{86.4{\pm}0.2}$ & $\underline{78.9{\pm}0.5}$\\
\bottomrule
\end{tabularx}
\end{table}

\begin{table}
\centering
\scriptsize
\caption{\textbf{AD cohort ablation: both constructions contribute distinct improvements.} Mean $\pm$ SE across $20$ held-out splits.}
\label{tab:abl_real}
\setlength{\tabcolsep}{4pt}
\begin{tabularx}{\linewidth}{X c c c c c c}
\toprule
Model & Held-out NLL\,$\downarrow$ & BAcc\,$\uparrow$ & AUC AD/(CN{+}MCI)\,$\uparrow$ & AUC AD/MCI\,$\uparrow$ & Masked NLL\,$\downarrow$ & $|$Cog $r|$\,$\uparrow$\\
\midrule
\textsc{Elpac}, no-anchor & $\underline{11.19{\pm}0.03}$ & $64.2{\pm}1.5$ & $84.7{\pm}0.3$ & $\underline{77.0{\pm}0.6}$ & $7.38{\pm}0.05$ & $\mathbf{59.2{\pm}0.5}$\\
Diagonal covariance & $12.83{\pm}0.05$ & $\underline{66.5{\pm}1.5}$ & $\underline{84.9{\pm}0.3}$ & $76.3{\pm}0.5$ & $8.98{\pm}0.04$ & $57.0{\pm}0.5$\\
Static-factor covariance & $11.85{\pm}0.04$ & $58.7{\pm}1.6$ & $84.0{\pm}0.3$ & $74.0{\pm}0.6$ & $\underline{7.25{\pm}0.05}$ & $50.2{\pm}0.6$\\
\midrule
\textbf{\textsc{Elpac} (full)} & $\mathbf{11.01{\pm}0.03}$ & $\mathbf{68.5{\pm}1.4}$ & $\mathbf{86.4{\pm}0.2}$ & $\mathbf{78.9{\pm}0.5}$ & $\mathbf{7.22{\pm}0.05}$ & $\underline{59.0{\pm}0.5}$\\
\bottomrule
\end{tabularx}
\end{table}

\begin{figure}
\centering
\includegraphics[width=\linewidth]{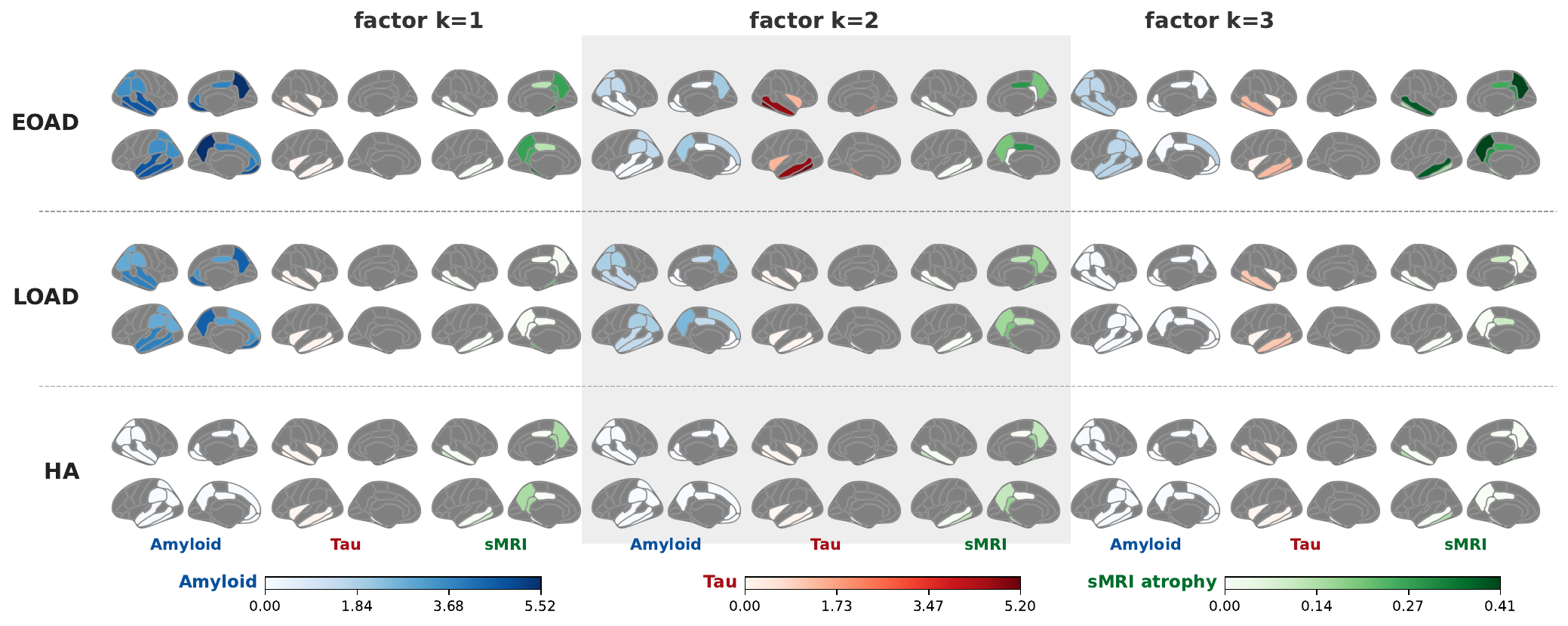}
\caption{\textbf{Coordination factors decompose pathology into subtype-specific within- and across-modality coalitions.} DK maps of $|\bm{A}_c|$ ordered by decreasing posterior loading magnitude.}
\label{fig:coord}
\end{figure}

\textbf{Recovered subtype trajectories and multimodal coordination programs.} 
The mean biomarker trajectories track subtype-specific dominant pathologies (\cref{fig:stage}): amyloid and tau in EOAD, amyloid in LOAD with structural change following and modest tau elevation, and a structural-only signal in HA that remains below the pathology threshold on amyloid and tau. 
The coordination programs differ qualitatively across subtypes (\cref{fig:coord}).
EOAD's leading axes are cross-modal A/T/N coalitions, including one axis that ties medial-temporal tau to lateral-temporal atrophy. LOAD's leading axis is amyloid-dominant within-modality, with cross-modal A/N and T/N structure on later axes. HA's leading axes are atrophy-dominant within-modality. 
The stage-varying variance share confirms the EOAD concurrent-multimodal versus LOAD sequential-amyloid-then-tau pattern (\cref{fig:appH_factor_multimodal}).

\cameraready{\textbf{Robustness to anchor corruption.} Across the four targeted corruptions, \textsc{Elpac} retains stage ordering and subtype separation: we corrupted anchors---subtype mislabeling, the true stage redrawn at random, an inverted group-mean direction, and a localized non-monotone bump---and refit at each level; the mislabeling arms are mirrored on the real cohort (\cref{app:anchor_corruption}). Anchor mislabeling degrades subtype and stage recovery smoothly as the labels get noisier (at 20\% uniform mislabeling, held-out synthetic subtype accuracy and stage $\rho$ fall by 3.4 and 5.7 pp and real-cohort BAcc by 7.0 pp); the stage redraw affects staging while leaving subtype recovery unchanged; and the two mean-shape perturbations have the smallest stage-ordering effects. On the real cohort, held-out likelihood stays well inside the $0.84$-nat full-vs-static-factor margin of \cref{tab:abl_real}.}

\textbf{Cross-cohort robustness and longitudinal consistency.} 
Leave-one-cohort-out folds preserve the stage axis and subtype mapping (\cref{app:loco}), indicating cross-cohort generalization. On subjects with longitudinal follow-up visits, \textsc{Elpac}'s cross-sectional stage estimates \cameraready{advance on average with subsequent visits}, and the same model attains the highest cognition recovery and three-year MCI-to-AD conversion AUC among the baselines (\cref{app:longitudinal}). These support the anchored stage axis as a transferable severity coordinate.

\section{Discussion}
\textsc{Elpac} resolves two coupled problems in latent-progression modeling: locating each subject on a disease axis whose endpoints are reliably labeled but whose origin is not, and capturing how multimodal biomarker coalitions reorganize along that axis. Endpoint anchoring inverts the supervision direction for the first; the stage-varying factor decomposition handles the second, with the inferred stage axis supplying the auxiliary variation that conditionally identifies its loadings and activations.

\textbf{Limitations.} \cameraready{The stage axis is pseudo-temporal, recovered from cross-sectional snapshots, so the coordination factors describe co-variation along it rather than mechanism.} The dependency layer in \cref{eq:latentfactor} models second-order multimodal coordination; higher-order joint dependencies are a future extension via copula or normalizing-flow residuals. Theorem~\ref{thm:identifiability} identifies the dynamic factor block under the conditions of \cref{app:thm_proof}; the full-model proposition (\cref{app:ident_warp_prop}) further invokes a Gaussian mixing-family hypothesis, with broader mixing families a separate theoretical direction. The amortized variational scheme produces cohort-scale joint inference and single-pass test-time deployment at the per-evaluation cost stated in \cref{sec:inference}; richer guide families (e.g., normalizing-flow posteriors) remain an inference-side extension.

\begin{ack}
{\ifcameraready\color{blue}\fi\textbf{Funding.} This work was partially funded by awards R01AG081413, R01EB034720, and R01AG068191 from the NIH. The content is solely the responsibility of the authors and does not necessarily represent the official views of the NIH.

\textbf{Competing interests.} The authors declare no competing interests.

\textbf{Data.} Data collection and sharing for the Alzheimer's Disease Neuroimaging Initiative (ADNI) is funded by the National Institute on Aging (National Institutes of Health Grant U19 AG024904). The grantee organization is the Northern California Institute for Research and Education. In the past, ADNI has also received funding from the National Institute of Biomedical Imaging and Bioengineering, the Canadian Institutes of Health Research, and private sector contributions through the Foundation for the National Institutes of Health (FNIH) including generous contributions from the following: AbbVie, Alzheimer's Association; Alzheimer's Drug Discovery Foundation; Araclon Biotech; BioClinica, Inc.; Biogen; Bristol-Myers Squibb Company; CereSpir, Inc.; Cogstate; Eisai Inc.; Elan Pharmaceuticals, Inc.; Eli Lilly and Company; EuroImmun; F. Hoffmann-La Roche Ltd and its affiliated company Genentech, Inc.; Fujirebio; GE Healthcare; IXICO Ltd.; Janssen Alzheimer Immunotherapy Research \& Development, LLC.; Johnson \& Johnson Pharmaceutical Research \&Development LLC.; Lumosity; Lundbeck; Merck \& Co., Inc.; Meso Scale Diagnostics, LLC.; NeuroRx Research; Neurotrack Technologies; Novartis Pharmaceuticals Corporation; Pfizer Inc.; Piramal Imaging; Servier; Takeda Pharmaceutical Company; and Transition Therapeutics.

The NACC database is funded by NIA/NIH Grant U24 AG072122. NACC data are contributed by the NIA-funded ADRCs: P30 AG062429 (PI James Brewer, MD, PhD), P30 AG066468 (PI Oscar Lopez, MD), P30 AG062421 (PI Teresa Gomez-Isla, MD), P30 AG066509 (PI Thomas Grabowski, MD), P30 AG066514 (PI Mary Sano, PhD), P30 AG066530 (PI Helena Chui, MD, Arthur Toga, PhD), P30 AG066507 (PI Marilyn Albert, PhD), P30 AG066444 (PI David Holtzman, MD), P30 AG066518 (PIs Lisa Silbert, MD, Kevin Duff, PhD), P30 AG066512 (PI Thomas Wisniewski, MD), P30 AG066462 (PI Scott Small, MD), P30 AG072979 (PI David Wolk, MD), P30 AG072972 (PIs Charles DeCarli, MD, Rachel Whitmer, PhD), P30 AG072976 (PI Andrew Saykin, PsyD), P30 AG072975 (PI Julie Schneider, MD, MS), P30 AG072978 (PI Ann McKee, MD), P30 AG072977 (PI Robert Vassar, PhD), P30 AG066519 (PI Joshua Grill, PhD), P30 AG062677 (PIs Brad Boeve, MD, Ronald Petersen, MD, PhD), P30 AG079280 (PI Jessica Langbaum, PhD), P30 AG062422 (PI Gil Rabinovici, MD), P30 AG066511 (PI Allan Levey, MD, PhD), P30 AG072946 (PI Linda Van Eldik, PhD), P30 AG062715 (PI Sanjay Asthana, MD, FRCP), P30 AG072973 (PI Russell Swerdlow, MD), P30 AG066506 (PIs Glenn Smith, PhD, ABPP, David Lowenstein, PhD, Ranjan Duara, MD), P30 AG066508 (PIs Stephen Strittmatter, MD, PhD, Christopher Van Dyck, MD), P30 AG066515 (PI Victor Henderson, MD, MS), P30 AG072947 (PI Suzanne Craft, PhD), P30 AG072931 (PI Henry Paulson, MD, PhD), P30 AG066546 (PIs Sudha Seshadri, MD, Gladys Maestre, MD, PhD), P30 AG086401 (PI Erik Roberson, MD, PhD), P30 AG086404 (PI Gary Rosenberg, MD), P30 AG086403 (PI Angela Jefferson, PhD), P30 AG072958 (PIs Heather Whitson, MD, Gwenn Garden, MD, PhD), P30 AG072959 (PI Jagan Pillai, MD, PhD), P30 AG092752 (Ihab Hajjar, MD, MS).

SCAN is a multi-institutional project that was funded as a U24 grant (AG067418) by the National Institute on Aging in May 2020. Data collected by SCAN and shared by NACC are contributed by the NIA-funded ADRCs as follows: Arizona Alzheimer's Center - P30 AG072980 (PI: Eric Reiman, MD); R01 AG069453 (PI: Eric Reiman (contact), MD); P30 AG019610 (PI: Eric Reiman, MD); and the State of Arizona which provided additional funding supporting our center; Boston University - P30 AG013846 (PI Neil Kowall MD); Cleveland ADRC - P30 AG062428 (James Leverenz, MD); Cleveland Clinic, Las Vegas -- P20AG068053; Columbia - P50 AG008702 (PI Scott Small MD); Duke/UNC ADRC -- P30 AG072958; Emory University - P30AG066511 (PI Levey Allan, MD, PhD); Indiana University - R01 AG19771 (PI Andrew Saykin, PsyD); P30 AG10133 (PI Andrew Saykin, PsyD); P30 AG072976 (PI Andrew Saykin, PsyD); R01 AG061788 (PI Shannon Risacher, PhD); R01 AG053993 (PI Yu-Chien Wu, MD, PhD); U01 AG057195 (PI Liana Apostolova, MD); U19 AG063911 (PI Bradley Boeve, MD); and the Indiana University Department of Radiology and Imaging Sciences; Johns Hopkins - P30 AG066507 (PI Marilyn Albert, Phd.); Mayo Clinic - P50 AG016574 (PI Ronald Petersen MD PhD); Mount Sinai - P30 AG066514 (PI Mary Sano, PhD); R01 AG054110 (PI Trey Hedden, PhD); R01 AG053509 (PI Trey Hedden, PhD); New York University - P30AG066512-01S2 (PI Thomas Wisniewski, MD); R01AG056031 (PI Ricardo Osorio, MD); R01AG056531 (PIs Ricardo Osorio, MD; Girardin Jean-Louis, PhD); Northwestern University - P30 AG013854 (PI Robert Vassar PhD); R01 AG045571 (PI Emily Rogalski, PhD); R56 AG045571, (PI Emily Rogalski, PhD); R01 AG067781, (PI Emily Rogalski, PhD); U19 AG073153, (PI Emily Rogalski, PhD); R01 DC008552, (M.-Marsel Mesulam, MD); R01 AG077444, (PIs M.-Marsel Mesulam, MD, Emily Rogalski, PhD); R01 NS075075 (PI Emily Rogalski, PhD); R01 AG056258 (PI Emily Rogalski, PhD); Oregon Health and Science University - P30 AG066518 (PIs Lisa Silbert, MD, Kevin Duff, PhD); R56 AG074321 (PI Jeffrey Kaye, MD); Rush University - P30 AG010161 (PI David Bennett MD); Stanford -- P30AG066515; P50 AG047366 (PI Victor Henderson MD MS); University of Alabama, Birmingham -- P20; University of California, Davis - P30 AG10129 (PI Charles DeCarli, MD); P30 AG072972 (PI Charles DeCarli, MD); University of California, Irvine - P50 AG016573 (PI Frank LaFerla PhD); University of California, San Diego - P30AG062429 (PI James Brewer, MD, PhD); University of California, San Francisco - P30 AG062422 (Rabinovici, Gil D., MD); University of Kansas - P30 AG035982 (Russell Swerdlow, MD); University of Kentucky - P30 AG028283-15S1 (PIs Linda Van Eldik, PhD and Brian Gold, PhD); University of Michigan ADRC - P30AG053760 (PI Henry Paulson, MD, PhD) P30AG072931 (PI Henry Paulson, MD, PhD) Cure Alzheimer's Fund 200775 - (PI Henry Paulson, MD, PhD) U19 NS120384 (PI Charles DeCarli, MD, University of Michigan Site PI Henry Paulson, MD, PhD) R01 AG068338 (MPI Bruno Giordani, PhD, Carol Persad, PhD, Yi Murphey, PhD) S10OD026738-01 (PI Douglas Noll, PhD) R01 AG058724 (PI Benjamin Hampstead, PhD) R35 AG072262 (PI Benjamin Hampstead, PhD) W81XWH2110743 (PI Benjamin Hampstead, PhD) R01 AG073235 (PI Nancy Chiaravalloti, University of Michigan Site PI Benjamin Hampstead, PhD) 1I01RX001534 (PI Benjamin Hampstead, PhD) IRX001381 (PI Benjamin Hampstead, PhD); University of New Mexico - P20 AG068077 (Gary Rosenberg, MD); University of Pennsylvania - State of PA project 2019NF4100087335 (PI David Wolk, MD); Rooney Family Research Fund (PI David Wolk, MD); R01 AG055005 (PI David Wolk, MD); University of Pittsburgh - P50 AG005133 (PI Oscar Lopez MD); University of Southern California - P50 AG005142 (PI Helena Chui MD); University of Washington - P50 AG005136 (PI Thomas Grabowski MD); University of Wisconsin - P50 AG033514 (PI Sanjay Asthana MD FRCP); Vanderbilt University -- P20 AG068082; Wake Forest - P30AG072947 (PI Suzanne Craft, PhD); Washington University, St. Louis - P01 AG03991 (PI John Morris MD); P01 AG026276 (PI John Morris MD); P20 MH071616 (PI Dan Marcus); P30 AG066444 (PI John Morris MD); P30 NS098577 (PI Dan Marcus); R01 AG021910 (PI Randy Buckner); R01 AG043434 (PI Catherine Roe); R01 EB009352 (PI Dan Marcus); UL1 TR000448 (PI Brad Evanoff); U24 RR021382 (PI Bruce Rosen); Avid Radiopharmaceuticals / Eli Lilly; Yale - P50 AG047270 (PI Stephen Strittmatter MD PhD); R01AG052560 (MPI: Christopher van Dyck, MD; Richard Carson, PhD); R01AG062276 (PI: Christopher van Dyck, MD); 1Florida - P30AG066506-03 (PI Glenn Smith, PhD); P50 AG047266 (PI Todd Golde MD PhD)}
\end{ack}

\bibliographystyle{unsrtnat}
\bibliography{paper_neurips2026}

\begin{thebibliography}{50}
\providecommand{\natexlab}[1]{#1}
\providecommand{\url}[1]{\texttt{#1}}
\expandafter\ifx\csname urlstyle\endcsname\relax
  \providecommand{\doi}[1]{doi: #1}\else
  \providecommand{\doi}{doi: \begingroup \urlstyle{rm}\Url}\fi

\bibitem[Young et~al.(2024)Young, Oxtoby, Garbarino, Fox, Barkhof, Schott, and
  Alexander]{Young2024}
Alexandra~L. Young, Neil~P. Oxtoby, Sara Garbarino, Nick~C. Fox, Frederik
  Barkhof, Jonathan~M. Schott, and Daniel~C. Alexander.
\newblock Data-driven modelling of neurodegenerative disease progression:
  thinking outside the black box.
\newblock \emph{Nature Reviews Neuroscience}, 25\penalty0 (2):\penalty0
  111--130, 2024.
\newblock ISSN 1471-003X.

\bibitem[Jack et~al.(2024)Jack, Andrews, Beach, Buracchio, Dunn, Graf, Hansson,
  Ho, Jagust, McDade, Molinuevo, Okonkwo, Pani, Rafii, Scheltens, Siemers,
  Snyder, Sperling, Teunissen, and Carrillo]{Jack2024}
Jr. Jack, Clifford~R., J.~Scott Andrews, Thomas~G. Beach, Teresa Buracchio,
  Billy Dunn, Ana Graf, Oskar Hansson, Carole Ho, William Jagust, Eric McDade,
  José~Luis Molinuevo, Ozioma~C. Okonkwo, Luca Pani, Michael~S. Rafii, Philip
  Scheltens, Eric Siemers, Heather~M. Snyder, Reisa Sperling, Charlotte~E.
  Teunissen, and Maria~C. Carrillo.
\newblock Revised criteria for diagnosis and staging of {Alzheimer}'s disease:
  {Alzheimer}'s association workgroup.
\newblock \emph{Alzheimer's \& Dementia}, 20\penalty0 (8):\penalty0 5143--5169,
  2024.
\newblock ISSN 1552-5260.

\bibitem[Salvadó et~al.(2024)Salvadó, Horie, Barthélemy, Vogel,
  Pichet~Binette, Chen, Aschenbrenner, Gordon, Benzinger, Holtzman, Morris,
  Palmqvist, Stomrud, Janelidze, Ossenkoppele, Schindler, Bateman, and
  Hansson]{Salvado2024}
Gemma Salvadó, Kanta Horie, Nicolas~R. Barthélemy, Jacob~W. Vogel, Alexa
  Pichet~Binette, Charles~D. Chen, Andrew~J. Aschenbrenner, Brian~A. Gordon,
  Tammie L.~S. Benzinger, David~M. Holtzman, John~C. Morris, Sebastian
  Palmqvist, Erik Stomrud, Shorena Janelidze, Rik Ossenkoppele, Suzanne~E.
  Schindler, Randall~J. Bateman, and Oskar Hansson.
\newblock Disease staging of {Alzheimer}'s disease using a {CSF}-based
  biomarker model.
\newblock \emph{Nature Aging}, 4\penalty0 (5):\penalty0 694--708, 2024.
\newblock ISSN 2662-8465.

\bibitem[Palmqvist et~al.(2025)Palmqvist, Warmenhoven, Anastasi, Pilotto,
  Janelidze, Tideman, Stomrud, Mattsson-Carlgren, Smith, Ossenkoppele, Tan,
  Dittrich, Skoog, Zetterberg, Quaresima, Tolassi, Höglund, Brugnoni,
  Puig-Pijoan, Fernández-Lebrero, Contador, Padovani, Monane, Verghese,
  Braunstein, Kern, Blennow, Ashton, Suárez-Calvet, and
  Hansson]{Palmqvist2025}
Sebastian Palmqvist, Noëlle Warmenhoven, Federica Anastasi, Andrea Pilotto,
  Shorena Janelidze, Pontus Tideman, Erik Stomrud, Niklas Mattsson-Carlgren,
  Ruben Smith, Rik Ossenkoppele, Kübra Tan, Anna Dittrich, Ingmar Skoog,
  Henrik Zetterberg, Virginia Quaresima, Chiara Tolassi, Kina Höglund, Duilio
  Brugnoni, Albert Puig-Pijoan, Aida Fernández-Lebrero, José Contador,
  Alessandro Padovani, Mark Monane, Philip~B. Verghese, Joel~B. Braunstein,
  Silke Kern, Kaj Blennow, Nicholas~J. Ashton, Marc Suárez-Calvet, and Oskar
  Hansson.
\newblock Plasma phospho-tau217 for {A}lzheimer's disease diagnosis in primary
  and secondary care using a fully automated platform.
\newblock \emph{Nature Medicine}, 31\penalty0 (6):\penalty0 2036--2043, 2025.
\newblock ISSN 1078-8956.

\bibitem[Fonteijn et~al.(2012)Fonteijn, Modat, Clarkson, Barnes, Lehmann,
  Hobbs, Scahill, Tabrizi, Ourselin, Fox, and Alexander]{Fonteijn2012}
Hubert~M. Fonteijn, Marc Modat, Matthew~J. Clarkson, Josephine Barnes, Manja
  Lehmann, Nicola~Z. Hobbs, Rachael~I. Scahill, Sarah~J. Tabrizi, Sebastien
  Ourselin, Nick~C. Fox, and Daniel~C. Alexander.
\newblock An event-based model for disease progression and its application in
  familial {Alzheimer}'s disease and {Huntington}'s disease.
\newblock \emph{NeuroImage}, 60\penalty0 (3):\penalty0 1880--1889, 2012.
\newblock ISSN 1053-8119.

\bibitem[Young et~al.(2018)Young, Marinescu, Oxtoby, Bocchetta, Yong, Firth,
  Cash, Thomas, Dick, Cardoso, van Swieten, Borroni, Galimberti, Masellis,
  Tartaglia, Rowe, Graff, Tagliavini, Frisoni, Laforce, Finger, de~Mendon{\c
  c}a, Sorbi, Warren, {The Genetic FTD Initiative (GENFI)}, and {The
  Alzheimer's Disease Neuroimaging Initiative (ADNI)}]{Young2018}
Alexandra~L. Young, Razvan~V. Marinescu, Neil~P. Oxtoby, Martina Bocchetta,
  Keir Yong, Nicholas~C. Firth, David~M. Cash, David~L. Thomas, Katrina~M.
  Dick, Jorge Cardoso, John van Swieten, Barbara Borroni, Daniela Galimberti,
  Mario Masellis, Maria~Carmela Tartaglia, James~B. Rowe, Caroline Graff,
  Fabrizio Tagliavini, Giovanni~B. Frisoni, Robert Laforce, Jr., Elizabeth
  Finger, Alexandre de~Mendon{\c c}a, Sandro Sorbi, Jason~D. Warren, {The
  Genetic FTD Initiative (GENFI)}, and {The Alzheimer's Disease Neuroimaging
  Initiative (ADNI)}.
\newblock Uncovering the heterogeneity and temporal complexity of
  neurodegenerative diseases with subtype and stage inference.
\newblock \emph{Nature Communications}, 9\penalty0 (1):\penalty0 4273, 2018.
\newblock ISSN 2041-1723.

\bibitem[Venkatraghavan et~al.(2019)Venkatraghavan, Bron, Niessen, and
  Klein]{Venkatraghavan2019}
Vikram Venkatraghavan, Esther~E. Bron, Wiro~J. Niessen, and Stefan Klein.
\newblock Disease progression timeline estimation for {Alzheimer's} disease
  using discriminative event based modeling.
\newblock \emph{NeuroImage}, 186:\penalty0 518--532, 2019.
\newblock ISSN 1053-8119.

\bibitem[Wijeratne and Alexander(2024)]{Wijeratne2024}
Peter~A. Wijeratne and Daniel~C. Alexander.
\newblock Unscrambling disease progression at scale: fast inference of event
  permutations with optimal transport.
\newblock In \emph{Advances in Neural Information Processing Systems 37}, pages
  63316--63341. Neural Information Processing Systems Foundation, 2024.

\bibitem[Koval et~al.(2021)Koval, Bône, Louis, Lartigue, Bottani, Marcoux,
  Samper-González, Burgos, Charlier, Bertrand, Epelbaum, Colliot,
  Allassonnière, and Durrleman]{Koval2021}
Igor Koval, Alexandre Bône, Maxime Louis, Thomas Lartigue, Simona Bottani,
  Arnaud Marcoux, Jorge Samper-González, Ninon Burgos, Benjamin Charlier, Anne
  Bertrand, Stéphane Epelbaum, Olivier Colliot, Stéphanie Allassonnière, and
  Stanley Durrleman.
\newblock {AD Course Map} charts {Alzheimer}'s disease progression.
\newblock \emph{Scientific Reports}, 11\penalty0 (1):\penalty0 8020, 2021.
\newblock ISSN 2045-2322.

\bibitem[Schiratti et~al.(2017)Schiratti, Allassonnière, Colliot, and
  Durrleman]{Schiratti2017}
Jean-Baptiste Schiratti, Stéphanie Allassonnière, Olivier Colliot, and
  Stanley Durrleman.
\newblock A {Bayesian} mixed-effects model to learn trajectories of changes
  from repeated manifold-valued observations.
\newblock \emph{Journal of Machine Learning Research}, 18\penalty0
  (133):\penalty0 1--33, 2017.

\bibitem[Lespinasse et~al.(2023)Lespinasse, Dufouil, and
  Proust-Lima]{Lespinasse2023}
Jérémie Lespinasse, Carole Dufouil, and Cécile Proust-Lima.
\newblock Disease progression model anchored around clinical diagnosis in
  longitudinal cohorts: example of {A}lzheimer's disease and related dementia.
\newblock \emph{BMC Medical Research Methodology}, 23\penalty0 (1):\penalty0
  199, 2023.
\newblock ISSN 1471-2288.

\bibitem[Viani et~al.(2025)Viani, Custo, d'Angremont, Garibotto, Frisoni,
  Gutman, and Lorenzi]{Viani2025}
Alessandro Viani, Anna Custo, Emile d'Angremont, Valentina Garibotto,
  Giovanni~B. Frisoni, Boris~A. Gutman, and Marco Lorenzi.
\newblock Disease progression modeling and stratification for detecting
  sub-trajectories in the natural history of pathologies: Application to
  {A}lzheimer's disease trajectory modeling.
\newblock \emph{Imaging Neuroscience}, 3:\penalty0 IMAG.a.954, 2025.
\newblock ISSN 2837-6056.

\bibitem[Zhang et~al.(2024)Zhang, Pu, Kawamura, Loza, Bengio, Shung, and
  Tong]{Zhang2024TFM}
Xi~Zhang, Yuan Pu, Yuki Kawamura, Andrew Loza, Yoshua Bengio, Dennis~L. Shung,
  and Alexander Tong.
\newblock Trajectory flow matching with applications to clinical time series
  modelling.
\newblock In \emph{Advances in Neural Information Processing Systems 37}, pages
  107198--107224. Neural Information Processing Systems Foundation, 2024.

\bibitem[Liu et~al.(2025)Liu, Yao, Yin, Cheung, and Qin]{Liu2025DiPro}
Chen Liu, Wenfang Yao, Kejing Yin, William~K. Cheung, and Jing Qin.
\newblock Multimodal disease progression modeling via spatio-temporal
  disentanglement and multiscale alignment.
\newblock In \emph{Advances in Neural Information Processing Systems 38}.
  Neural Information Processing Systems Foundation, 2025.

\bibitem[Wilson and Ghahramani(2011)]{Wilson2011}
Andrew~Gordon Wilson and Zoubin Ghahramani.
\newblock Generalised {Wishart} processes.
\newblock In \emph{Proceedings of the Twenty-Seventh Conference on Uncertainty
  in Artificial Intelligence}, pages 736--744. AUAI Press, 2011.

\bibitem[Hoff and Niu(2012)]{HoffNiu2012}
Peter~D. Hoff and Xiaoyue Niu.
\newblock A covariance regression model.
\newblock \emph{Statistica Sinica}, 22\penalty0 (2):\penalty0 729--753, 2012.
\newblock ISSN 1017-0405.

\bibitem[Nejatbakhsh et~al.(2023)Nejatbakhsh, Garon, and
  Williams]{Nejatbakhsh2023}
Amin Nejatbakhsh, Isabel Garon, and Alex~H. Williams.
\newblock Estimating noise correlations across continuous conditions with
  {Wishart} processes.
\newblock In \emph{Advances in Neural Information Processing Systems 36}, pages
  54032--54045. Neural Information Processing Systems Foundation, 2023.

\bibitem[Tu et~al.(2024)Tu, Mahony, Moore, Bertolero, Alexander-Bloch, Gur,
  Bassett, Satterthwaite, Raznahan, and Shinohara]{Tu2024}
Danni Tu, Bridget Mahony, Tyler~M. Moore, Maxwell~A. Bertolero, Aaron~F.
  Alexander-Bloch, Ruben Gur, Dani~S. Bassett, Theodore~D. Satterthwaite, Armin
  Raznahan, and Russell~T. Shinohara.
\newblock {CoCoA}: conditional correlation models with association size.
\newblock \emph{Biostatistics}, 25\penalty0 (1):\penalty0 154--170, 2024.
\newblock ISSN 1465-4644.

\bibitem[Chen et~al.(2021)Chen, Lu, Weng, Chen, Williamson, Manz, Shady, and
  Mahmood]{Chen2021MCAT}
Richard~J. Chen, Ming~Y. Lu, Wei-Hung Weng, Tiffany~Y. Chen, Drew F.~K.
  Williamson, Trevor Manz, Maha Shady, and Faisal Mahmood.
\newblock Multimodal co-attention transformer for survival prediction in
  gigapixel whole slide images.
\newblock In \emph{Proceedings of the IEEE/CVF International Conference on
  Computer Vision}, pages 4015--4025. IEEE, 2021.

\bibitem[Chen et~al.(2022)Chen, Lu, Wang, Williamson, Rodig, Lindeman, and
  Mahmood]{Chen2022Pathomic}
Richard~J. Chen, Ming~Y. Lu, Jingwen Wang, Drew F.~K. Williamson, Scott~J.
  Rodig, Neal~I. Lindeman, and Faisal Mahmood.
\newblock Pathomic fusion: An integrated framework for fusing histopathology
  and genomic features for cancer diagnosis and prognosis.
\newblock \emph{IEEE Transactions on Medical Imaging}, 41\penalty0
  (4):\penalty0 757--770, 2022.
\newblock ISSN 0278-0062.

\bibitem[Wu and Goodman(2018)]{WuGoodman2018PoE}
Mike Wu and Noah~D. Goodman.
\newblock Multimodal generative models for scalable weakly-supervised learning.
\newblock In \emph{Advances in Neural Information Processing Systems 31}, 2018.

\bibitem[Sutter et~al.(2021)Sutter, Daunhawer, and Vogt]{Sutter2021MoPoE}
Thomas~M. Sutter, Imant Daunhawer, and Julia~E. Vogt.
\newblock Generalized multimodal {ELBO}.
\newblock In \emph{International Conference on Learning Representations
  ({ICLR})}, 2021.

\bibitem[Hemker et~al.(2024)Hemker, Simidjievski, and Jamnik]{Hemker2024}
Konstantin Hemker, Nikola Simidjievski, and Mateja Jamnik.
\newblock {HEALNet}: Multimodal fusion for heterogeneous biomedical data.
\newblock In \emph{Advances in Neural Information Processing Systems 37}, pages
  64479--64498. Neural Information Processing Systems Foundation, 2024.

\bibitem[Hyvärinen and Morioka(2016)]{HyvarinenMorioka2016}
Aapo Hyvärinen and Hiroshi Morioka.
\newblock Unsupervised feature extraction by time-contrastive learning and
  nonlinear ica.
\newblock In \emph{Advances in Neural Information Processing Systems},
  volume~29, 2016.

\bibitem[Zheng and Zhang(2023)]{ZhengZhang2023}
Yujia Zheng and Kun Zhang.
\newblock Generalizing nonlinear {ICA} beyond structural sparsity.
\newblock In \emph{Advances in Neural Information Processing Systems 36}, pages
  13326--13355. Neural Information Processing Systems Foundation, 2023.

\bibitem[von Kügelgen et~al.(2023)von Kügelgen, Besserve, Wendong, Gresele,
  Kekić, Bareinboim, Blei, and Schölkopf]{vonKugelgen2023}
Julius von Kügelgen, Michel Besserve, Liang Wendong, Luigi Gresele, Armin
  Kekić, Elias Bareinboim, David~M. Blei, and Bernhard Schölkopf.
\newblock Nonparametric identifiability of causal representations from unknown
  interventions.
\newblock In \emph{Advances in Neural Information Processing Systems 36}, pages
  48603--48638. Neural Information Processing Systems Foundation, 2023.

\bibitem[Anderson and Rubin(1956)]{AndersonRubin1956}
Theodore~W. Anderson and Herman Rubin.
\newblock Statistical inference in factor analysis.
\newblock In \emph{Proceedings of the Third Berkeley Symposium on Mathematical
  Statistics and Probability}, volume~5, pages 111--150. University of
  California Press, 1956.

\bibitem[Bhattacharya and Dunson(2011)]{BhattacharyaDunson2011}
Anirban Bhattacharya and David~B. Dunson.
\newblock Sparse {Bayesian} infinite factor models.
\newblock \emph{Biometrika}, 98\penalty0 (2):\penalty0 291--306, 2011.
\newblock ISSN 0006-3444.

\bibitem[Pati et~al.(2014)Pati, Bhattacharya, Pillai, and Dunson]{Pati2014}
Debdeep Pati, Anirban Bhattacharya, Natesh~S. Pillai, and David~B. Dunson.
\newblock Posterior contraction in sparse {Bayesian} factor models for massive
  covariance matrices.
\newblock \emph{The Annals of Statistics}, 42\penalty0 (3):\penalty0
  1102--1130, 2014.
\newblock ISSN 0090-5364.

\bibitem[Geweke and Zhou(1996)]{Geweke1996}
John Geweke and Guofu Zhou.
\newblock Measuring the pricing error of the arbitrage pricing theory.
\newblock \emph{Review of Financial Studies}, 9\penalty0 (2):\penalty0
  557--587, 1996.
\newblock ISSN 0893-9454.

\bibitem[Kingma and Welling(2014)]{KingmaWelling2014VAE}
Diederik~P. Kingma and Max Welling.
\newblock Auto-encoding variational {B}ayes.
\newblock In \emph{International Conference on Learning Representations
  ({ICLR})}, 2014.

\bibitem[Rezende et~al.(2014)Rezende, Mohamed, and
  Wierstra]{RezendeMohamed2014}
Danilo~Jimenez Rezende, Shakir Mohamed, and Daan Wierstra.
\newblock Stochastic backpropagation and approximate inference in deep
  generative models.
\newblock In \emph{International Conference on Machine Learning ({ICML})},
  2014.

\bibitem[Kingma and Ba(2015)]{KingmaBa2015Adam}
Diederik~P. Kingma and Jimmy Ba.
\newblock {Adam}: A method for stochastic optimization.
\newblock In \emph{International Conference on Learning Representations
  ({ICLR})}, 2015.

\bibitem[Henderson and Searle(1981)]{HendersonSearle1981}
Harold~V. Henderson and Shayle~R. Searle.
\newblock On deriving the inverse of a sum of matrices.
\newblock \emph{SIAM Review}, 23\penalty0 (1):\penalty0 53--60, 1981.
\newblock ISSN 0036-1445.

\bibitem[Chen and Guestrin(2016)]{Chen2016XGBoost}
Tianqi Chen and Carlos Guestrin.
\newblock {XGBoost}: A scalable tree boosting system.
\newblock In \emph{Proceedings of the 22nd ACM SIGKDD International Conference
  on Knowledge Discovery and Data Mining}, pages 785--794. ACM, 2016.

\bibitem[Zhu and Ghahramani(2002)]{Zhu2002LabelProp}
Xiaojin Zhu and Zoubin Ghahramani.
\newblock Learning from labeled and unlabeled data with label propagation.
\newblock Technical Report CMU-CALD-02-107, Carnegie Mellon University, 2002.

\bibitem[Weiner et~al.(2025)Weiner, Kanoria, Miller, Aisen, Beckett, Conti,
  Diaz, Flenniken, Green, Harvey, Jack, Jagust, Lee, Morris, Nho, Nosheny,
  Okonkwo, Perrin, Petersen, Rivera-Mindt, Saykin, Shaw, Toga, Tosun, and
  Veitch]{ADNI}
Michael~W. Weiner, Shaveta Kanoria, Melanie~J. Miller, Paul~S. Aisen, Laurel~A.
  Beckett, Catherine Conti, Adam Diaz, Derek Flenniken, Robert~C. Green,
  Danielle~J. Harvey, Clifford~R. Jack, William Jagust, Edward~B. Lee, John~C.
  Morris, Kwangsik Nho, Rachel Nosheny, Ozioma~C. Okonkwo, Richard~J. Perrin,
  Ronald~C. Petersen, Monica Rivera-Mindt, Andrew~J. Saykin, Leslie~M. Shaw,
  Arthur~W. Toga, Duygu Tosun, and Dallas~P. Veitch.
\newblock Overview of {Alzheimer's} {Disease} {Neuroimaging} {Initiative} and
  future clinical trials.
\newblock \emph{Alzheimer's \& Dementia}, 21\penalty0 (1):\penalty0 e14321,
  2025.
\newblock ISSN 1552-5260.

\bibitem[Beekly et~al.(2007)Beekly, Ramos, Lee, Deitrich, Jacka, Wu, Hubbard,
  Koepsell, Morris, Kukull, and {NIA Alzheimer's Disease Centers}]{NACC}
Duane~L. Beekly, Erin~M. Ramos, William~W. Lee, Woodrow~D. Deitrich, Mary~E.
  Jacka, Joylee Wu, Janene~L. Hubbard, Thomas~D. Koepsell, John~C. Morris,
  Walter~A. Kukull, and {NIA Alzheimer's Disease Centers}.
\newblock The {National} {Alzheimer's} {Coordinating} {Center} ({NACC})
  database: the {Uniform} {Data} {Set}.
\newblock \emph{Alzheimer Disease \& Associated Disorders}, 21\penalty0
  (3):\penalty0 249--258, 2007.
\newblock ISSN 0893-0341.

\bibitem[Cody et~al.(2025)Cody, Sokolowski, Guerra, Insel, Jagust, Harrison,
  and Mormino]{SCAN}
Karly~A. Cody, Andrzej Sokolowski, Lucah~Medina Guerra, Philip~S. Insel,
  William~J. Jagust, Theresa~M. Harrison, and Elizabeth~C. Mormino.
\newblock Comparison of amyloid {PET} acquired through standardized and
  unstandardized protocols.
\newblock \emph{Alzheimer's \& Dementia}, 21\penalty0 (9):\penalty0 e70697,
  2025.
\newblock ISSN 1552-5260.
\newblock {SCAN}: {Standardized} {Centralized} {Alzheimer's} \& {Related}
  {Dementias} {Neuroimaging}, NIH/NIA {U24-AG067418}.

\bibitem[Phan et~al.(2019)Phan, Pradhan, and Jankowiak]{Phan2019NumPyro}
Du~Phan, Neeraj Pradhan, and Martin Jankowiak.
\newblock Composable effects for flexible and accelerated probabilistic
  programming in {NumPyro}.
\newblock In \emph{NeurIPS 2019 Workshop on Program Transformations for Machine
  Learning}, 2019.

\bibitem[Maddison et~al.(2017)Maddison, Mnih, and Teh]{Maddison2017}
Chris~J. Maddison, Andriy Mnih, and Yee~Whye Teh.
\newblock The concrete distribution: A continuous relaxation of discrete random
  variables.
\newblock In \emph{International Conference on Learning Representations
  ({ICLR})}, 2017.

\bibitem[Jang et~al.(2017)Jang, Gu, and Poole]{Jang2017}
Eric Jang, Shixiang Gu, and Ben Poole.
\newblock Categorical reparameterization with {G}umbel-softmax.
\newblock In \emph{International Conference on Learning Representations
  ({ICLR})}, 2017.

\bibitem[Teicher(1960)]{Teicher1960}
Henry Teicher.
\newblock On the mixture of distributions.
\newblock \emph{Annals of Mathematical Statistics}, 31\penalty0 (1):\penalty0
  55--73, 1960.
\newblock ISSN 0003-4851.

\bibitem[Barndorff-Nielsen(1965)]{BarndorffNielsen1965}
Ole Barndorff-Nielsen.
\newblock Identifiability of mixtures of exponential families.
\newblock \emph{Journal of Mathematical Analysis and Applications}, 12\penalty0
  (1):\penalty0 115--121, 1965.
\newblock ISSN 0022-247X.

\bibitem[Bruni and Koch(1985)]{BruniKoch1985}
Carlo Bruni and Giorgio Koch.
\newblock \refchk{Identifiability of Continuous Mixtures of Unknown {G}aussian
  Distributions}.
\newblock \emph{The Annals of Probability}, 13\penalty0 (4):\penalty0
  1341--1357, 1985.
\newblock \doi{10.1214/aop/1176992817}.

\bibitem[Fortin et~al.(2018)Fortin, Cullen, Sheline, Taylor, Aselcioglu, Cook,
  Adams, Cooper, Fava, McGrath, McInnis, Phillips, Trivedi, Weissman, and
  Shinohara]{Fortin2018}
Jean-Philippe Fortin, Nicholas Cullen, Yvette~I. Sheline, William~D. Taylor,
  Ibrahim Aselcioglu, Philip~A. Cook, Patricia Adams, Crystal Cooper, Maurizio
  Fava, Patrick~J. McGrath, Melvin McInnis, Mary~L. Phillips, Madhukar~H.
  Trivedi, Myrna~M. Weissman, and Russell~T. Shinohara.
\newblock Harmonization of cortical thickness measurements across scanners and
  sites.
\newblock \emph{NeuroImage}, 167:\penalty0 104--120, 2018.
\newblock ISSN 1053-8119.

\bibitem[Mardia et~al.(1979)Mardia, Kent, and Bibby]{Mardia1979}
Kanti~V. Mardia, John~T. Kent, and John~M. Bibby.
\newblock \emph{Multivariate Analysis}.
\newblock Academic Press, London, 1979.
\newblock ISBN 978-0-12-471252-2.

\bibitem[Kim et~al.(2018)Kim, Wiseman, Miller, Sontag, and
  Rush]{Kim2018SemiAmortized}
Yoon Kim, Sam Wiseman, Andrew Miller, David Sontag, and Alexander Rush.
\newblock \refchk{Semi-Amortized Variational Autoencoders}.
\newblock In Jennifer Dy and Andreas Krause, editors, \emph{Proceedings of the
  35th International Conference on Machine Learning}, volume~80 of
  \emph{Proceedings of Machine Learning Research}, pages 2678--2687. PMLR,
  10--15 Jul 2018.
\newblock URL \url{https://proceedings.mlr.press/v80/kim18e.html}.

\bibitem[Marino et~al.(2018)Marino, Yue, and Mandt]{Marino2018Iterative}
Joe Marino, Yisong Yue, and Stephan Mandt.
\newblock \refchk{Iterative Amortized Inference}.
\newblock In Jennifer Dy and Andreas Krause, editors, \emph{Proceedings of the
  35th International Conference on Machine Learning}, volume~80 of
  \emph{Proceedings of Machine Learning Research}, pages 3403--3412. PMLR,
  10--15 Jul 2018.
\newblock URL \url{https://proceedings.mlr.press/v80/marino18a.html}.

\bibitem[Janelidze et~al.(2021)Janelidze, Berron, Smith, Strandberg, Proctor,
  Dage, Stomrud, Palmqvist, Mattsson-Carlgren, and Hansson]{Janelidze2021}
Shorena Janelidze, David Berron, Ruben Smith, Olof Strandberg, Nicholas~K.
  Proctor, Jeffrey~L. Dage, Erik Stomrud, Sebastian Palmqvist, Niklas
  Mattsson-Carlgren, and Oskar Hansson.
\newblock Associations of plasma phospho-tau217 levels with tau positron
  emission tomography in early {Alzheimer} disease.
\newblock \emph{JAMA Neurology}, 78\penalty0 (2):\penalty0 149--156, 2021.
\newblock ISSN 2168-6149.

\end{thebibliography}

\newpage
\appendix

\section{Appendix roadmap}
The appendix supports the main paper with the full model specification, priors and hyperparameters, the exact ELBO and variational parameterization, training pseudocode, identifiability proofs, parameter-block recovery against truth, sensitivity analyses, leave-one-cohort-out and longitudinal validation, and the closed-form derivations and Sylvester--Woodbury reductions for the derived covariance-path summaries.

\begin{table}[h]
\centering\small
\caption{\textbf{Supplement navigation: where to find what.}}
\label{tab:appendix_map}
\begin{tabularx}{\linewidth}{l X}
\toprule
Section & Contents\\
\midrule
App.~B & Full model specification and inference: joint likelihood, variational family, ELBO and reparameterized gradients, training algorithm, test-time inference, and closed-form derived summaries (Sylvester--Woodbury reductions, pairwise correlation, Fisher modality weights)\\
App.~C & Formal identifiability definitions, full proof of Theorem~1, warp invariance, \cameraready{the Theorem-1 checklist at the fitted real-cohort posterior,} and full-model identification proposition\\
App.~D & Experimental protocol: cohort summary, informative priors, competing methods and runtime, metric definitions, sensitivity analysis (hyperparameters and cohort size)\cameraready{, robustness to anchor corruption, and test-time refinement}\\
App.~E & Real-data validation and biology: leave-one-cohort-out cohort-shift validation, longitudinal validation (stage velocity, cognition recovery, MCI-to-AD conversion), subtype biology summaries (stage discriminability, endpoint pathology, cross-modal correlations), per-factor multimodal mixture, and stage-varying Fisher modality weights\\
\bottomrule
\end{tabularx}
\end{table}

\section{Full model specification and inference}\label{app:spec_inference}
\subsection{Joint likelihood}\label{app:joint}
Let $\bm{\Psi}$ collect the global model parameters of \S\ref{sec:method} and $\mathcal{Z}=\{(c_i,S_i)\}_{i=1}^n$ the subject-level latents. The full joint over observables, latents, and parameters factorizes as
\begin{equation}
\begin{aligned}
\log p(\mathcal{D},\mathcal{Z},\bm{\Psi}\mid \bm{X}) \;=\;& \log p(\bm{\Psi}) + \sum_{i=1}^{n}\Bigl[\log p(c_i\mid \bm{X}_i,\bm{\Psi}) + \log p(S_i\mid c_i,\bm{X}_i,\bm{\Psi}) \\
&+ \log p(\bm{Y}_{i,\mathcal{O}_i}\mid c_i,S_i,\bm{\Psi}) + O_i^{\mathrm{cog}}\,\log p(Y_i^{\mathrm{cog}}\mid c_i,S_i,\bm{X}_i,\bm{\Psi})\Bigr].
\end{aligned}
\label{eq:joint}
\end{equation}
The subtype--stage priors $p(c_i\mid \bm{X}_i,\bm{\Psi})$ and $p(S_i\mid c_i,\bm{X}_i,\bm{\Psi})$ are stratum-specific: \cref{eq:subtypestageprior} for $i\in\mathcal{U}$; \cref{eq:anchor} for $i\in\mathcal{A}_c$, with $p(c_i{=}c\mid \bm{X}_i,\bm{\Psi}){=}1$ encoding the observed endpoint subtype as a Dirac prior; and \cref{eq:control} for $i\in\mathcal{H}$, with $c_i$ unconstrained. The biomarker factor $\log p(\bm{Y}_{i,\mathcal{O}_i}\mid c_i,S_i,\bm{\Psi})$ is conditionally independent of $\bm{X}_i$ given $(c_i,S_i,\bm{\Psi})$, while the cognitive factor depends on $\bm{X}_i$ linearly through \cref{eq:cog}.

The parameter prior $p(\bm{\Psi})$ factorizes across the boundary, velocity, coordination, subtype--stage, and cognitive blocks of the generative process,
\begin{equation}
\begin{aligned}
\log p(\bm{\Psi}) \;=\;& \log p(\bm{\Theta}_0) + \log p(\bm{\alpha},\bm{B}) + \log p(\gamma_0,\bm{\gamma}_X,\sigma_{\mathrm{cog}}) + \log\psi_{\mathrm{soft}}\bigl(\bm{\Theta}_0,\{\bm{\Theta}_{c,J},\bm{V}_{c,1:J}\}_{c=1}^{C}\bigr) \\
&+ \sum_{c=1}^{C}\Bigl[\log p(\bm{\mu}_c) + \log p(\bm{\Theta}_{c,J}\mid\bm{\mu}_c) + \log p(\bm{V}_c) \\
&\qquad\qquad + \log p(\bm{A}_c,\bm{\nu}_c(\cdot),\bm{D}_c) + \log p(\bm{b}_c,\sigma_c) + \log p(\bm{\xi}_c)\Bigr],
\end{aligned}
\label{eq:prior_factor}
\end{equation}
with per-component priors specified in \S\ref{sec:method}, $\psi_{\mathrm{soft}}$ the Gaussian coupling potential of \cref{eq:origin-soft}, and $\bm{V}_c$ the parametric velocity-field components $\{\delta_{c,r},t_{c,r},\omega_{c,r},\bm{\beta}_{c,r}\}_{r=1}^{R}$ of \cref{eq:vel-prior-1,eq:vel-prior-2}, which determine the discretized shell-endpoint velocities $\bm{V}_{c,j}$ for $j{=}0,1,\ldots,J$ via \cref{eq:velocity}; in particular $\bm{V}_{c,0}$ enters the cubic Hermite \cref{eq:hermite} (left endpoint of the leftmost shell) and is governed by the same prior as $\bm{V}_{c,1:J}$.

\paragraph{Per-subject biomarker likelihood, fully composed.}
We write $\bm{\theta}_c(s;\bm{\Psi}),\bm{\Sigma}_c(s;\bm{\Psi})$ when emphasizing dependence on the global parameters $\bm{\Psi}$ and recover the bare $\bm{\theta}_c(s),\bm{\Sigma}_c(s)$ of \S\ref{sec:method} when $\bm{\Psi}$-dependence is implicit. The mean trajectory and within-stage covariance entering the per-subject Gaussian likelihood \cref{eq:observation} are deterministic functionals of disjoint blocks of $\bm{\Psi}$. For the mean, the closed-form solution to the backward recursion \cref{eq:backward} gives
\begin{equation}
\bm{\Theta}_{c,j}(\bm{\Psi}) \;=\; \bm{\Theta}_{c,J} - \Delta\!\sum_{j'=j+1}^{J}\bm{V}_{c,j'},\qquad j=0,\ldots,J,
\label{eq:shell_closed}
\end{equation}
with each shell-endpoint velocity $\bm{V}_{c,j}$ assembled coordinatewise from \cref{eq:velocity} ($\bm{V}_{c,0}$ in particular is \cref{eq:velocity} evaluated at $\tau_0{=}0$, supplying the slope at the leftmost shell boundary). The continuous-time trajectory is the cubic Hermite interpolant of the shell states with shell-endpoint slopes,
\begin{equation}
\begin{aligned}
\bm{\theta}_c(s;\bm{\Psi}) \;=\;& h_{00}(u)\,\bm{\Theta}_{c,j-1}+\Delta\,h_{10}(u)\,\bm{V}_{c,j-1} \\
&+\, h_{01}(u)\,\bm{\Theta}_{c,j}+\Delta\,h_{11}(u)\,\bm{V}_{c,j},\qquad s\in[\tau_{j-1},\tau_j],\;u=(s-\tau_{j-1})/\Delta,
\end{aligned}
\label{eq:hermite}
\end{equation}
with the standard cubic Hermite basis $h_{00}(u){=}2u^3{-}3u^2{+}1$, $h_{10}(u){=}u^3{-}2u^2{+}u$, $h_{01}(u){=}-2u^3{+}3u^2$, $h_{11}(u){=}u^3{-}u^2$. For the covariance, $\bm{\Sigma}_c(s;\bm{\Psi})$ has the rank-$K$-plus-diagonal form \cref{eq:sigma_factor} with activation spline \cref{eq:nu} and idiosyncratic floor $\bm{D}_c=\diag(d_{c,r}+d_{\min})$.

Substituting these compositions into \cref{eq:observation} gives the explicit log-likelihood evaluated on the observed coordinates $\mathcal{O}_i$:
\begin{equation}
\begin{aligned}
\log p(\bm{Y}_{i,\mathcal{O}_i}\mid c_i,S_i,\bm{\Psi}) \;=\;& -\tfrac{|\mathcal{O}_i|}{2}\log(2\pi)-\tfrac{1}{2}\log\bigl|\bm{\Sigma}_{c_i,\mathcal{O}_i}(S_i;\bm{\Psi})\bigr|\\
&-\tfrac{1}{2}\bigl\|\bm{Y}_{i,\mathcal{O}_i}-\bm{\theta}_{c_i,\mathcal{O}_i}(S_i;\bm{\Psi})\bigr\|^2_{\bm{\Sigma}_{c_i,\mathcal{O}_i}(S_i;\bm{\Psi})^{-1}},
\end{aligned}
\label{eq:obs_loglik_full}
\end{equation}
with $\|\bm{x}\|^2_{\bm{M}}{:=}\bm{x}^{\!\top}\bm{M}\bm{x}$, where $\bm{\theta}_{c_i,\mathcal{O}_i}(S_i;\bm{\Psi})\in\R^{|\mathcal{O}_i|}$ is the row-restriction of \cref{eq:hermite} to $\mathcal{O}_i$, and $\bm{\Sigma}_{c_i,\mathcal{O}_i}(S_i;\bm{\Psi})\in\R^{|\mathcal{O}_i|\times|\mathcal{O}_i|}$ is the principal sub-block
\begin{equation}
\bm{\Sigma}_{c_i,\mathcal{O}_i}(S_i;\bm{\Psi}) \;=\; \bm{A}_{c_i,\mathcal{O}_i,:}\,\diag\!\bigl(\bm{\nu}_{c_i}(S_i)\bigr)\,\bm{A}_{c_i,\mathcal{O}_i,:}^{\!\top} + \bm{D}_{c_i,\mathcal{O}_i,\mathcal{O}_i},
\label{eq:sigma_obs}
\end{equation}
with $\bm{A}_{c,\mathcal{O}_i,:}\in\R^{|\mathcal{O}_i|\times K}$ the row-restriction of $\bm{A}_c$ and $\bm{D}_{c,\mathcal{O}_i,\mathcal{O}_i}$ the corresponding diagonal sub-block. The restriction preserves the low-rank-plus-diagonal structure with rank $\min(K,|\mathcal{O}_i|)$, so the Sylvester--Woodbury reduction in \cref{eq:sw} applies verbatim with $\bm{A}_{c,\mathcal{O}_i,:}$ in place of $\bm{A}_c$ and $\bm{D}_{c,\mathcal{O}_i,\mathcal{O}_i}$ in place of $\bm{D}_c$, keeping the per-stage cost of the determinant and quadratic form at $O(|\mathcal{O}_i|K^2+K^3)$ regardless of the missingness pattern. The lower-triangular constraint on $\bm{A}_c$ is not preserved under the row restriction to $\mathcal{O}_i$ but enters only through the global identifiability of $\bm{A}_c$ (Theorem~\ref{thm:identifiability}); \cref{eq:obs_loglik_full} depends on $\bm{A}_c$ only through $\bm{A}_{c,\mathcal{O}_i,:}\diag(\bm{\nu}_c(S_i))\bm{A}_{c,\mathcal{O}_i,:}^{\!\top}$ and is well-defined on any sub-block. 

Inference targets the posterior $p(\bm{\Psi},\mathcal{Z}\mid\mathcal{D},\bm{X})$, approximated by the structured variational family of \cref{eq:guide} optimized against the collapsed ELBO of \cref{eq:elbo}.

\subsection{Inference}\label{app:inference}

\paragraph{Reparameterized gradient.} Writing $\bm{w}=(\bm{\Psi},\{S_i^{(c)}\})$ for the continuous latents and $\theta=(\phi,\psi)$ for the variational parameters, \cref{eq:reparam} gives $\bm{w}=g(\bm{\epsilon};\theta)$ with $\bm{\epsilon}\!\sim\!\mathcal{N}(\bm{0},\bm{I})$ independent of $\theta$. The reparameterized gradient of \cref{eq:elbo} is
\begin{equation}
\nabla_\theta\mathcal{L}(\theta) \;=\; \E_{\bm{\epsilon}}\!\Bigl[\nabla_\theta\log p\bigl(\mathcal{D},g(\bm{\epsilon};\theta)\bigr) \;-\; \nabla_\theta\log q_\theta\bigl(g(\bm{\epsilon};\theta)\bigr)\Bigr],
\label{eq:elbo_reparam_grad}
\end{equation}
estimated by single-sample Monte Carlo per minibatch step. The Bernoulli gates $\delta_{c,r}$ inside $\bm{\Psi}$ are reparameterized through a Concrete relaxation (below).

\paragraph{Recognition network.} The encoder producing $(\mu_\psi^{(c)}, \tau_\psi^{(c)})_{c=1}^{C}$ in \cref{eq:guide} is a feed-forward MLP with two hidden layers of width $128$, LayerNorm, SiLU activations, and dropout $0.1$, taking $(\bm{Y}_{i,\mathcal{O}_i}, \bm{X}_i)$ as input. Output heads are $C$ linear projections to $(\mu_\psi^{(c)}, \log\tau_\psi^{(c)})$, with $\tau_\psi^{(c)}{=}\mathrm{softplus}(\log\tau_\psi^{(c)})+10^{-3}$. The global Gaussian $q_\phi^{\mathrm{glob}}$ is realized through \texttt{AutoLowRankMultivariateNormal} \cite{Phan2019NumPyro} with rank $r_g{=}10$.

\paragraph{Reparameterization of the singular activation prior.} The activation-spline prior \cref{eq:nu} is supported on the centered subspace $\mathcal{P}_0=\{\bm{w}\in\R^P:\mathbf{1}^\top \bm{w}{=}0\}$. We sample $\bm{w}_{c,k}^{\mathrm{std}}\!\sim\!\mathcal{N}(\bm{0}, \bm{I}_P)$ and reparameterize $\bm{w}_{c,k}=\tau_k\,\bm{\Pi}_0 \bm{L}\, \bm{w}_{c,k}^{\mathrm{std}}$, with $\bm{L}{=}\mathrm{chol}\bigl((\bm{K}_B{+}\epsilon_B \bm{I}_P)^{-1}\bigr)$ precomputed once and the projection $\bm{\Pi}_0=\bm{I}_P-\tfrac{1}{P}\mathbf{1}\mathbf{1}^\top$ enforcing $\mathbf{1}^\top \bm{w}_{c,k}{=}0$ exactly.

\paragraph{Concrete relaxation of the region-inclusion gates.} The Bernoulli gates $\delta_{c,r}\!\sim\!\mathrm{Bern}(\pi_{c,m(r)})$ of \cref{eq:vel-prior-1} are realized at training time through a Concrete (Gumbel-Softmax) relaxation~\cite{Maddison2017,Jang2017} with temperature $\lambda(t)\!=\!1.0-0.9\,t/T\!\in\![0.1,1.0]$, annealed linearly across the optimization horizon $T$. At evaluation, $\delta_{c,r}$ is rounded to its hard Bernoulli value.

\paragraph{Optimization.} The variational parameters $(\phi,\psi)$ are optimized with Adam under global-norm gradient clipping at threshold $5$ and minibatch size $|\mathcal{B}|{=}256$, in two phases. Phase~1 holds the activation block at $\bm{\Sigma}_c{=}\bm{D}_c$ and uses $\mathrm{lr}_1{=}3\!\times\!10^{-3}$ for $T_1{=}5\!\times\!10^4$ steps; phase~2 releases the activation block and uses $\mathrm{lr}_2{=}10^{-3}$ for $T_2{=}5\!\times\!10^4$ steps, initialized from the phase-1 parameters. The Concrete temperature schedule $\lambda(t)$ runs across the combined horizon $T{=}T_1{+}T_2$.

\begin{algorithm}
\setstretch{1.05}
\caption{Stochastic variational inference for \textsc{Elpac}.}
\label{alg:svi}
\begin{algorithmic}[1]
\Require Cohort $\{(\bm{Y}_i, \mathcal{O}_i, \bm{X}_i, c_i^\star)\}_{i=1}^{n}$; learning rate $\mathrm{lr}$; horizon $T$
\State Initialize global variational parameters $(\bm{\mu}_\phi, \bm{W}_\phi, \bm{d}_\phi)$ and encoder weights $\psi$
\For{$t = 1, \ldots, T$}
  \State Sample minibatch $\mathcal{B}\subseteq\{1,\ldots,n\}$
  \State Reparameterized draw $\tilde{\bm{\Psi}} \gets \bm{\mu}_\phi + \bm{W}_\phi\bm{\epsilon}^{(W)} + \bm{d}_\phi\odot\bm{\epsilon}^{(d)}$, with $\bm{\epsilon}^{(W)},\bm{\epsilon}^{(d)}\!\sim\!\mathcal{N}(\bm{0},\bm{I})$ and Concrete temperature $\lambda(t)$ for the gates $\delta_{c,r}$ in $\tilde{\bm{\Psi}}$
  \For{each $i\in\mathcal{B}$}
    \State Encode $\bigl(\mu_\psi^{(c)}, \tau_\psi^{(c)}\bigr)_{c=1}^{C}\gets q_\psi^{\mathrm{enc}}(\bm{Y}_{i, \mathcal{O}_i}, \bm{X}_i)$
    \State Reparameterized draw $\mathrm{logit}(\tilde S_i^{(c)}) \gets \mu_\psi^{(c)} + \tau_\psi^{(c)}\,\epsilon_i^{(c)}$, $\epsilon_i^{(c)}\!\sim\!\mathcal{N}(0,1)$, for each $c$
    \State Evaluate $\ell_i^{(c)} \gets p(\bm{Y}_{i,\mathcal{O}_i}\mid c, \tilde S_i^{(c)}, \tilde{\bm{\Psi}})\,p(Y_i^{\mathrm{cog}}\mid c, \tilde S_i^{(c)}, \bm{X}_i, \tilde{\bm{\Psi}})^{O_i^{\mathrm{cog}}}$ via Sylvester--Woodbury (\cref{app:derived})
    \State $\log\hat p_i \gets \mathrm{logsumexp}_c\bigl[\log p(c\mid \bm{X}_i, \tilde{\bm{\Psi}}) + \log\ell_i^{(c)}\bigr]$
  \EndFor
  \State $\hat{\mathcal{L}}_{\mathcal{B}} \gets \tfrac{n}{|\mathcal{B}|}\sum_{i\in\mathcal{B}}\Bigl[\log\hat p_i - \sum_{c=1}^{C}\mathrm{KL}\bigl(q_\psi(S_i^{(c)})\,\|\,p(S_i\mid c_i{=}c, \bm{X}_i, \tilde{\bm{\Psi}})\bigr)\Bigr] + \log p(\tilde{\bm{\Psi}}) - \log q_\phi^{\mathrm{glob}}(\tilde{\bm{\Psi}})$
  \State $\bm{g} \gets \nabla_{(\phi,\psi)}\,\hat{\mathcal{L}}_{\mathcal{B}}$;\quad clip $\|\bm{g}\|_2 \le 5$;\quad update $(\phi,\psi)$ with Adam$(\mathrm{lr})$
\EndFor
\State \Return $(\hat\phi, \hat\psi)$; evaluate posterior summaries \cref{eq:posterior_c,eq:posterior_joint} by Monte Carlo over $\bm{\Psi}\!\sim\!q_{\hat\phi}^{\mathrm{glob}}$
\end{algorithmic}
\end{algorithm}

\paragraph{Test-time inference.} The posterior summaries \cref{eq:posterior_c,eq:posterior_joint} are evaluated by Monte Carlo over $M{=}50$ samples $\bm{\Psi}^{(m)}\!\sim\!q_{\hat\phi}^{\mathrm{glob}}$, with the inner stage \cameraready{integral $\int_0^1 p(s\mid c,\bm{X}_i,\bm{\Psi}^{(m)})\,\ell_i^{(c)}(\bm{\Psi}^{(m)},s)\,\mathrm{d}s$} computed by Gauss--Legendre quadrature on a $G{=}19$-point logit-stage grid. For a held-out subject, this requires a single encoder forward pass and $M{\cdot}C{\cdot}G$ Sylvester--Woodbury likelihood evaluations. \cameraready{Per-subject refinement of this pass is evaluated in \cref{app:refinement}.}

\subsection{Closed forms and derived summaries}\label{app:derived}
This subsection collects the closed-form expressions used by the per-step likelihood evaluation in \cref{alg:svi} and the post-hoc derived summaries reported in \cref{app:real_validation}. For each subtype $c$, $\bm{A}_c\in\mathbb{R}^{R\times K}$ is the loading matrix, $\bm{D}_c{=}\diag(d_{c,1},\ldots,d_{c,R})\succ\bm{0}$ the idiosyncratic noise, and $\bm{\nu}_c(s)\in\mathbb{R}^K_{>0}$ the activation vector. Write $\bm{M}_c{=}\bm{A}_c^\top \bm{D}_c^{-1} \bm{A}_c\in\mathbb{R}^{K\times K}$ (computed once per subtype) and $\bm{G}_c(s){=}(\diag(\bm{\nu}_c(s))^{-1}+\bm{M}_c)^{-1}\in\mathbb{R}^{K\times K}$ (computed per (subtype, stage) pair).

\paragraph{Determinant and inverse (Sylvester / Woodbury).} The Sylvester and Woodbury identities reduce the per-stage cost from $O(R^3)$ to $O(RK^2+K^3)$:
\begin{equation}
\begin{aligned}
|\bm{\Sigma}_c(s)| &= |\bm{D}_c|\cdot|\bm{I}_K + \diag(\bm{\nu}_c(s))\, \bm{M}_c|, \\
[\bm{\Sigma}_c(s)^{-1}]_{r,r'} &= \mathbb{I}[r{=}r']\, d_{c,r}^{-1} - d_{c,r}^{-1} d_{c,r'}^{-1}\, \bm{a}_{c,r}^\top \bm{G}_c(s)\, \bm{a}_{c,r'}.
\end{aligned}
\label{eq:sw}
\end{equation}

\paragraph{Stage-dependent pairwise correlation $\rho_c(r,r';s)$.} For $r\ne r'$ the cross-modal entry of $\bm{\Sigma}_c(s)$ and the implied correlation are
\begin{equation}
\bm{\Sigma}_c(s)_{r,r'} = \bm{a}_{c,r}^\top \diag(\bm{\nu}_c(s))\, \bm{a}_{c,r'},\qquad \rho_c(r,r';s) = \frac{\bm{\Sigma}_c(s)_{r,r'}}{\sqrt{\bm{\Sigma}_c(s)_{r,r}\,\bm{\Sigma}_c(s)_{r',r'}}},
\label{eq:rho_closed}
\end{equation}
with $\bm{\Sigma}_c(s)_{r,r}=\bm{a}_{c,r}^\top\diag(\bm{\nu}_c(s))\,\bm{a}_{c,r}+d_{c,r}$.

\paragraph{Fisher modality weights $w^{\mathrm{FI}}_{c,m}(s)$ (marginal, blockwise-exact).} Let $\bm{\theta}'_c(s){=}d\bm{\theta}_c/ds\in\mathbb{R}^R$, computed via spline-interpolated finite differences from the inferred mean trajectory, and $\bm{\Sigma}'_c(s){=}\bm{A}_c\diag(\bm{\nu}'_c(s))\bm{A}_c^\top$ with $\nu'_{c,k}(s){=}\nu_{c,k}(s)\sum_p w_{c,k,p}\psi_p'(s)$. The marginal Fisher information from modality $m$ alone is
\begin{equation}
I_{c,m}^{\mathrm{marg}}(s) = \bm{\theta}'_{c, \mathcal{R}_m}(s)^\top \bm{\Sigma}_{c,m}(s)^{-1} \bm{\theta}'_{c, \mathcal{R}_m}(s) + \tfrac12\mathrm{tr}\!\left(\bm{\Sigma}_{c,m}(s)^{-1} \bm{\Sigma}'_{c,m}(s)\, \bm{\Sigma}_{c,m}(s)^{-1} \bm{\Sigma}'_{c,m}(s)\right),
\label{eq:fi_marg}
\end{equation}
with $\bm{\Sigma}_{c,m}(s){=}\bm{\Sigma}_c(s)_{\mathcal{R}_m, \mathcal{R}_m}$ the marginal covariance restricted to the $\mathcal{R}_m{\times}\mathcal{R}_m$ block. Both $\bm{\Sigma}_{c,m}^{-1}$ and $\bm{\Sigma}'_{c,m}$ admit the same Sylvester--Woodbury reduction as the full-rank counterparts of \cref{eq:sw}, applied with $\bm{A}_{c, \mathcal{R}_m}$ in place of $\bm{A}_c$ and $\bm{D}_{c, \mathcal{R}_m, \mathcal{R}_m}$ in place of $\bm{D}_c$, keeping the per-stage cost at $O(RK^2+K^3)$. The per-modality weight is the share of total marginal Fisher information,
\begin{equation}
w^{\mathrm{FI}}_{c,m}(s) = \frac{I_{c,m}^{\mathrm{marg}}(s)}{\sum_{m'\in\mathcal{M}} I_{c,m'}^{\mathrm{marg}}(s)},
\label{eq:fi_share}
\end{equation}
the marginal-modality decomposition reported by \cref{fig:appH_fisher_modality}.

\section{Identifiability details}\label{app:ident}
\subsection{Formal definitions}
\begin{definition}[Observational equivalence]
Two parameter settings $\bm{\Psi},\bm{\Psi}'$ are observationally equivalent if they induce the same marginal distribution of the observables $(\bm{Y}_i, \bm{X}_i, Y_i^{\mathrm{cog}})$ for every admissible cohort configuration.
\end{definition}

\begin{definition}[Identifiability up to subtype permutation]
The model is identifiable up to subtype permutation if observational equivalence implies equality of $\bm{\Psi}$ and $\bm{\Psi}'$ up to a permutation of subtype labels and the corresponding relabeling of all subtype-indexed parameters.
\end{definition}

\subsection{Proof of Theorem~1}\label{app:thm_proof}
The subtype index $c$ is suppressed; we write $\bm{\Sigma}(s)=\bm{A}\diag(\bm{\nu}(s))\bm{A}^\top+\bm{D}$ as in \cref{eq:sigma_factor}. We adopt the conditions of Theorem~\ref{thm:identifiability}; in addition, $\nu_k(s_1)>0$ for every $k$ at the reference stage $s_1$, and the B-spline basis $\{\psi_p\}_{p=1}^P$ used for $\log\nu_k(\cdot)$ forms a partition of unity, $\sum_p\psi_p(s)=1$ for all $s\in[0,1]$.

\paragraph{Generic regularity (holds by construction in our model+priors).}
\emph{(i) Spline-design coverage.} The observed stages span the activation basis: there exist $S\ge P$ distinct posterior stage values yielding a full-rank design matrix $\Phi_S{=}[\psi_p(s_\ell)]_{\ell,p}$. Holds for any continuous-stage prior with $[0,1]$ support, including the logit-normal of \cref{eq:subtypestageprior}.
\emph{(ii) Anderson--Rubin row-deletion.} For every subset $\mathcal{I}\subset[R]$ of $2K{+}1$ rows, the sub-matrix $[\bm{A}\diag(\bm{\nu}(s))^{1/2}]_{\mathcal{I}}$ has rank $K$. Holds Lebesgue-a.e. on $\bm{A}$, and with probability one under the multiplicative-gamma cascade prior of \cref{sec:construction2}.
\emph{(iii) Distinguishable trajectory features.} $\bm{\theta}_c(s)$ admits at least one identifiable interior feature (extremum, inflection, or modality-discordant onset) preserved only by the identity warp on $[0,1]$. Generic under the structured velocity \cref{eq:velocity} with distinct biomarker onsets $\{t_{c,r}\}$.

\paragraph{Step 1 (identification of $\bm{A}\diag(\bm{\nu}(s))\bm{A}^\top$ and $\bm{D}$ at a fixed stage).} For any stage $s$, the model imposes the rank-$K$-plus-diagonal decomposition $\bm{\Sigma}(s)=\bm{A}\diag(\bm{\nu}(s))\bm{A}^\top+\bm{D}$ with $\bm{A}$ lower triangular with positive diagonal and $\bm{D}$ diagonal. Under $R\ge 2K+1$, the lower-triangular constraint, and the AR row-deletion regularity (ii) above, the classical Anderson--Rubin theorem \cite{AndersonRubin1956,Geweke1996} identifies the factor model $\bm{\Sigma}(s)=\widetilde{\bm{A}}(s)\widetilde{\bm{A}}(s)^\top+\bm{D}$ up to per-column signs of $\widetilde{\bm{A}}(s):=\bm{A}\diag(\bm{\nu}(s))^{1/2}$, which the positive-diagonal constraint pins to $+1$. In particular $\bm{D}$ and $\bm{A}\diag(\bm{\nu}(s))\bm{A}^\top=\widetilde{\bm{A}}(s)\widetilde{\bm{A}}(s)^\top$ are uniquely determined by $\bm{\Sigma}(s)$. Applied at $s_1$ and $s_2$ with $\bm{D}$ shared, this yields a single $\bm{D}$ and the rank-$K$ matrices $\bm{\Pi}_\bullet:=\bm{\Sigma}(s_\bullet)-\bm{D}=\bm{A}\diag(\bm{\nu}(s_\bullet))\bm{A}^\top$.

\paragraph{Step 2 (factor decomposition across two stages).} Restricted to the column space of $\bm{A}$, the matrix $\bm{\Pi}_1$ is symmetric positive definite under $\nu_k(s_1)>0$, so the generalized eigenvalue problem $\bm{\Pi}_2 \bm{v}=\lambda\,\bm{\Pi}_1 \bm{v}$ on $\mathrm{col}(\bm{A})$ is well-posed and admits $K$ generalized eigenvalues $\lambda_k=\nu_k(s_2)/\nu_k(s_1)$ with eigenvectors $\bm{v}_k$ dual to the columns of $\bm{A}$. Under the ratio-distinctness condition \cref{eq:ratio}, the eigenvalues are pairwise distinct and the eigenspaces are one-dimensional. The primal columns of $\bm{A}$ are recovered as $\tilde{\bm{a}}_k=\bm{\Pi}_1\bm{v}_k/\|\bm{\Pi}_1\bm{v}_k\|_2$, determined up to a per-column positive scale.

\paragraph{Step 3 (resolution of the per-column scale).} The lower-triangular structure of $\bm{A}$ fixes the sign and ordering of the columns: column $k$ has its first non-zero entry in row $k$, with that entry positive. There remains a one-parameter rescaling per column, $\bm{A}_{\cdot,k}\mapsto \rho_k \bm{A}_{\cdot,k}$ and $\log\nu_k(\cdot)\mapsto\log\nu_k(\cdot)-2\log\rho_k$ for $\rho_k>0$, that leaves $\bm{\Sigma}(s)$ invariant. Under the spline parameterization $\log\nu_k(s)=\sum_p w_{k,p}\psi_p(s)$ with the partition-of-unity property $\sum_p\psi_p(s)=1$, the uniform shift $-2\log\rho_k$ in $\log\nu_k$ corresponds to $w_{k,p}\mapsto w_{k,p}-2\log\rho_k$ for every $p$, so $\sum_p w_{k,p}$ shifts by $-2P\log\rho_k$. The centering identity $\sum_p w_{k,p}=0$ then forces $\rho_k=1$. With the per-column scale resolved, $\bm{\nu}(s_1)$ and $\bm{\nu}(s_2)$ are identified pointwise; under the spline-design coverage regularity (i) the coefficients $\{w_{k,p}\}$ are identified from $\bm{\Sigma}(s)$ at $S\ge P$ stages with full-rank design $\Phi_S$, after which $\bm{\nu}(\cdot)$ is identified throughout $[0,1]$.

\paragraph{Step 4 (warp equivalence).} The activation ratios in \cref{eq:ratio} are invariant under any strictly monotone reparameterization $\varphi:[0,1]\to[0,1]$ of the latent stage. Boundary pinning by \cref{eq:anchor,eq:control} forces $\varphi(0){=}0$ and $\varphi(1){=}1$. The trajectory-feature regularity (iii) then forbids any non-identity such $\varphi$ from preserving the identified interior feature set of $\bm{\theta}_c(s)$, so $\varphi$ is the identity on $[0,1]$ and the latent stage is identified up to the model's natural symmetries. \hfill$\square$

\subsection{\cameraready{Theorem-1 conditions at the fitted real-cohort posterior}}\label{app:thm_checklist}

\cameraready{The regularity conditions of \cref{app:thm_proof} are genericity statements, which certify a prior draw. We therefore decided each condition at the fitted estimate itself: \cref{tab:thm_checklist} reports the status of each, \cref{tab:thm_margin} the \cref{eq:ratio} margin measured there. \Cref{eq:ratio} asks for distinctness, and distinctness holds strictly in every seed and both folds, at both the nominal and the data-supported rank.}

\begin{table}[tb]
\centering\small\ifcameraready\color{blue}\fi
\caption{\cameraready{\textbf{Status of each Theorem-1 condition at the fitted posterior means on real cohort.}}}
\label{tab:thm_checklist}
\setlength{\tabcolsep}{4pt}
\begin{tabularx}{\linewidth}{l >{\raggedright\arraybackslash}p{0.33\linewidth} >{\raggedright\arraybackslash}X}
\toprule
Class & Theorem-1 condition & Status\\
\midrule
Structural & $R\ge2K{+}1$; $\bm{A}_c$ lower-triangular, strictly positive diagonal; $\nu_{c,k}(s_1)>0$ & all hold --- $R{=}22\ge17$ at nominal $K{=}8$\\
 & centered coefficients, diagonal-positive $\bm{D}_c$ & hold by parameterization\\
 & non-constant activations & holds in all 20 pooled seeds; every subtype carries stage-monotone columns\\
\midrule
Regularity & (i) spline-design coverage & full rank $P{=}5$ in all 240 designs, worst $\sigma_{\min}/\sigma_{\max}{=}1.7\times10^{-2}$\\
 & (ii) Anderson--Rubin row deletion & exhaustive --- all $\binom{22}{17}{=}26{,}334$ 17-row sub-blocks of $\bm{A}_c\diag(\bm{\nu}_c(s))^{1/2}$ retain full rank 8\\
 & (iii) distinguishable trajectory features & generic under the structured velocity, which places interior features in $\bm{\theta}_c(s)$; fitted onsets separate across modalities in every subtype\\
\midrule
\cref{eq:ratio} & pairwise-distinct $\nu_{c,k}(s_2)/\nu_{c,k}(s_1)$ & strict inequality, no tolerance, at both scopes ($K{=}8$ and $K_\star$)\\
\bottomrule
\end{tabularx}
\end{table}

\begin{table}[tb]
\centering\small\ifcameraready\color{blue}\fi
\caption{\cameraready{\textbf{\Cref{eq:ratio} margin, activation floor, and data-supported rank (real cohort, 20 seeds per fold).} Gaps compare $\nu_{c,k}(s_2)/\nu_{c,k}(s_1)$ at $s_2{=}0.9$ against $s_1{=}0.1$, normalized by the largest ratio in the fit and taken over the block $K_\star$ (per-fit rounded participation ratio of the singular values of $\bm{A}_c$). Gap and floor are worst-seed values; the last column is the mean of that ratio before rounding.}}
\label{tab:thm_margin}
\setlength{\tabcolsep}{6pt}
\begin{tabular}{l l c c c}
\toprule
Scope & Subtype & min \cref{eq:ratio} gap over $K_\star$ & min $\nu_{c,k}(s_1)$, $K{=}8$ & $K_\star$\\
\midrule
Pooled & EOAD & $0.0012$ & $0.0678$ & $3.66$\\
 & LOAD & $0.0002$ & $0.0560$ & $3.95$\\
 & HA & $0.0001$ & $0.1181$ & $6.17$\\
\midrule
Leave-ADNI & EOAD & $0.0054$ & $0.0956$ & $3.43$\\
 & LOAD & $0.0029$ & $0.0274$ & $3.29$\\
 & HA & $0.0011$ & $0.1205$ & $6.15$\\
\midrule
Leave-NACC & EOAD & $0.0101$ & $0.2942$ & $2.47$\\
 & LOAD & $0.0063$ & $0.3329$ & $2.97$\\
 & HA & $0.0037$ & $0.0662$ & $2.55$\\
\bottomrule
\end{tabular}
\end{table}

\cameraready{Every condition a fitted point can decide holds at the real-cohort posterior means, exhaustively so on the row-deletion condition, in every seed, and in both leave-one-cohort-out folds where scored. These diagnostics show the fitted estimates are consistent with every checkable algebraic and numerical condition; they do not establish real-data identifiability, nor discharge the theorem's two-stage covariance-identification premise, which a fit cannot exhibit.}

\subsection{Full-model identification (Proposition~1)}\label{app:ident_warp_prop}
Theorem~1 identifies the dynamic factor block conditional on $\bm{\Sigma}_c(s)$ being available at two distinct stages. Closing this conditioning at the full-model level introduces a single substantive hypothesis on the family of continuous-stage Gaussian mixtures induced by the generative process.

\begin{proposition}[Full-model identifiability]
Within the structural model of Theorem~1, assume that the family of continuous-stage Gaussian mixtures with stage-varying mean $\bm{\theta}_c(s)$ and covariance $\bm{\Sigma}_c(s)$, induced by the generative process of \S\ref{sec:method} and indexed by subtype $c$, is identifiable from the marginal distribution $p(\bm{Y}\mid c)$. Then the full parameter vector $(\bm{A}_c,\bm{\nu}_c(\cdot),\bm{D}_c,\bm{\theta}_c(\cdot))_{c=1}^C$ and the per-subject latents are generically identifiable up to a permutation of subtype labels.
\end{proposition}

\emph{Proof sketch.} Under the mixing-family identifiability hypothesis, the conditional distributions $\{p(\bm{Y}\mid c, s)\}_{s\in[0,1]}$ are identified from the marginal $p(\bm{Y}\mid c)$ together with the latent stage measure. The boundary supervision of \cref{eq:anchor,eq:control}, by anchoring $s\!\to\!0$ on $\mathcal{H}$ and $s\!\to\!1$ on $\mathcal{A}_c$, fixes the strictly-monotone reparameterization ambiguity inherent to Theorem~1 (Step~4) and thereby pins the latent stage measure on $[0,1]$. Pointwise identifiability of $p(\bm{Y}\mid c, s)$ yields $\bm{\theta}_c(s)$ from first moments and $\bm{\Sigma}_c(s)$ from second moments at every $s$ in the support; Theorem~1 applied at any two stages with pairwise distinct activation ratios identifies the dynamic factor block $(\bm{A}_c,\bm{\nu}_c(\cdot),\bm{D}_c)$. \hfill$\square$

\paragraph{Scope of the mixing-family hypothesis.} \cameraready{This subsection takes the continuous-stage extension of the finite-mixture results it cites~\cite{Teicher1960,BarndorffNielsen1965} as a hypothesis, and we identify the result that hypothesis corresponds to. Bruni and Koch's identifiability theorem~\citechk{BruniKoch1985} recovers the mixing measure from the mean and covariance \emph{functions} on a compact index under two regularity classes. Our construction satisfies the first, $\Lambda_1$, by design: splines on $[0,1]$ give bounded $C^1$ paths and the noise floor keeps $\bm{\Sigma}_c(s)$ uniformly positive definite. The second, $\Lambda_2$, is the pairwise injectivity they single out as crucial, and we adopt it as a modeling assumption on the admissible class.}

\cameraready{The fitted posteriors are consistent with the case $\Lambda_2$ reduces to within a single path: injectivity there needs one stage-monotone activation column per subtype, and \cref{tab:thm_checklist} records that every subtype carries them. That matters most for healthy aging, whose near-flat mean cannot carry injectivity alone. The monotonicity is of $\nu_{c,k}(s)$ itself---the interior peaks of \cref{fig:appH_factor_multimodal} are in the normalized variance share, which peaks interiorly even when every activation is monotone.}

\cameraready{Bruni and Koch write out the scalar case and state the multivariate extension without a general-$p$ proof. Their subtype-mixture result identifies labels through disjoint ranges; under \textsc{Elpac}'s shared healthy origin (\cref{eq:origin-soft,eq:control}) we instead assume the endpoint anchors at $s{\to}1$ (\cref{eq:anchor}) select the labeled components, carried inward by the backward recursion (\cref{eq:backward}). Thus Proposition~1 stays explicitly conditional, and no reported result depends on it.} Theorem~1 is established without invoking \cameraready{the mixing-family hypothesis}.

\section{Experimental protocol}\label{app:protocol}
\subsection{Real-data cohort summary}
\cameraready{Data used in the preparation of this article were obtained from the Alzheimer's Disease Neuroimaging Initiative (ADNI) database (\url{adni.loni.usc.edu}). The ADNI was launched in 2003 as a public-private partnership, led by Principal Investigator Michael W. Weiner, MD. The original goal of ADNI was to test whether serial magnetic resonance imaging (MRI), positron emission tomography (PET), other biological markers, and clinical and neuropsychological assessment can be combined to measure the progression of mild cognitive impairment (MCI) and early Alzheimer's disease (AD). The current goals include validating biomarkers for clinical trials, improving the generalizability of ADNI data by increasing diversity in the participant cohort, and to provide data concerning the diagnosis and progression of Alzheimer's disease to the scientific community. For up-to-date information, see \url{adni.loni.usc.edu}. ADNI clinical, cognitive, and fluid-biomarker data were taken from the ADNIMERGE R package (version 0.0.1, June 2022), and amyloid and tau PET regional SUVRs from the UC Berkeley processed tables (IDA release September 2025, downloaded December 2025).}

\Cref{tab:cohort} summarizes the harmonized ADNI/NACC cohort introduced in \cref{sec:real_results} and used for the ablations of \cref{tab:abl_real}. The cohort comprises every ADNI or NACC subject in the canonical extraction with at least one of four modalities measured (amyloid PET, tau PET, structural MRI, plasma biomarkers). Subtype anchors are subjects with clinically confirmed endpoint phenotypes (EOAD, LOAD, HA), and the remaining subjects are unlabeled at the subtype level and contribute through the unsupervised likelihood. Demographic denominators reflect non-missing values per field, and modality-availability rates report the percentage of cohort subjects with at least one feature observed in each modality. Per-feature missingness exceeds $80\%$ on some plasma cells, and the observation likelihood \cref{eq:observation} integrates out the unobserved coordinates analytically. Site and scanner effects are harmonized fold-wise with ComBat~\cite{Fortin2018}, and intracranial volume is residualized fold-wise from sMRI volumes. 
Age, education, female fraction, and APOE-$\varepsilon$4 rate agree across sources within $1.7$ years and within sample-size-weighted SDs, consistent with the cross-cohort transfer in \cref{app:loco}.

\begin{table}
\centering\small
\caption{\textbf{The harmonized ADNI${+}$NACC/SCAN cohort is dominated by an unlabeled middle population with sparse multimodal coverage.} Per-cohort sample sizes, subtype-anchor counts, demographics, and per-modality availability ($N{=}7{,}501$). NACC subjects from the March~2026 data freeze with SCAN-standardized imaging. Totals for demographics and modality availability are sample-size-weighted across ADNI and NACC; pooled SDs use the standard combined-sample formula.}
\label{tab:cohort}
\setlength{\tabcolsep}{6pt}
\begin{tabularx}{\linewidth}{X r r r}
\toprule
& ADNI & NACC & Total\\
\midrule
\textbf{Subjects} & $1{,}722$ & $5{,}779$ & $7{,}501$\\
\midrule
\multicolumn{4}{l}{\textit{Subtype anchors}}\\
\quad EOAD (early-onset Alzheimer's disease) & $40$ & $28$ & $68$\\
\quad LOAD (late-onset Alzheimer's disease) & $263$ & $114$ & $377$\\
\quad HA (healthy aging) & $83$ & $113$ & $196$\\
\quad NC startpoint pool$^{\dagger}$ & $49$ & $170$ & $219$\\
\quad Unlabeled (general cohort) & $1{,}287$ & $5{,}354$ & $6{,}641$\\
\midrule
\multicolumn{4}{l}{\textit{Demographics, mean (SD) or \%}}\\
\quad Age at scan (years) & $72.9\,(7.8)$ & $71.2\,(8.5)$ & $71.6\,(8.4)$\\
\quad Female (\%) & $49.0$ & $59.6$ & $57.2$\\
\quad Education (years) & $16.3\,(2.6)$ & $16.2\,(2.7)$ & $16.2\,(2.7)$\\
\quad APOE-$\varepsilon$4 carriers (\%) & $43.8$ & $39.1$ & $40.2$\\
\midrule
\multicolumn{4}{l}{\textit{Modality availability (\% of cohort with $\ge 1$ feature observed)}}\\
\quad Amyloid PET & $72.6$ & $34.0$ & $42.9$\\
\quad Tau PET & $16.9$ & $15.4$ & $15.7$\\
\quad Structural MRI & $79.9$ & $70.9$ & $73.0$\\
\quad Plasma biomarkers & $58.0$ & $0.0$ & $13.3$\\
\bottomrule
\end{tabularx}

\smallskip
\footnotesize $^{\dagger}$ NC startpoint pool: cognitively normal subjects (diagnosis CN, amyloid-negative, age $\in [40, 65]$, \cameraready{CDR\textsuperscript{\textregistered} Dementia Staging Instrument Sum of Boxes (CDR-SB)}${=}0$) pinned at $s\!\approx\!0$ to supply the $\bm{\Theta}_{c,0}$ healthy-reference prior, excluded from the unlabeled row.
\end{table}

\subsection{Prior specifications}\label{app:ad_priors}
The boundary supervision priors \cref{eq:anchor,eq:control} use $s_A=0.95$, $s_H=0.01$, and $\epsilon_S=0.1$ throughout. 
For the AD experiments, the diffuse priors on the modality-level onset centers $\{\tilde t_{c,m}\}$ in \cref{eq:vel-prior-2} and on the age coordinate of $\bm{b}_c$ in \cref{eq:subtypestageprior} are replaced by informative location and ordering priors. \emph{(a) AD biomarker cascade.} For $c\in\{\mathrm{EOAD},\mathrm{LOAD}\}$, the modality-level centers are softly ordered amyloid-PET $\to$ tau-PET $\to$ structural-MRI through a Gaussian penalty $\sum_{m<m'}\bigl(\max(0,\tilde t_{c,m}-\tilde t_{c,m'})\bigr)^2/\sigma_{\mathrm{ord}}^2$ with $\sigma_{\mathrm{ord}}{=}0.1$, encoding the NIA-AA cascade \cite{Jack2024}. \emph{(b) Preserved-aging endpoint.} For the HA subtype, $\tilde t_{\mathrm{HA},m}\sim\mathcal{N}(\cameraready{\mathrm{logit}\,0.9}, 0.05^2)$ for $m\in\{\text{Amyloid PET}, \text{Tau PET}\}$, placing pathology onsets near the endpoint. \emph{(c) Subtype age priors.} The age coordinate of $\bm{b}_c$ is drawn from $\mathcal{N}(-2.0,0.5)$ for EOAD, $\mathcal{N}(+1.5, 0.5)$ for LOAD, and $\mathcal{N}(-0.5, 0.5)$ for HA, with the remaining coordinates keeping $\mathcal{N}(0,1)$. The synthetic experiments use the diffuse $\bm{b}_c\equiv \bm{0}$ and the unconditional categorical $\bm{\alpha}\!\sim\!\mathcal{N}(\bm{0},\bm{I}_C)$ on subtype.

\subsection{Competing methods and runtime}\label{app:methods}
The external baselines (\cref{tab:synth,tab:stage}) and internal variants (\cref{tab:abl_synth,tab:abl_real}) are introduced in \cref{sec:synth,sec:real_results}. \Cref{tab:runtime} reports per-fit wall-clock time and training sample size on the synthetic and real-data benchmarks. The intended read of the table is end-to-end cost for a single posterior fit, so the \textsc{Elpac} numbers cover both phases of the curriculum optimizer of \cref{app:inference} (initialization with diagonal covariance plus the joint phase under the rank-$K$-plus-diagonal covariance) and the test-time encoder pass.

\textbf{Hardware.} \textsc{Elpac} is fit on a single NVIDIA RTX Pro 6000 Blackwell GPU (48~GB VRAM), with $2$ CPU cores and $6$~GB host memory per task. CPU baselines run on shared partitions of mixed Intel Xeon and AMD EPYC nodes ($64$ to $128$ cores per node, up to $2.3$~TB RAM per node), with each task allocated $4$ CPU cores and $16$~GB RAM (XGBoost and Label-prop use $2$ cores and $8$~GB).

\begin{table}
\centering\small
\caption{\textbf{\textsc{Elpac} fits the full multimodal cohort jointly within wall-clock budgets comparable to the joint-fusion baselines.} Hardware, sample size, and per-fit wall-clock (seconds, mean${\pm}$SE) over $100$ synthetic replicates and $20$ AD splits. $n$ is the total sample size (train${+}$held-out). Supervised label-consuming baselines (XGBoost, Label-prop) operate on the labeled anchor partition, others on the full cohort.}
\label{tab:runtime}
\setlength{\tabcolsep}{4pt}
\begin{tabularx}{\linewidth}{X l r r r r}
\toprule
& & \multicolumn{2}{c}{Synthetic} & \multicolumn{2}{c}{Real}\\
\cmidrule(lr){3-4}\cmidrule(lr){5-6}
Method & Device & $n$ & Time (s) & $n$ & Time (s)\\
\midrule
SuStaIn~\cite{Young2018}                & CPU & $1{,}000^{\dagger}$ & $3{,}089.28{\pm}49.67$ & $7{,}501^{\dagger}$ & $29{,}707.17{\pm}951.63$\\
DEBM~\cite{Venkatraghavan2019}          & CPU & $1{,}000^{\dagger}$ & $7.21{\pm}0.15$        & $7{,}501^{\dagger}$ & $55.34{\pm}6.60$\\
Static-FA~\cite{AndersonRubin1956,Mardia1979}      & CPU & $1{,}000^{\dagger}$ & $0.78{\pm}0.06$        & $7{,}501^{\dagger}$ & $14.18{\pm}1.14$\\
MoPoE-VAE~\cite{Sutter2021MoPoE}        & CPU & $1{,}000$   & $6.30{\pm}0.20$        & $7{,}501$ & $5.55{\pm}0.32$\\
XGBoost~\cite{Chen2016XGBoost}          & CPU & ${\sim}150$ & $0.90{\pm}0.01$        & $641$     & $1.20{\pm}0.06$\\
Label-prop~\cite{Zhu2002LabelProp}      & CPU & ${\sim}150$ & $0.03{\pm}0.00$        & $641$     & $102.91{\pm}0.91$\\
HEALNet~\cite{Hemker2024}               & CPU & $1{,}000$   & $258.11{\pm}5.54$      & $7{,}501$ & $926.95{\pm}175.06$\\
\midrule
Diagonal covariance                     & GPU & $1{,}000$ & $60.77{\pm}0.56$       & $7{,}501$ & $198.13{\pm}0.62$\\
Static-factor covariance                & GPU & $1{,}000$ & $114.66{\pm}0.64$      & $7{,}501$ & $511.27{\pm}50.92$\\
\midrule
\textbf{\textsc{Elpac} (full)}          & GPU & $1{,}000$ & $167.25{\pm}1.67$      & $7{,}501$ & $384.61{\pm}1.39$\\
\bottomrule
\end{tabularx}

\smallskip
\footnotesize $^{\dagger}$ SuStaIn, DEBM, and Static-FA are fit per biomarker on observed cells; reported $n$ is the cohort size, while the effective per-feature training set ranges from ${\sim}1{,}000$ (plasma) to ${\sim}5{,}500$ (sMRI) on the real cohort. MoPoE-VAE, HEALNet, and \textsc{Elpac} variants are fit jointly on all modalities.
\end{table}

\textsc{Elpac} fits the full $7{,}501$-subject AD cohort in ${\sim}385$~s on a single GPU, ${\sim}77{\times}$ faster than SuStaIn and ${\sim}2.4{\times}$ faster than HEALNet, and returns a posterior over subtype, stage, trajectories, and stage-varying coordination. Two implementation choices keep the cost in this range: the structured low-rank guide of \cref{eq:guide} makes the global-parameter cost linear in $r_g{=}10$, and the Sylvester--Woodbury reduction of \cref{eq:sw} keeps each per-stage likelihood evaluation at $O(RK^2{+}K^3)$ regardless of the missingness pattern. The $187$~s gap between the diagonal-covariance variant and \textsc{Elpac} (full) buys the held-out NLL and masked reconstruction gains in \cref{tab:abl_real}.

\subsection{Metric definitions}\label{app:metrics}
We define the non-standard metrics here. Standard ones (BAcc, Macro-F1, AUC, RMSE) follow conventional usage. \emph{NLL/cell} is the held-out negative log-likelihood per observed coordinate, normalized by the average number of observed cells per subject for cross-model comparability. \emph{Masked NLL} averages cross-modal predictive log-likelihood across four leave-one-modality-out rotations. \emph{Stage-score $\rho$} is the sign-aware Spearman correlation between the recovered stage scalar and the simulator stage, while $|\rho|$ vs.\ MMSE uses the absolute value to be sign-agnostic across methods. \emph{Cog Pearson $r$} is the held-out Pearson correlation between observed cognition $C_{\rm cog}$ and a regression of $C_{\rm cog}$ on $(\hat S_i, \mathrm{onehot}(\hat c_i))$ fit on training anchors, with absolute-value variants reported when subtype-conditional sign conventions differ across methods. \emph{Stage-binned PPC} is the cross-pair correlation between the model-implied $\rho_c(r,r';s)$ and the empirical $\hat\rho_c(s)$ binned over five stage bins. \emph{$P\%$ PI coverage} is the empirical fraction of held-out biomarker observations inside their nominal $P\%$ posterior predictive interval. AUC paired bootstraps use $200$ resamples, while all other paired-bootstrap CIs use $1{,}000$.

\subsection{Synthetic recovery: subtype trajectories and stage-varying coordination}\label{app:synth_recovery_traj_nu}
\begin{figure}
\centering
\includegraphics[width=\linewidth]{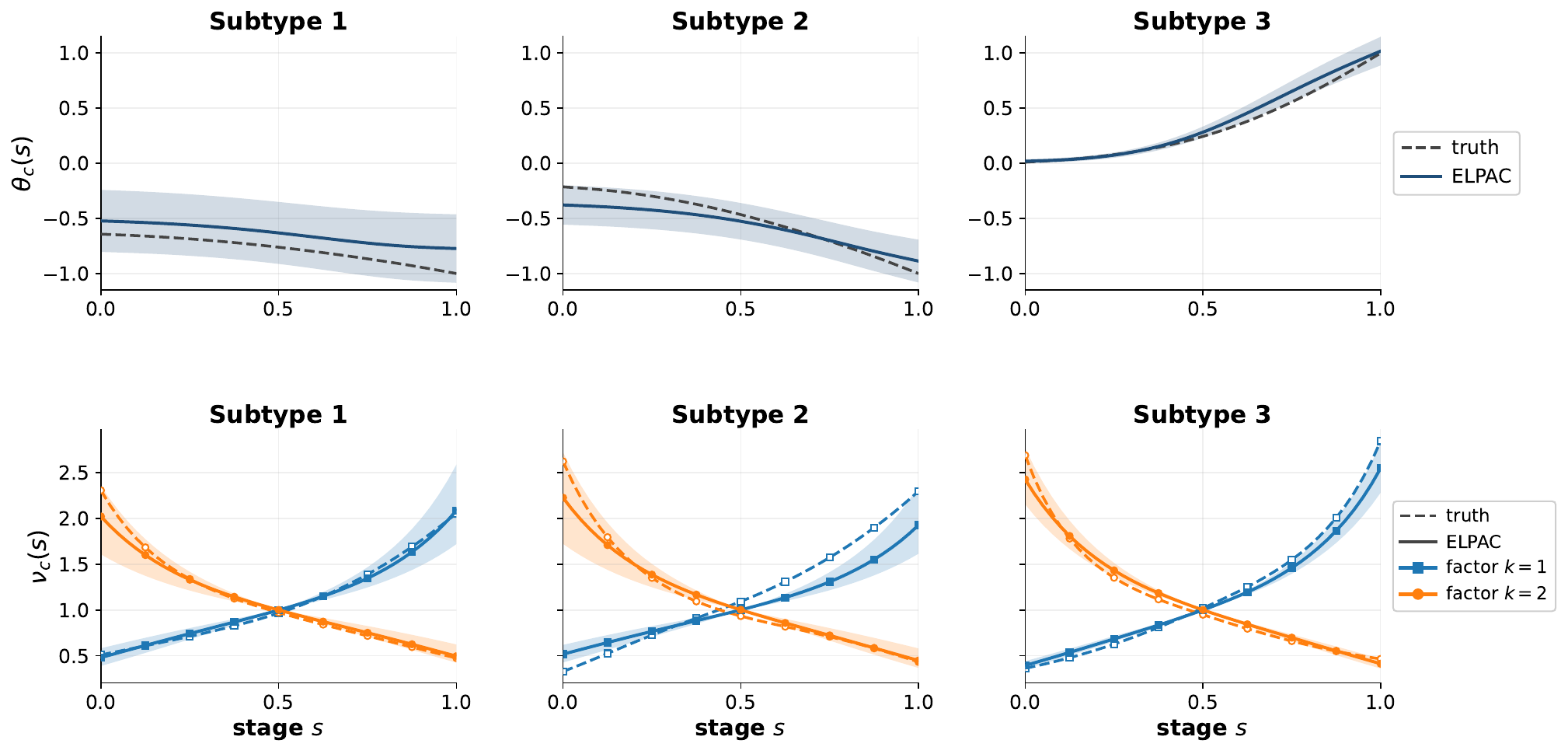}
\caption{\textbf{Subtype trajectories $\bm{\theta}_c(s)$ and stage-varying activations $\bm{\nu}_c(s)$ are recovered against simulator truth.} \textbf{Top row}: representative $\bm{\theta}_c(s)$, one biomarker per subtype, amplitude-normalized; truth dashed, \textsc{Elpac} solid; shaded band is the across-replicate $95\%$ CI of the mean over $100$ replicates. \textbf{Bottom row}: $\bm{\nu}_c(s)$ for the two factors retained by the multiplicative-gamma prior; per-factor color and marker; truth dashed, \textsc{Elpac} solid; shaded interquartile range over $100$ replicates.}
\label{fig:appH_synth_traj_nu}
\end{figure}

\Cref{fig:appH_synth_traj_nu} probes the two parameter blocks targeted by Constructions~I and~II separately, against the simulator truth, on $100$ replicate cohorts at $N{=}1{,}000$. Construction~I is summarized through the recovered subtype mean trajectory $\bm{\theta}_c(s)$ on a representative biomarker per subtype (top row), and Construction~II through the stage-varying activation $\bm{\nu}_c(s)$ for the factors retained by the multiplicative-gamma prior (bottom row).

\textsc{Elpac}'s recovered subtype trajectories overlay the simulator truth across the full stage range (top row), and the across-replicate $95\%$ CI on the posterior mean stays tight in the unsupervised middle region between the boundary anchors. The endpoint-anchored backward recursion \cref{eq:backward} therefore propagates the boundary supervision inward and preserves amplitude at intermediate stages. The retained activation profiles $\bm{\nu}_c(s)$ (bottom row) recover stage-localized peaks per subtype that match the simulator-truth dashed curves, and the multiplicative-gamma cascade shrinks the third loading column of $\bm{A}_c$ toward zero on every replicate. Construction~I (mean trajectory) and Construction~II (factor activations) are jointly recovered under the variational family of \cref{eq:guide}, matching the conditional identifiability claim of \cref{thm:identifiability}.

\subsection{Sensitivity analysis}\label{app:sensitivity}
\begin{figure}
\centering
\includegraphics[width=0.8\linewidth]{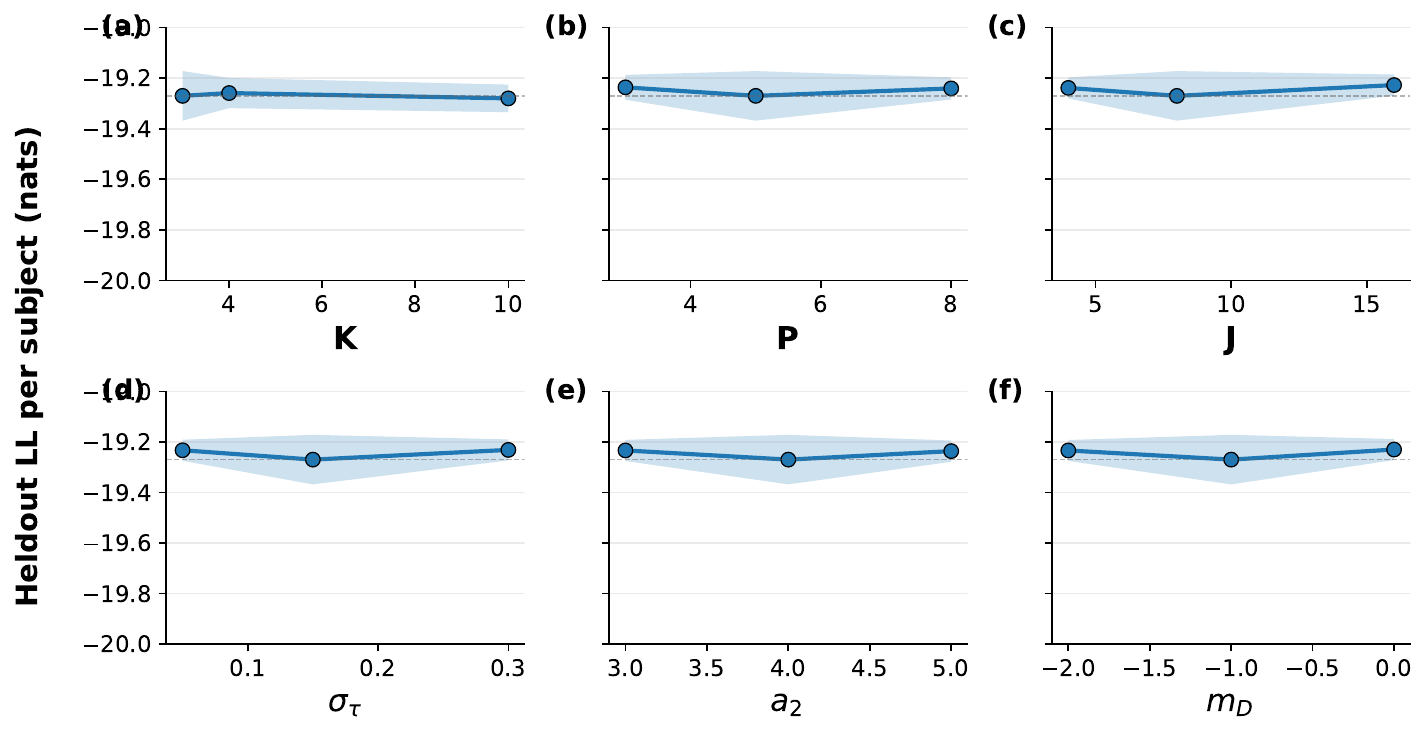}
\caption{\textbf{\textsc{Elpac} is robust to factor-module hyperparameters.} Held-out log-likelihood per subject across one-at-a-time sweeps of the six hyperparameters around the canonical configuration, with bands at $\pm 1.96\,\mathrm{SE}$ over $50$ paired replicate cohorts per cell.}
\label{fig:appH_sensitivity}
\end{figure}

\begin{figure}
\centering
\includegraphics[width=\linewidth]{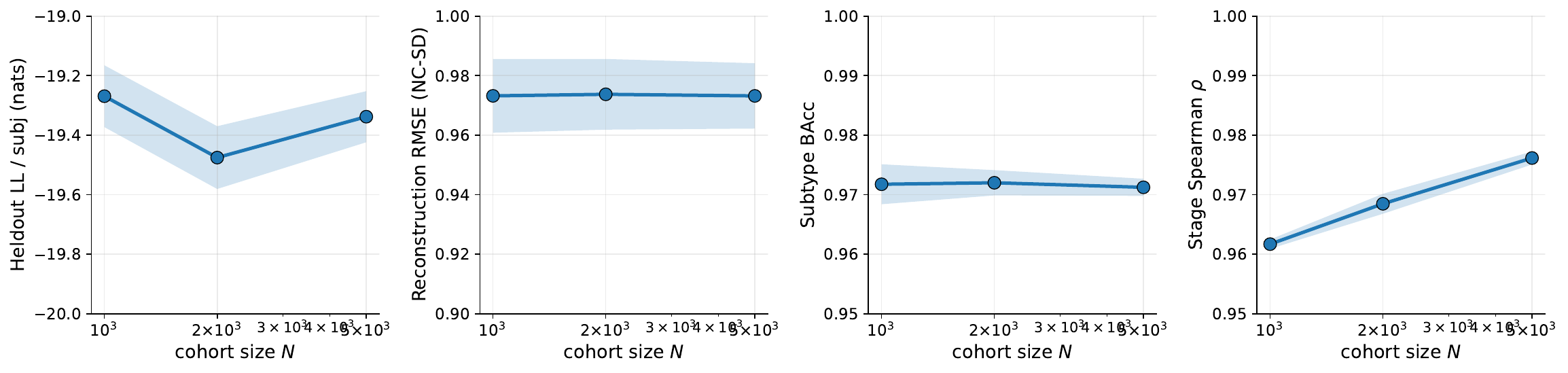}
\caption{\textbf{Latent-structure recovery is stable across $N\in\{1{,}000,\,2{,}000,\,5{,}000\}$.} Sample-size sweep: held-out log-likelihood, reconstruction RMSE, BAcc, and stage Spearman $\rho$, with bands at $\pm 1.96\,\mathrm{SE}$ over $100$ paired replicate cohorts per cell.}
\label{fig:appH_sample_size_100}
\end{figure}

\Cref{fig:appH_sensitivity,fig:appH_sample_size_100} report the hyperparameter and sample-size sweeps that support the sensitivity claims in §\ref{sec:synth}. Each sweep is paired across replicate cohorts, so the reported standard errors absorb cohort-level variability and isolate the marginal effect of the hyperparameter under variation.

\Cref{fig:appH_sensitivity} sweeps the factor rank $K$, the activation-spline basis dimension $P$, the trajectory shell count $J$, the activation-prior scale $\sigma_\tau$, the multiplicative-gamma cascade strength $a_2$, and the log-noise mean $m_D$ one axis at a time around the canonical configuration. \textsc{Elpac}'s held-out log-likelihood per subject stays inside the $\pm 1.96\,\mathrm{SE}$ band of the canonical value on every panel, so the per-construction attribution in \cref{tab:abl_synth,tab:abl_real} is robust to the hyperparameter choice.

\cameraready{We have also validated on synthetic data that the coordination structure is not driven by the priors: the informative NIA-AA prior acts on the mean-trajectory onset centers alone and \cameraready{does not act directly on} the coordination block, and relaxing the priors that shape it leaves the factorization essentially unchanged (Tucker congruence of matched loading columns $\phi{=}0.967$ across the $a_2$ and $\sigma_\tau$ sweeps of \cref{fig:appH_sensitivity}, against $0.29$ between fits on different simulated cohorts under the same prior). It follows the data instead---there the model-implied stage-varying correlations track the empirically binned ones (\cref{tab:abl_synth}), and qualitatively so on the real cohort (\cref{fig:supp_biology}c).}

\textsc{Elpac}'s four recovery metrics in \cref{fig:appH_sample_size_100} (held-out log-likelihood, reconstruction RMSE, subtype BAcc, stage Spearman $\rho$) are stable across $N\in\{1{,}000, 2{,}000, 5{,}000\}$, so the $N{=}1{,}000$ numbers in \cref{tab:synth} hold across the full sample-size range we test.

\begin{table}[t]
\centering\small\ifcameraready\color{blue}\fi
\caption{\cameraready{\textbf{Paired change from clean anchors} (pp, or nats/subject for LL; mean$\pm$SE). Real-cohort changes are paired against each split's own clean-anchor refit.}}
\label{tab:anchor_corruption}
\setlength{\tabcolsep}{4pt}
\begin{tabular}{l l l c c c}
\toprule
Scenario & Data & Metric & $\Delta$ 5\% & $\Delta$ 10\% & $\Delta$ 20\%\\
\midrule
Mislabel, clinical confusion & Synthetic & Subtype acc & $-0.2{\pm}0.1$ & $-0.6{\pm}0.1$ & $-1.3{\pm}0.1$\\
 &  & Stage $\rho$ & $-0.5{\pm}0.2$ & $-1.0{\pm}0.2$ & $-1.9{\pm}0.3$\\
 & Real & Subtype BAcc & $-1.7{\pm}0.7$ & $-4.2{\pm}1.2$ & $-6.4{\pm}1.3$\\
 &  & Macro-F1 & $+1.8{\pm}1.5$ & $+1.0{\pm}2.5$ & $+0.6{\pm}2.8$\\
 &  & Held-out LL & $-0.02{\pm}0.01$ & $-0.02{\pm}0.01$ & $-0.05{\pm}0.02$\\
\midrule
Mislabel, uniform-random & Synthetic & Subtype acc & $-0.1{\pm}0.1$ & $-0.6{\pm}0.2$ & $-3.4{\pm}0.6$\\
 &  & Stage $\rho$ & $-0.7{\pm}0.2$ & $-1.7{\pm}0.2$ & $-5.7{\pm}0.4$\\
 & Real & Subtype BAcc & $-2.2{\pm}0.8$ & $-5.1{\pm}1.0$ & $-7.0{\pm}1.2$\\
 &  & Macro-F1 & $-0.8{\pm}1.4$ & $-0.8{\pm}1.9$ & $+1.3{\pm}2.5$\\
 &  & Held-out LL & $-0.02{\pm}0.02$ & $-0.06{\pm}0.02$ & $-0.09{\pm}0.02$\\
\midrule
Stage redrawn $\mathcal{U}(0,1)$ & Synthetic & Stage $\rho$ & $-2.2{\pm}0.2$ & $-2.9{\pm}0.3$ & $-3.3{\pm}0.3$\\
 &  & Subtype acc & $+0.0{\pm}0.1$ & $-0.0{\pm}0.1$ & $-0.0{\pm}0.1$\\
\midrule
Inverted group-mean direction & Synthetic & Stage $\rho$ & $-0.2{\pm}0.1$ & $-0.6{\pm}0.2$ & $-1.3{\pm}0.3$\\
 &  & Subtype acc & $-0.0{\pm}0.1$ & $-0.1{\pm}0.2$ & $-1.3{\pm}0.5$\\
\midrule
Localized non-monotone bump & Synthetic & Stage $\rho$ & $-0.0{\pm}0.1$ & $-0.0{\pm}0.1$ & $-0.0{\pm}0.1$\\
 &  & Subtype acc & $-0.0{\pm}0.0$ & $-0.0{\pm}0.1$ & $-0.1{\pm}0.1$\\
\bottomrule
\end{tabular}
\end{table}
\subsection{\cameraready{Robustness to anchor corruption}}\label{app:anchor_corruption}

{\ifcameraready\color{blue}\fi
The monotonicity constraint applies to the latent group-mean trajectory, not to individual observations, which are noisy realizations of it, with subject-wise heterogeneity in the residual factor layer. A subject whose measurements stabilize or reverse is variation \emph{about} the mean, not a constraint on its shape. On that mean, \cref{eq:velocity} fixes the sign of the shell-endpoint velocities, and so the direction of shell-to-shell progression, leaving magnitude free; the cubic Hermite interpolation retains limited freedom for between-shell overshoot. It also admits flat segments, all three within \cref{eq:velocity}: the onset sigmoid holds a marker at rest before onset, the spline amplitude $F_{c,r}$ can fall arbitrarily close to zero over any stage range, and $\delta_{c,r}{=}0$ drops a marker outright---so temporary stabilization, plateaus and saturation all sit inside the family, as the biology of these markers requires. That direction is established for them~\cite{Jack2024} and is the inductive bias event-based staging relies on~\cite{Fonteijn2012,Young2018}. Thus the only shape excluded is a sustained reversal of a group mean---which we inject deliberately below.

We corrupt anchor correctness at graded levels along four axes: subtype mislabeling, both uniform-random and under a clinical confusion scheme that concentrates errors on the EOAD/LOAD boundary where diagnostic ambiguity actually falls; the anchor's true stage redrawn; an inverted group-mean direction; and a localized non-monotone bump. We refit at each level. Synthetic replicates run against known truth, and the mislabeling arms are mirrored on the real cohort (\cref{tab:anchor_corruption}).

Across the four targeted corruptions, \textsc{Elpac} retains stage ordering and subtype separation. Anchor mislabeling degrades subtype and stage recovery smoothly as the labels get noisier, with clinical confusion the gentler of the two arms at 20\%; the stage redraw affects staging while leaving subtype recovery unchanged; and the two mean-shape perturbations have the smallest stage-ordering effects---inverting a marker's direction is milder in stage ordering than any mislabeling, and a localized non-monotone bump produces no detectable loss in stage ordering. The losses are also small: on the real cohort, held-out likelihood stays well inside the $0.84$-nat full-vs-static-factor margin of \cref{tab:abl_real}, and Macro-F1 shows no monotone degradation.
}

\subsection{\cameraready{Test-time refinement}}\label{app:refinement}
\begin{table}[tb]
\centering\small\ifcameraready\color{blue}\fi
\caption{\cameraready{\textbf{Per-subject refinement sharpens held-out staging at every step budget.} Relative change vs.\ the single encoder pass ($T{=}0$), globals at their posterior medians; mean over replicates (synthetic) or splits (real), $\pm$SE at $T{=}50$. Synthetic: stage $\rho$ against the true stage; real: $|\rho|$ of the path-aggregated severity score $G^{\mathrm{path}}$ (\cref{app:longitudinal}) against MMSE on all refined held-out subjects.}}
\label{tab:refinement}
\setlength{\tabcolsep}{5pt}
\begin{tabular}{l c c c c c c}
\toprule
$T$ Adam steps & 1 & 2 & 5 & 10 & 20 & 50\\
\midrule
Synthetic (100 replicates) & $+0.5\%$ & $+1.0\%$ & $+2.5\%$ & $+4.7\%$ & $+8.3\%$ & $+13.0{\pm}0.5\%$\\
Real cohort (20 splits) & $+1.6\%$ & $+2.9\%$ & $+7.1\%$ & $+10.9\%$ & $+13.1\%$ & $+13.6{\pm}2.7\%$\\
\bottomrule
\end{tabular}
\end{table}

\cameraready{Subject-specific refinement is a drop-in extension of the amortized inference path of \crefrange{eq:guide}{eq:posterior_joint}. The mechanism is the local-refinement step of semi-amortized inference~\citechk{Kim2018SemiAmortized,Marino2018Iterative}: with the fitted globals frozen at their posterior medians, each subject's subtype-conditional stage posteriors are warm-started from the encoder pass and optimized on that subject's own data---no labels, no retraining, no global update. Subtype remains exactly enumerated; during refinement its posterior is computed from the per-subtype terms of the local bound, which approach \cref{eq:posterior_c} as those stage posteriors sharpen; the factor scores remain analytically integrated. The local objective is a variational lower bound on that subject's evidence under the same frozen globals and $G{=}19$ grid. The step ladder $T\in\{1,2,5,10,20,50\}$ (Adam steps) and learning rate $0.05$ were fixed in advance.}

\cameraready{Across the tested budgets the per-subject objective improves and stage $\rho$ rises above the encoder pass (\cref{tab:refinement}); the gain at $T{=}50$ holds in every synthetic replicate and persists when results are averaged over 20 global-posterior draws. Subtype inference stays stable, matching the frozen-global exact-grid reference on synthetic data and approaching it from below on the real cohort. The measured gain is therefore in staging, with subtype inference stable, giving an optional per-patient accuracy-for-compute tradeoff.}

\FloatBarrier
\section{Real-data validation and biology}\label{app:real_validation}
\subsection{Leave-one-cohort-out validation}\label{app:loco}
\begin{figure}
\centering
\includegraphics[width=\linewidth]{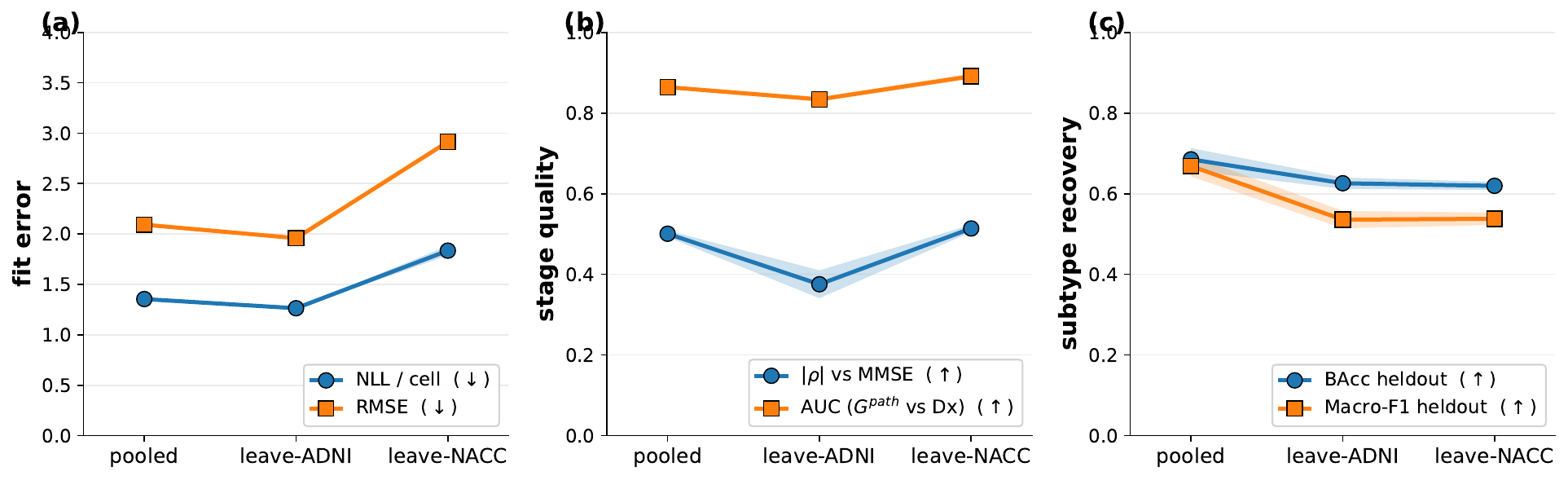}
\caption{\textbf{Stage and subtype recovery transfer across cohorts.} Leave-one-cohort-out validation. \textbf{(a)} Fit error (NLL/cell, RMSE). \textbf{(b)} Stage quality ($|\rho|$ vs.\ MMSE, AUC of $G^{\mathrm{path}}$ vs.\ diagnosis). \textbf{(c)} Subtype recovery (BAcc, Macro-F1). Lines connect pooled, leave-ADNI, and leave-NACC folds; bands are $\pm 1.96\,\mathrm{SE}$ over $20$ splits.}
\label{fig:appH_loco}
\end{figure}

\Cref{fig:appH_loco} compares the pooled fit with leave-ADNI (train on NACC, evaluate on ADNI) and leave-NACC (train on ADNI, evaluate on NACC). ADNI and NACC differ in scanner pool, plasma availability ($58\%$ vs.\ $0\%$, \cref{tab:cohort}), and recruitment composition, so the leave-one-cohort folds test cross-source transfer of the constructions.

The stage-quality and subtype-recovery panels (\cref{fig:appH_loco}b,c) overlap across the three folds within their $\pm 1.96\,\mathrm{SE}$ bands, so the anchored stage axis and its subtype mapping transfer across acquisition sources. The small NLL/RMSE increase under leave-NACC (panel~a) tracks the smaller ADNI-only training pool (${\sim}1.7$k vs.\ ${\sim}5.8$k subjects) and does not propagate into the stage-quality or subtype-recovery panels.

\subsection{Longitudinal validation: stage velocity, cognition, and MCI conversion}\label{app:longitudinal}
\begin{figure}
\centering
\includegraphics[width=\linewidth]{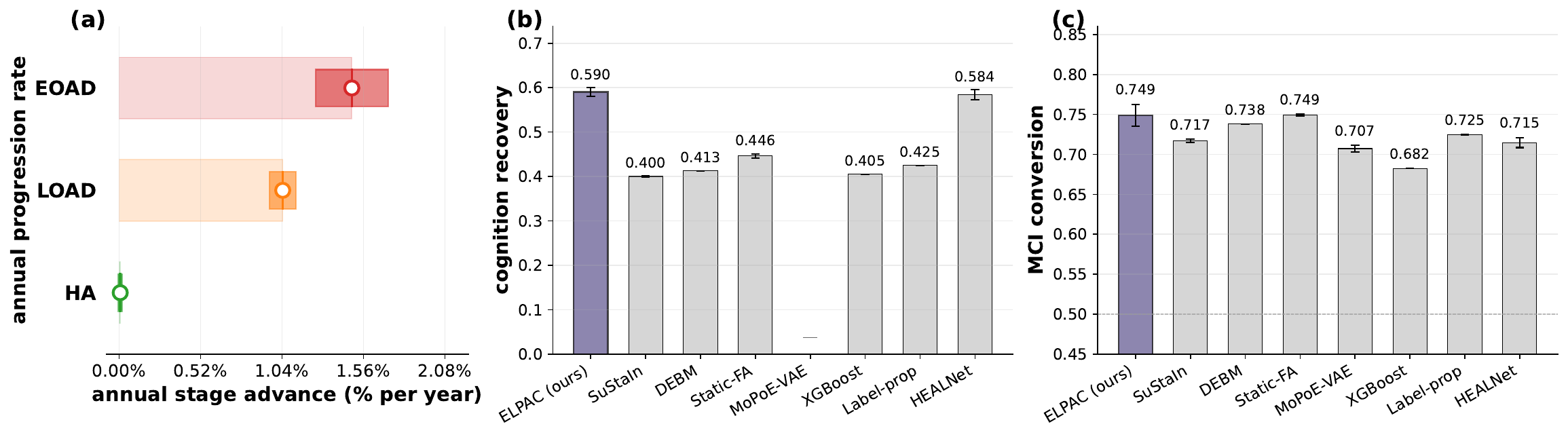}
\caption{\textbf{The recovered stage tracks longitudinal progression and predicts MCI$\to$AD conversion at the top of the external-baseline set.} \textbf{(a)} Annual stage advance per subtype. \textbf{(b)} Held-out cognition correlation across the \cref{tab:stage} comparison set. \textbf{(c)} Three-year MCI$\to$AD conversion AUC across the same set. \textsc{Elpac} highlighted, with bars at $\pm 1.96\,\mathrm{SE}$.}
\label{fig:appH_longitudinal}
\end{figure}

\Cref{fig:appH_longitudinal} reports three longitudinal probes of the cross-sectional stage axis on subjects with follow-up visits: per-subject annual stage advance by subtype (panel~a), held-out cognition correlation across the \cref{tab:stage} comparison set (panel~b), and three-year MCI-to-AD conversion AUC on the same set (panel~c). Panels~a and~c target signals absent from the training likelihood, so they read out longitudinal behavior of the recovered stage axis.
The annual stage advance orders the subtypes EOAD${>}$LOAD${>}$HA along a temporal axis unobserved at training, matching the clinical expectation that early-onset disease progresses fastest and healthy aging is approximately stable~\cite{Jack2024}. \textsc{Elpac} leads panel~b, so the inferred stage is the strongest single predictor of cognition among the methods evaluated. The path-aggregated severity score $G^{\mathrm{path}}_i$, defined as the integrated mean-trajectory pathology up to subject $i$'s posterior stage, leads panel~c, ranking short-horizon conversion risk on subjects unseen at training.

\begin{figure}
\centering
\includegraphics[width=\linewidth]{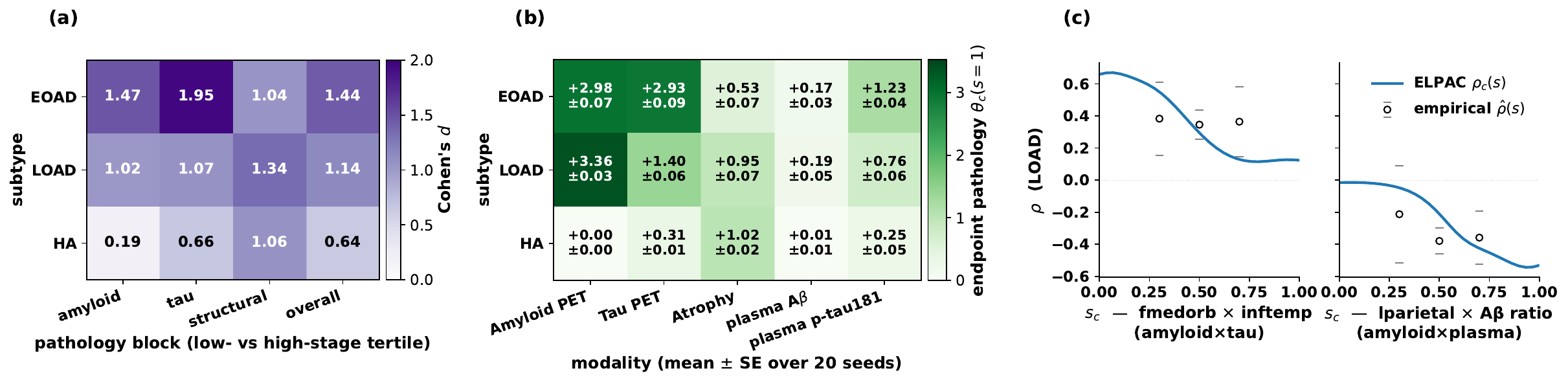}
\caption{\textbf{Each subtype separates biomarkers along its dominant pathology channel.} \textbf{(a)} Stage discriminability (Cohen's $d$, tertile contrast) by subtype and pathology block. \textbf{(b)} Endpoint pathology heatmap: NC-anchored $\bm{\theta}_c(s{=}1)$ (\cameraready{$\pm\mathrm{SE}$}, $20$ splits). \textbf{(c)} Two LOAD cross-modal correlation trajectories $\rho_c(r,r';s)$: model-implied curve vs.\ empirical-binned $\hat\rho_c(s)$.}
\label{fig:supp_biology}
\end{figure}

\subsection{Subtype biology and coordination programs}\label{app:subtype_bio}
\Cref{fig:supp_biology} reports three biological summaries of the recovered subtype structure. The discriminability heatmap (panel~a) localizes the largest effect sizes to \cameraready{the tau pathology block} in EOAD, structural atrophy in LOAD, and structural-only signal in HA, mirroring the modality-direction pattern in \cref{fig:stage}. The endpoint pathology heatmap (panel~b) shows HA at the healthy reference on amyloid while EOAD and LOAD reach pathological endpoints on every modality, so the inferred trajectories terminate at the biologically expected magnitudes per subtype.

Panel~c overlays the model-implied stage-dependent correlation $\rho_c(r,r';s)$ from \cref{eq:rho_closed} on the empirical-binned $\hat\rho_c(s)$ for two representative LOAD biomarker pairs, and the two curves track each other across the inferred stage axis on both pairs. The fit captures the stage shape, not only the stage-marginal value, consistent with the stage-binned PPC ablation gain in \cref{tab:abl_synth}.
\begin{figure}
\centering
\includegraphics[width=0.8\linewidth]{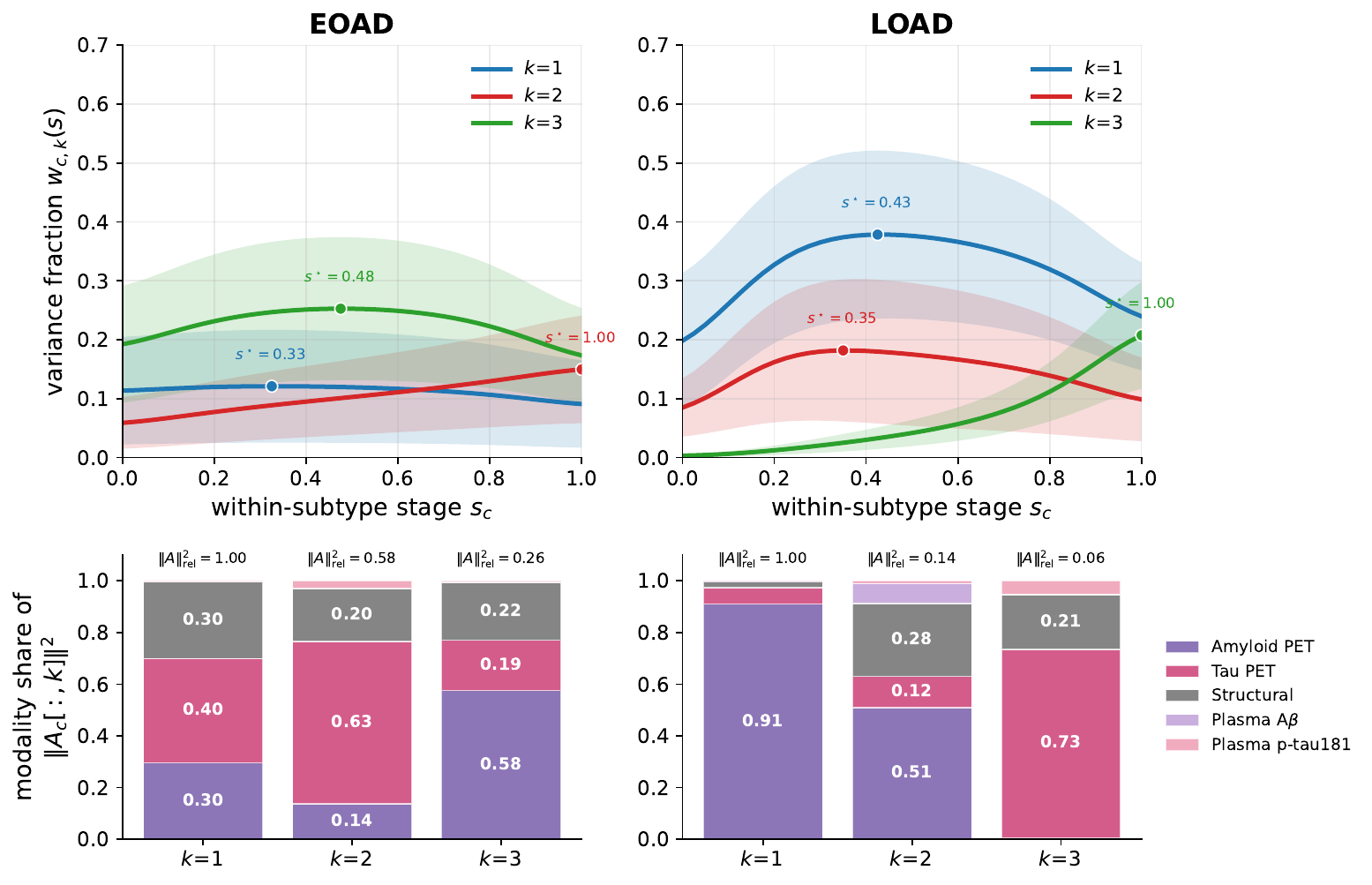}
\caption{\textbf{Subtype-specific coordination structures emerge: EOAD is concurrent multimodal whereas LOAD is sequential amyloid-then-tau.} Top: stage-varying variance share $\Phi_{c,k}(s_c)$ for the three leading factors, with peak markers and $\pm 1.96\,\mathrm{SE}$ bands. Bottom: modality decomposition of loading energy (sum $=1$ per bar). HA omitted; see text.}
\label{fig:appH_factor_multimodal}
\end{figure}
\paragraph{Per-factor multimodal mixture (cross-modal coupling).}
\Cref{fig:appH_factor_multimodal} resolves the residual coordination structure into per-factor variance shares and modality decompositions of loading energy, addressing \emph{which biomarkers couple in each subtype, and when each coalition strengthens or attenuates}. EOAD engages a three-way amyloid, tau, and structural coalition (across-modality) with mid-stage peak. LOAD instead exhibits an \cameraready{amyloid-dominant} leading factor (within-modality) that peaks earlier on the stage axis, with a smaller late-emerging tau-dominant factor (within-modality), \cameraready{consistent with} the sequential amyloid-then-tau cascade reported in late-onset disease. HA is omitted: \cameraready{its leading factors are atrophy-dominant within-modality axes (\cref{sec:real_results}) carrying no pathology-discriminative signal}, as expected for an amyloid-negative endpoint.

\begin{figure}
\centering
\includegraphics[width=\linewidth]{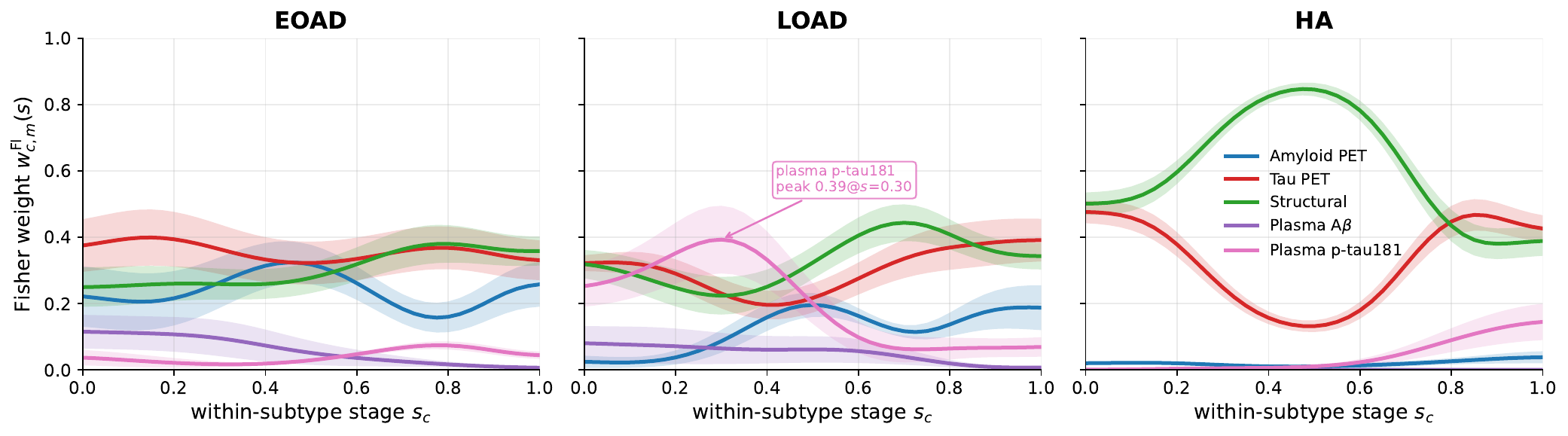}
\caption{\textbf{The most stage-informative modality varies by subtype: tau-PET/sMRI in EOAD, plasma \cameraready{p-tau181} then tau-PET in LOAD, sMRI throughout in HA.} Per-subtype Fisher modality weights $w^{\mathrm{FI}}_{c,m}(s_c)$, plasma split into A$\beta$ ratio and \cameraready{p-tau181}; bands $\pm 1.96\,\mathrm{SE}$ over $20$ splits.}
\label{fig:appH_fisher_modality}
\end{figure}

\paragraph{Stage-varying Fisher modality weights.}
A complementary view targets \emph{which modality is most informative about stage at a given stage}, derived from the mean-trajectory derivative $\bm{\theta}'_c(s)$ and the inverse covariance (\cref{eq:fi_marg,eq:fi_share}). \Cref{fig:appH_fisher_modality} plots the per-subtype Fisher modality weights $w^{\mathrm{FI}}_{c,m}(s_c)$ on the $5$-modality split (plasma A$\beta$ ratio and \cameraready{p-tau181} separated). In EOAD, tau-PET and structural MRI jointly carry the bulk of the stage information across mid-to-late stage, indicating that tau and structural readouts are the most informative modalities in early-onset disease. In LOAD, plasma \cameraready{p-tau181} reaches a mid-stage informativeness peak before tau-PET takes over near the endpoint, \cameraready{consistent with reports that plasma p-tau leads} PET-detectable tau in amyloid-positive individuals~\cite{Janelidze2021}, so plasma readouts are the most stage-informative biomarkers earlier in late-onset disease. In HA, structural MRI dominates throughout, indicating that healthy aging is a structural-only progression direction without informative amyloid or tau signal.

\newpage
\section*{NeurIPS Paper Checklist}

\begin{enumerate}

\item {\bf Claims}
    \item[] Question: Do the main claims made in the abstract and introduction accurately reflect the paper's contributions and scope?
    \item[] Answer: \answerYes{}
    \item[] Justification: The abstract and the three numbered contributions in §1 (asymmetric endpoint supervision, conditional identifiability of stage-varying coordination, validation on synthetic and harmonized AD cohorts) match the experimental claims of §3.1 and §3.2 and the conditional-identifiability claim of Theorem~\ref{thm:identifiability} in §2.5.

\item {\bf Limitations}
    \item[] Question: Does the paper discuss the limitations of the work performed by the authors?
    \item[] Answer: \answerYes{}
    \item[] Justification: §4 Discussion contains an explicit \emph{Limitations} paragraph covering \cameraready{four} items: \cameraready{the stage axis is pseudo-temporal, so the coordination factors describe co-variation along it rather than mechanism;} the Gaussian dependency layer captures only second-order coordination; the full-model identifiability proposition (\cref{app:ident_warp_prop}) is conditional on a Gaussian mixing-family hypothesis; and variational inference is an approximation to the true posterior. \cameraready{Each of the last three} is paired with a natural extension path.

\item {\bf Theory assumptions and proofs}
    \item[] Question: For each theoretical result, does the paper provide the full set of assumptions and a complete (and correct) proof?
    \item[] Answer: \answerYes{}
    \item[] Justification: Theorem~\ref{thm:identifiability} is stated in §2.5 with conditions and a four-step proof sketch; the full proof appears in \cref{app:thm_proof}. Proposition~1 (full-model identifiability) is stated and proved in \cref{app:ident_warp_prop}. All theorems are numbered and cross-referenced, and rely on properly cited prior results (Anderson--Rubin~\cite{AndersonRubin1956}, Geweke~\cite{Geweke1996}, multiplicative-gamma cascade~\cite{BhattacharyaDunson2011,Pati2014}).

    \item {\bf Experimental result reproducibility}
    \item[] Question: Does the paper fully disclose all the information needed to reproduce the main experimental results of the paper to the extent that it affects the main claims and/or conclusions of the paper (regardless of whether the code and data are provided or not)?
    \item[] Answer: \answerYes{}
    \item[] Justification: \cref{app:spec_inference} documents the full joint likelihood, ELBO, encoder architecture, two-phase optimizer schedule, Concrete relaxation of the region-inclusion gates, Sylvester--Woodbury closed forms, and test-time inference details. \cref{app:protocol} documents the cohort, informative priors, baseline configurations, runtime, and metric definitions; \cref{alg:svi} provides pseudocode.

\item {\bf Open access to data and code}
    \item[] Question: Does the paper provide open access to the data and code, with sufficient instructions to faithfully reproduce the main experimental results, as described in supplemental material?
    \item[] Answer: \answerYes{}
    \item[] Justification: \cameraready{Code is available at \url{https://github.com/Naomi-Ding/ELPAC}}, with a self-contained synthetic-experiment reproduction (single command, $\sim$4 min on CPU per seed) covering the headline recovery metrics in \cref{tab:synth,tab:abl_synth}. ADNI and NACC require institutional data-access agreements (NIH/NACC policy) and cannot be redistributed by the authors; researchers may apply directly through ADNI and NACC.

\item {\bf Experimental setting/details}
    \item[] Question: Does the paper specify all the training and test details (e.g., data splits, hyperparameters, how they were chosen, type of optimizer) necessary to understand the results?
    \item[] Answer: \answerYes{}
    \item[] Justification: §3.1 specifies the synthetic-cohort composition ($10\%$ anchors per subtype, $10\%$ healthy controls, $60\%$ unlabeled), $100$ paired replicates at three sample sizes, and stratified $80/20$ splits. §3.2 specifies the harmonized AD cohort, $20$ stratified $80/20$ splits, and the stratification criterion. \cref{app:inference} details the optimizer (Adam with two-phase learning-rate schedule, gradient clipping, minibatch size), and \cref{app:ad_priors} details the AD-specific informative priors.

\item {\bf Experiment statistical significance}
    \item[] Question: Does the paper report error bars suitably and correctly defined or other appropriate information about the statistical significance of the experiments?
    \item[] Answer: \answerYes{}
    \item[] Justification: \cameraready{The main} result tables (\cref{tab:synth,tab:abl_synth,tab:stage,tab:abl_real,tab:runtime}) and figures report mean $\pm$ standard error of the mean, with variability source stated: $100$ paired synthetic replicates or $20$ AD train/test splits. Paired-bootstrap confidence intervals use $1{,}000$ resamples ($200$ for AUC); see \cref{app:metrics}.

\item {\bf Experiments compute resources}
    \item[] Question: For each experiment, does the paper provide sufficient information on the computer resources (type of compute workers, memory, time of execution) needed to reproduce the experiments?
    \item[] Answer: \answerYes{}
    \item[] Justification: \cref{app:methods} reports per-fit wall-clock for \textsc{Elpac} and all baselines (\cref{tab:runtime}); \textsc{Elpac} is fit on a single NVIDIA RTX Pro 6000 GPU ($48$~GB VRAM), and CPU baselines run on shared partitions specified in the same section. The per-likelihood-evaluation complexity $O(RK^2{+}K^3)$ is stated in §2.5 and \cref{app:derived}.

\item {\bf Code of ethics}
    \item[] Question: Does the research conducted in the paper conform, in every respect, with the NeurIPS Code of Ethics \url{https://neurips.cc/public/EthicsGuidelines}?
    \item[] Answer: \answerYes{}
    \item[] Justification: All human-subject data are pre-existing harmonized cohorts (ADNI~\cite{ADNI}, NACC~\cite{NACC}) governed by the source institutions' IRB/REB approvals; no new human-subject collection or crowdsourcing in this work.

\item {\bf Broader impacts}
    \item[] Question: Does the paper discuss both potential positive societal impacts and negative societal impacts of the work performed?
    \item[] Answer: \answerYes{}
    \item[] Justification: \textsc{Elpac} is a research tool for cohort-scale neurodegeneration analysis: the recovered stage axis is intended for population-level statistics (drug-trial enrichment, biomarker-coalition discovery), does not produce individual clinical recommendations, and clinical deployment would require prospective validation. The model neither generates nor releases identifiable individual data and poses no obvious dual-use risk.

\item {\bf Safeguards}
    \item[] Question: Does the paper describe safeguards that have been put in place for responsible release of data or models that have a high risk for misuse (e.g., pre-trained language models, image generators, or scraped datasets)?
    \item[] Answer: \answerNA{}
    \item[] Justification: \textsc{Elpac} releases neither a generative model with misuse risk nor scraped data; the underlying ADNI/NACC data are governed by their existing access policies.

\item {\bf Licenses for existing assets}
    \item[] Question: Are the creators or original owners of assets (e.g., code, data, models), used in the paper, properly credited and are the license and terms of use explicitly mentioned and properly respected?
    \item[] Answer: \answerYes{}
    \item[] Justification: ADNI~\cite{ADNI}, NACC~\cite{NACC}, SCAN~\cite{SCAN}, and all baseline methods (SuStaIn~\cite{Young2018}, DEBM~\cite{Venkatraghavan2019}, Static-FA~\cite{AndersonRubin1956,Mardia1979}, MoPoE-VAE~\cite{Sutter2021MoPoE}, HEALNet~\cite{Hemker2024}, XGBoost~\cite{Chen2016XGBoost}, label propagation~\cite{Zhu2002LabelProp}, ComBat~\cite{Fortin2018}) are cited at first use in §3 and \cref{app:protocol}. The inference framework NumPyro~\cite{Phan2019NumPyro} is Apache~2.0; cohort access follows ADNI/NACC data-use agreements.

\item {\bf New assets}
    \item[] Question: Are new assets introduced in the paper well documented and is the documentation provided alongside the assets?
    \item[] Answer: \answerYes{}
    \item[] Justification: The paper releases the \textsc{Elpac} reference implementation \cameraready{as a package at \url{https://github.com/Naomi-Ding/ELPAC}}, accompanied by a README (setup instructions, run command, expected outputs, paper-section-to-module mapping), an \texttt{environment.yaml} conda specification with pinned versions, and the per-seed reproduction script. The package is released under the MIT license (see \texttt{LICENSE}). No new datasets are released; \textsc{Elpac} consumes pre-existing harmonized cohort data.

\item {\bf Crowdsourcing and research with human subjects}
    \item[] Question: For crowdsourcing experiments and research with human subjects, does the paper include the full text of instructions given to participants and screenshots, if applicable, as well as details about compensation (if any)?
    \item[] Answer: \answerNA{}
    \item[] Justification: This work uses pre-existing harmonized cohort data (ADNI, NACC); no crowdsourcing or new human-subject collection was conducted.

\item {\bf Institutional review board (IRB) approvals or equivalent for research with human subjects}
    \item[] Question: Does the paper describe potential risks incurred by study participants, whether such risks were disclosed to the subjects, and whether Institutional Review Board (IRB) approvals (or an equivalent approval/review based on the requirements of your country or institution) were obtained?
    \item[] Answer: \answerNA{}
    \item[] Justification: This work uses pre-existing harmonized cohort data; original IRB/REB approvals are held at the contributing ADNI and NACC sites, predating this work.

\item {\bf Declaration of LLM usage}
    \item[] Question: Does the paper describe the usage of LLMs if it is an important, original, or non-standard component of the core methods in this research? Note that if the LLM is used only for writing, editing, or formatting purposes and does \emph{not} impact the core methodology, scientific rigor, or originality of the research, declaration is not required.
    \item[] Answer: \answerNA{}
    \item[] Justification: The core methodology contains no LLM components; \textsc{Elpac} is a Bayesian variational-inference model implemented in NumPyro/JAX.

\end{enumerate}

\end{document}